\documentclass{article} %
\usepackage{preprint,times}

\usepackage{amsmath,amsfonts,bm}

\def\eqref#1{equation~\ref{#1}}

\def\1{\bm{1}}

\DeclareMathAlphabet{\mathsfit}{\encodingdefault}{\sfdefault}{m}{sl}
\SetMathAlphabet{\mathsfit}{bold}{\encodingdefault}{\sfdefault}{bx}{n}

\usepackage{hyperref}
\usepackage{xurl} %
\usepackage{graphicx}
\usepackage{mathtools}
\usepackage{amsmath}
\usepackage[ruled,vlined,linesnumbered]{algorithm2e}
\SetAlgoNlRelativeSize{-1} 
\SetKwInput{KwIn}{\hspace*{-\algomargin}Input}
\SetKwInput{KwOut}{\hspace*{-\algomargin}Output}
\usepackage{multirow}
\usepackage{float}
\usepackage{placeins} 
\usepackage{titletoc} 
\usepackage{booktabs}
\usepackage{colortbl}
\usepackage{xcolor}
\usepackage{makecell}
\usepackage{subcaption}
\usepackage{wrapfig}
\usepackage{upgreek}
\usepackage{pifont}
\usepackage{nicefrac}
\usepackage{marvosym}
\usepackage{enumitem}
\usepackage{listings}
\usepackage[skins,breakable]{tcolorbox}

\hypersetup{
  citebordercolor={0.667 0.694 0.969},
  urlbordercolor={0.667 0.694 0.969},
  pdftitle={Consistent Plan-Act for Long-Horizon Agentic Tasks},
  pdfauthor={Heng-Zhuang Li, Yi-Kai Zhang, Yu Wang, Yueqing Sun, Jiayuan Zhang, Qi Gu, Han-Jia Ye},
}

\newcommand{\TableCols}{13}
\newcommand{\GroupHead}[1]{\rowcolor{black!8}\multicolumn{\TableCols}{c}{\textit{#1}}}
\definecolor{OursHighlight}{HTML}{AAB1F7}
\definecolor{AssertionString}{HTML}{343B78}
\lstdefinestyle{assertionjson}{
  basicstyle=\ttfamily\footnotesize,
  columns=fullflexible, keepspaces=true,
  showstringspaces=false, breaklines=true,
  aboveskip=0pt, belowskip=0pt,
  morestring=[b]", stringstyle=\color{AssertionString},
  morekeywords={null}, keywordstyle=\color{AssertionString}\bfseries
}
\newtcolorbox{assertionexample}[1]{
  enhanced, colback=black!1, colframe=OursHighlight!65!black,
  colbacktitle=OursHighlight!22, coltitle=black,
  boxrule=0.45pt, arc=1mm,
  left=7pt, right=7pt, top=6pt, bottom=6pt,
  fonttitle=\small\bfseries, title={#1},
  before skip=8pt, after skip=9pt
}
\lstdefinestyle{conpactprompttext}{
  basicstyle=\ttfamily\footnotesize,
  columns=fullflexible, keepspaces=true,
  showstringspaces=false, breaklines=true, breakatwhitespace=false,
  aboveskip=0pt, belowskip=0pt
}
\newtcolorbox{conpactprompt}[1]{
  enhanced, breakable, lines before break=4,
  colback=black!1, colframe=OursHighlight!65!black,
  colbacktitle=OursHighlight!22, coltitle=black,
  boxrule=0.45pt, arc=1mm,
  left=7pt, right=7pt, top=6pt, bottom=6pt,
  fonttitle=\small\bfseries, title={#1},
  title after break={#1 (continued)},
  before skip=8pt, after skip=9pt
}
\newcommand{\OursRow}{\rowcolor{OursHighlight!20}[\tabcolsep][\dimexpr\tabcolsep+0.4pt\relax]}
\newlength{\MetricWidth}
\newcolumntype{E}{>{\centering\arraybackslash}p{\MetricWidth}}
\newcolumntype{O}{>{\columncolor{OursHighlight!20}[\tabcolsep][0pt]\centering\arraybackslash}p{\MetricWidth}}
\newcolumntype{I}{!{\color{black!35}\vrule width 0.4pt}}

\title{Consistent Plan-Act for Long-Horizon\\Agentic Tasks}

\author{Heng-Zhuang Li$^{1, 2, 3, }$\thanks{Equal contribution.}\quad Yi-Kai Zhang$^{1, 2, \ast}$\quad Yu Wang$^{3, 4}$\quad Yueqing Sun$^{3}$\\
\bfseries Jiayuan Zhang$^{3}$\quad Qi Gu$^{3, }$\thanks{Corresponding to \texttt{guqi03@meituan.com} and \texttt{yehj@lamda.nju.edu.cn}.}\quad\quad Han-Jia Ye$^{1, 2, \text{\Letter}}$\\
$^1$School of Artificial Intelligence, Nanjing University\\
$^2$National Key Laboratory for Novel Software Technology, Nanjing University\\
$^3$LongCat Team, Meituan\quad $^4$University of Science and Technology of China\\
}

\date{}
\begin{document}

\begingroup
\makeatletter
\let\originalfnsymbol\@fnsymbol
\renewcommand{\@fnsymbol}[1]{%
  \ifnum#1=2\relax\text{\Letter}\else\originalfnsymbol{#1}\fi}
\makeatother
\maketitle
\endgroup

\begin{abstract}

Long-horizon agentic tasks demand strong reasoning and efficient execution across successive interactions with dynamic environments.
A common approach decouples high-level planning from low-level execution through separate planner and actor roles.
To investigate coordination failures in these tasks, we prompt both agents for structured state assertions and compare their reports programmatically to detect explicit contradictions.
Our analyses reveal systematic disagreement about the same task-relevant state facts, a phenomenon we term \textbf{\textit{planner-actor state mismatch}}.
We further find that providing agents with task-relevant state information reduces mismatch and improves coordination and task performance.
Based on the systematic analysis of the state mismatch, we propose \textbf{Con}sistent \textbf{P}lan-\textbf{Act} (\textbf{ConPAct}), which feeds detected contradictions back to both agents to form consistent state interpretations and fine-tunes them on curated consistent interactions for better coordination.
ConPAct improves performance across various environments and model configurations, e.g., increasing MiniGrid success rate from 38.6\% to 54.4\% with GPT-5.6-sol/terra as planner and actor respectively, demonstrating that state consistency can guide both inference-time correction and coordination training.
Code: \url{https://github.com/Fir-lat/ConPAct}.

\end{abstract}

\section{Introduction}

\suppressfloats[t]
\begin{figure*}[t]
\centering
\begin{subfigure}[b]{0.66\textwidth}
  \centering
  \begin{minipage}[c][0.5\linewidth][c]{\linewidth}
    \includegraphics[width=\linewidth]{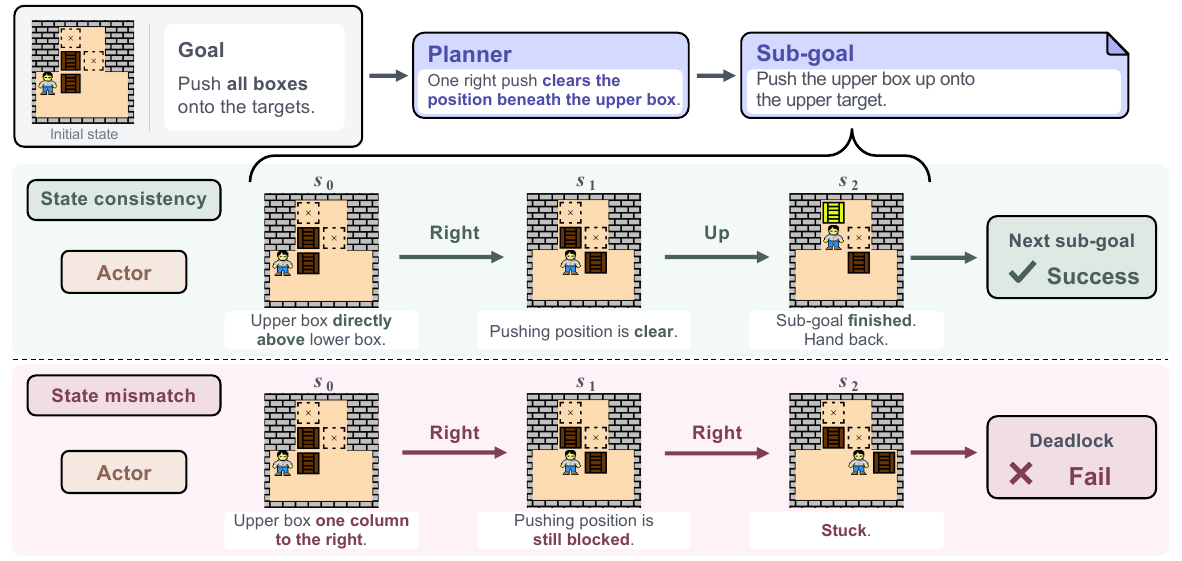}
  \end{minipage}
  \vspace{-8pt}
  \caption{Planner-Actor state mismatch and consistency.}
  \label{fig:case_study}
\end{subfigure}\hfill
\begin{subfigure}[b]{0.33\textwidth}
  \centering
  \begin{minipage}[c][\linewidth][c]{\linewidth}
    \includegraphics[width=\linewidth]{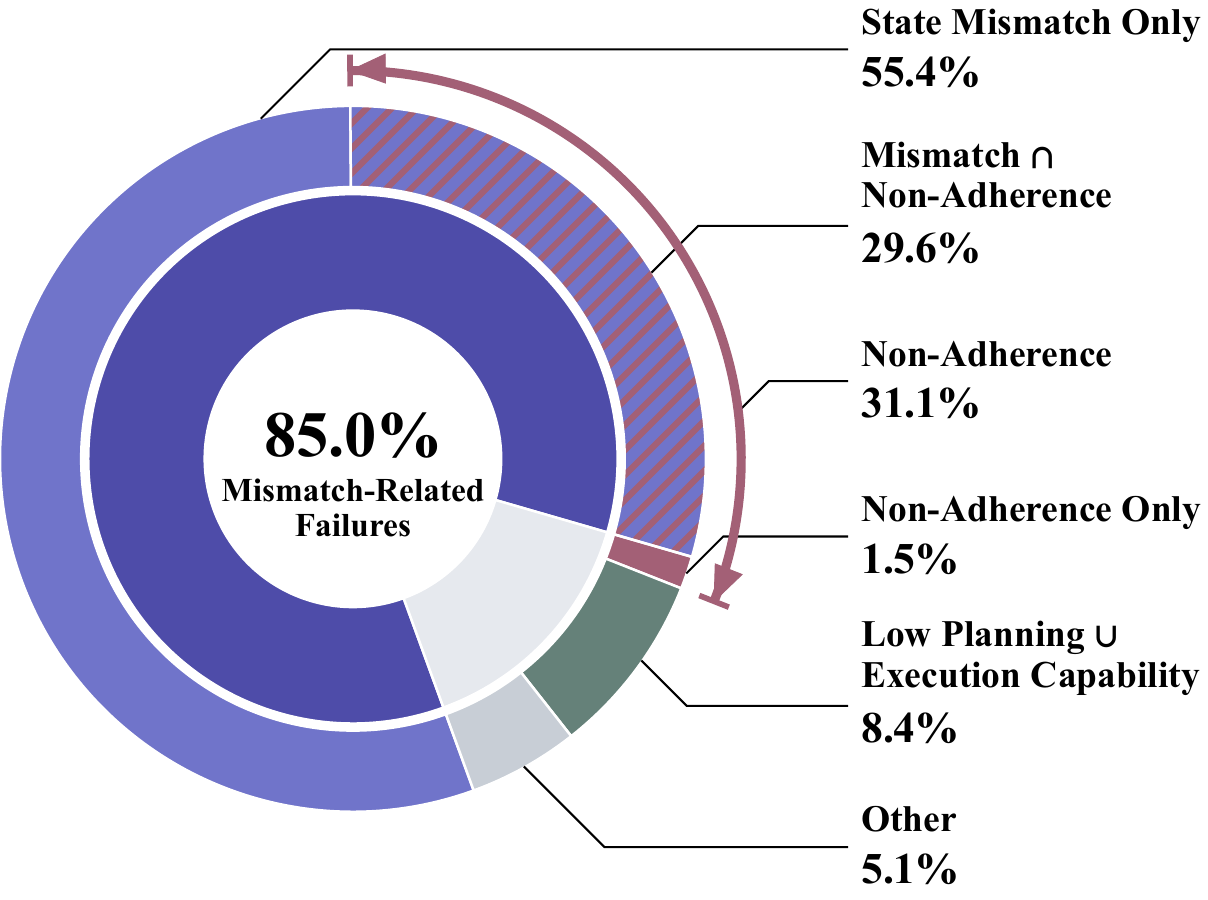}
  \end{minipage}
  \vspace{-8pt}
  \caption{Failure cases.}
  \label{fig:failure_mismatch_overview}
\end{subfigure}
\vspace{-8pt}
\caption{\textbf{Planner-actor state mismatch}.
(a) Misreading the upper box as one column farther right, the actor pushes \texttt{Right} twice, causing deadlock; the correct sequence is \texttt{Right} $\rightarrow$ \texttt{Up}, with consistent interpretation as the planner.
(b) Mismatch occurs in \textbf{85.0\%} of failed trajectories, far exceeding non-adherence (31.1\%) and low planning/execution capability (8.4\%).}
\label{fig:state_mismatch_overview}
\vspace{-16pt}
\end{figure*}

Large language models (LLMs)~\citep{openai2026gpt5,deepseekai2026deepseekv4} increasingly serve as agents in interactive environments, including games~\citep{paglieri2025balrog}, web navigation~\citep{yao2022webshop,zhou2024webarena}, GUI automation~\citep{rawles2023androidinthewild,xie2024osworld,rawles2025androidworld}, and embodied tasks~\citep{shridhar2021alfworld,ahn2022saycan,liu2023libero}.
These tasks require agents to combine long-term planning~\citep{chen2025enhancing,li2026beyond} with action selection guided by observations and environmental feedback~\citep{wang2025steca}.
Alongside frameworks that interleave reasoning and action~\citep{yao2023react} or incorporate reflection on task feedback~\citep{shinn2023reflexion}, plan-act architectures assign high-level planning and low-level execution to separate agents~\citep{erdogan2025planandact,jin2026hira,molinari2026reasonplanreact}.
A planner translates the overall objective into sub-goals~\citep{hu2025divide}, while an actor selects and adapts actions to achieve them~\citep{erdogan2025planandact,peng2026hiper}.
This division reduces the burden of simultaneously managing planning and execution~\citep{erdogan2025planandact} and allows models with different capabilities to serve the two roles~\citep{tan2026from}.
Their coordination through sub-goals and execution feedback requires common ground~\citep{clark1991grounding}: compatible interpretations of the task-relevant information referenced in sub-goals and execution feedback.

We use GPT-5.6-sol~\citep{openai2026gpt56} to examine sub-goal execution in failed plan-act trajectories across tasks and model configurations.
Categorizing these failures reveals recurring disagreements between the planner and actor about the same task-relevant environment state.
These disagreements can lead the actor to deviate from the planner's intent or treat feasible sub-goals as infeasible, contributing to task failure.
This failure mode differs from limited planning~\citep{li2026hiplan,tan2026from,dou2026plan} or execution~\citep{hu2025divide,zhen2026hierarchical,wang2025steca} capabilities, and sub-goal non-adherence~\citep{jia2025what,kim2026ppaplan,venkatesan2026plover}. We term it \textit{\textbf{planner-actor state mismatch}}.
Figure~\ref{fig:case_study} illustrates how individually coherent decisions can fail to support coordinated progress.
State mismatch occurs at least once in \textbf{85\%} of the analyzed failed trajectories (Figure~\ref{fig:failure_mismatch_overview}).
Improving planning~\citep{erdogan2025planandact,li2026hiplan}, execution~\citep{hu2025divide}, or adherence~\citep{jeong2026tape} does not by itself require the two roles to agree on the state facts underlying their decisions.
Prior work examines inter-agent failures~\citep{cemri2025mast}, explicit belief-state representations~\citep{li2023theory}, and belief alignment through dialogue~\citep{qiu2024minddial,dongre2026embodied}; here we focus on agreement about the same observed state facts at planner-actor handoffs.

To measure state mismatch, we use state assertions inspired by explicit belief-state representations~\citep{li2023theory}.
Each agent reports its perceived task-relevant state in a shared structured format alongside its sub-goal or action.
These reports serve as proxies for state interpretations, enabling programmatic detection of explicit contradictions without ground truth, or external verifiers. Our analysis is summarized as follows.
First, mismatch occurs across environments and models and increases with the capability gap between roles.
Controlled comparisons show that it persists under both visual and textual observations, with cross-role disagreement exceeding within-role sampling variability.
Second, controlled interventions establish its causal impact on coordination: correcting the actor's state interpretation restores coordinated behavior while holding the sub-goal and observation fixed.
Finally, our mismatch metric enables us to examine whether providing reliable state information improves agreement between the roles. Supplying state assertions from a stronger model or environment oracle reduces the measured disagreement while increasing task success.

Building on these findings, we propose \textbf{Con}sistent \textbf{P}lan-\textbf{Act} (\textbf{ConPAct}), which feeds programmatically detected state contradictions back to both agents.
Unlike model-generated critiques~\citep{madaan2023selfrefine,du2024multiagentdebate}, this feedback identifies mutual conflicting fields in their state assertions without any external information.
The agents then discuss these disagreements and revise their interpretations, sub-goals, or actions before subsequent execution. There are three variants:
\textbf{ConPAct-I}nference applies this intervention during inference without training.
\textbf{ConPAct-S}ampling trains on sampled collaboration and correction sequences selected for state consistency, whereas \textbf{ConPAct-R}eplaying uses a teacher model to reconstruct such supervision from existing ReAct~\citep{yao2023react} trajectories.
Experiments across various environments demonstrate benefits from both inference-time intervention and training.
On MiniGrid~\citep{chevalierboisvert2023minigrid,chevalierboisvert2019babyai}, ConPAct-I raises success from 38.6\% to 54.4\% with a GPT-5.6-sol planner and GPT-5.6-terra actor~\citep{openai2026gpt56}.
ConPAct-S reaches 39.7\% with Qwen3-VL-32B~\citep{bai2025qwen3vl}, exceeding the strongest evaluated training baseline by a large margin (29.1\%).
Ablations support the contributions of explicit conflict feedback, participation by both roles, and supervision of correction sequences.
Our contributions are summarized as follows:

\begin{itemize}[leftmargin=*, topsep=0pt, partopsep=0pt, itemsep=2pt, parsep=0pt]
\item We identify and characterize \textbf{\textit{planner-actor state mismatch}}, a relational failure mode. We establish its prevalence, impact on coordination, and mitigation potential through systematic analyses.
\item We develop \textbf{ConPAct}, which uses programmatically detected state contradictions to guide joint revision during inference and trains agents on consistent collaboration and correction sequences obtained through sampling or replaying from existing trajectories.
\item Experiments across environments and model configurations demonstrate improved coordination and task performance, with ablations supporting the contributions of each components.
\end{itemize}

\section{Characterizing Planner-Actor State Mismatch}
\label{sec:characterizing}

\begin{figure}[t]
  \centering
  \begin{subfigure}[t]{0.51\textwidth}
    \centering
    \includegraphics[height=1.74in,keepaspectratio]{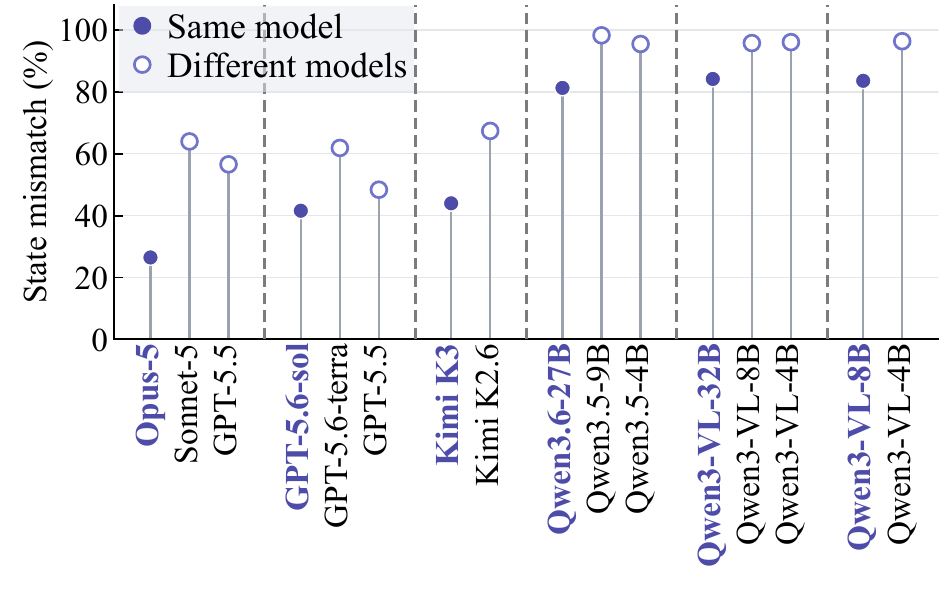}
    \vspace{-0pt}
    \caption{Prevalence of planner--actor state mismatch.}
    \label{fig:mismatch_prevalence}
  \end{subfigure}\hfill
  \begin{subfigure}[t]{0.485\textwidth}
    \centering
    \includegraphics[height=1.74in,keepaspectratio]{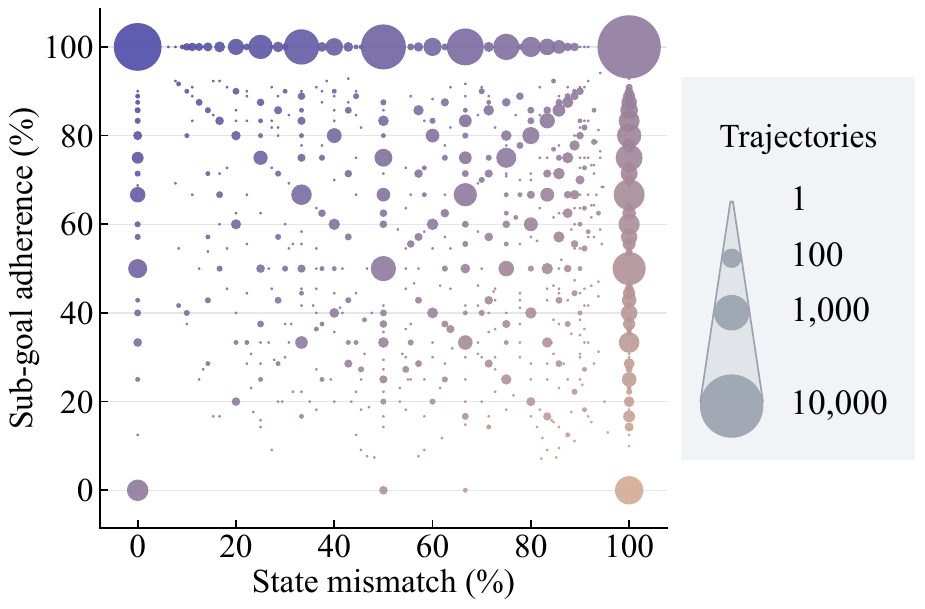}
    \vspace{-0pt}
    \caption{State mismatch despite high sub-goal adherence.}
    \label{fig:adherence_mismatch}
  \end{subfigure}
  \vspace{-0pt}
\caption{\textbf{Prevalence of planner-actor state mismatch.} (a) Mismatch occurs across various model configurations, with even homogeneous configurations have notable state mismatch; (b) State mismatch coexists with high sub-goal adherence, with marker size indicating trajectory frequency. Model sources are provided in Appendix~\ref{app:models-baselines}.}
  \label{fig:mismatch_overview}
  \vspace{-0pt}
\end{figure}

\subsection{Problem Setup and Measurement}
\label{problem:plan_act_agents}

\textbf{Long-Horizon Agentic Tasks.}
We study planner-actor state mismatch in Sokoban~\citep{hu2025lmgame} and MiniGrid~\citep{chevalierboisvert2023minigrid,chevalierboisvert2019babyai}, which require agents to pursue goals over extended sequences of interdependent actions and coordinate planning and execution throughout the task.
Their underlying grid states are also readily accessible, providing ground truth for evaluating agents' state interpretations.
We evaluate 200 Sokoban levels and 150 MiniGrid levels, with eight rollouts per level.
Our evaluation covers Claude~\citep{anthropic2026claudeopus5,anthropic2026claudesonnet5}, GPT~\citep{openai2026gpt56,openai2026gpt55}, Kimi~\citep{kimiteam2026kimik3openfrontier,moonshotai2026kimik26}, and Qwen~\citep{qwenteam2026qwen3627b,qwenteam2026qwen35,bai2025qwen3vl}, using homogeneous/heterogeneous model configurations.

\textbf{Plan-Act Agents.}
Given task instruction $x_i$, trajectory $\tau_i$ consists of model-call turns $t$, each with observation $o_{i,t}$ and response $y_{i,t}$.
The planner generates a sub-goal $g_{i,t}$ from its observation; the actor uses the latest sub-goal and its own observation to select action $a_{i,t}$ and report execution status $z_{i,t}$~\citep{erdogan2025planandact,peng2026hiper}.
Consecutive actor turns under one sub-goal form an \emph{execution segment}.
Completion or execution difficulties trigger \textit{dynamic replanning}, where the planner revises the sub-goal using the latest observation and actor feedback~\citep{erdogan2025planandact,lin2023swiftsage,paglieri2025learning}.
Coordination through sub-goals and feedback thus depends on the roles' interpretations of task-relevant state.

\label{problem:state_assertion}
\textbf{State Assertions.}
We prompt each role to report its perceived state as \texttt{JSON} alongside its sub-goal or action.
For each environment, we specify a compact schema $\mathcal F$ shared by both roles, with fixed fields and a predefined value domain $\mathcal V_f$ for each field $f$.
Both roles use the shared schema $\mathcal F$ describing task-relevant properties inferable from a single observation, such as object identities, action prerequisites, and goal conditions.
Each field $f$ has a predefined set $\mathcal V_f$ of canonical, mutually exclusive values.
Parsing response $y_{i,t}$ yields the \emph{state assertion}
$\mathbf v_{i,t}=\operatorname{parse}_{\mathcal F}(y_{i,t})$, where $\mathbf v_{i,t}[f]\in\mathcal V_f$.
At planner-actor handoffs, we compare assertions from consecutive role calls on the same observation ($o_{i,t}=o_{i,t+1}$), without either role seeing the other's assertion.
For a valid pair, we detect \emph{state mismatch} if any corresponding field differs:
\begin{equation}
d_{i,t} = \mathbf{1}\!\left[\exists f\in\mathcal F:\mathbf v_{i,t}[f]\neq\mathbf v_{i,t+1}[f]\right].
\label{eq:state_mismatch}
\end{equation}

\textbf{Metrics for analyzing state mismatch.}
(1) \emph{State mismatch rate} ($D$) is the fraction of valid handoff pairs with $d_{i,t}=1$;
(2) \emph{State correctness} ($A^{\text{plan}}$, $A^{\text{act}}$) is each role's exact-match assertion accuracy against environment ground truth on the same paired observations;
(3) \emph{Sub-goal adherence} ($H$) is the fraction of execution segments in which an LLM judge~\citep{zheng2023mtbench} determines that the actor pursues the assigned sub-goal;
(4) \emph{Task success} ($S$) is the fraction of trajectories with environment-confirmed task completion.
The detailed definitions are placed in Appendix~\ref{app:add_pre}.

\subsection{Prevalence of Planner-Actor State Mismatch}
\label{subsec:mismatch-existence}

\begin{figure}[t]
  \centering
  \begin{subfigure}[t]{0.300\textwidth}
    \centering
    \includegraphics[width=\textwidth]{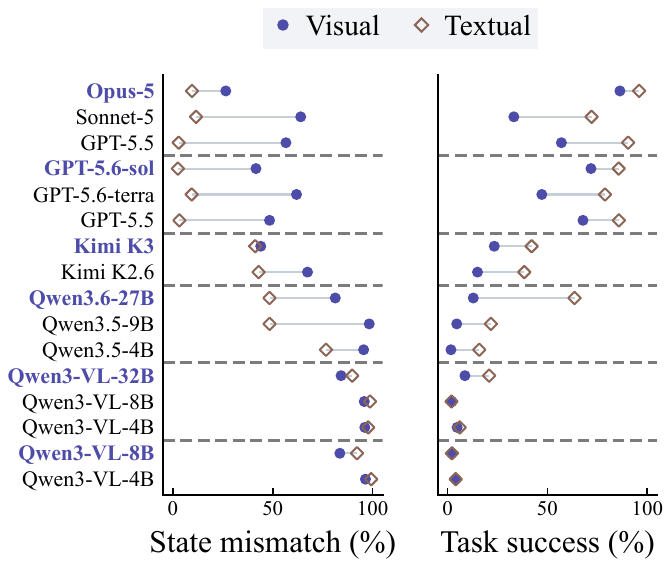}
    \caption{Visual vs. textual mismatch.}
    \vspace{-8pt}
    \label{fig:visual_text_mismatch}
  \end{subfigure}\hfill
  \begin{subfigure}[t]{0.355\textwidth}
    \centering
    \includegraphics[width=\textwidth]{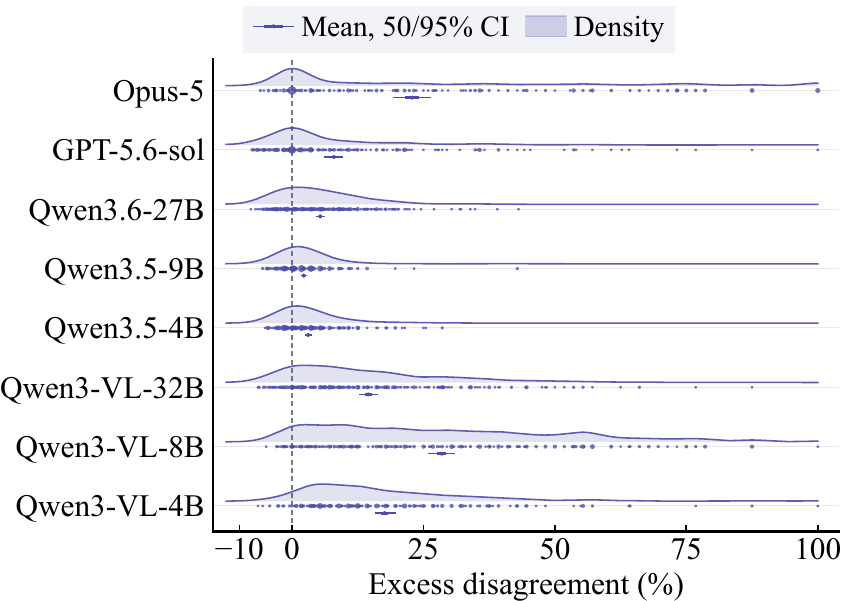}
    \caption{Excess cross-role disagreement.}
    \vspace{-8pt}
    \label{fig:sampling_noise_visual}
  \end{subfigure}\hfill
  \begin{subfigure}[t]{0.332\textwidth}
    \centering
    \includegraphics[width=\textwidth]{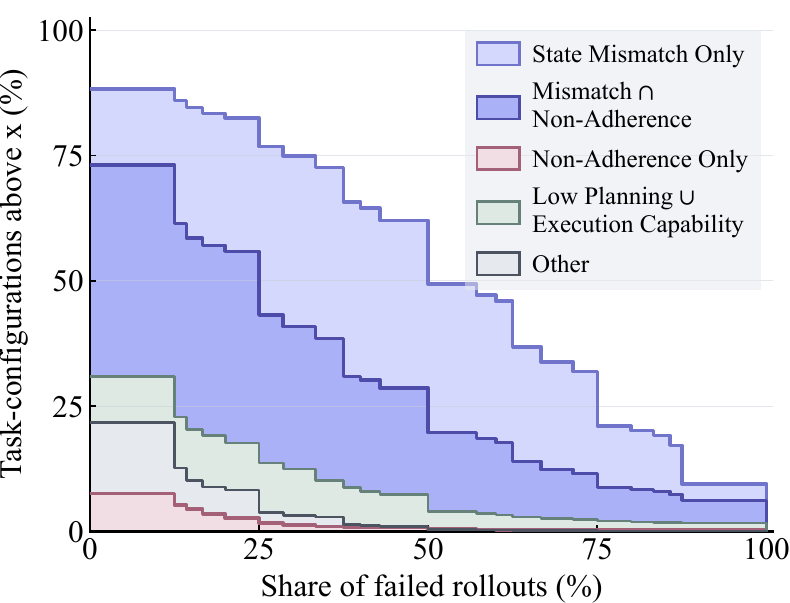}
    \caption{Failure patterns.}
    \vspace{-8pt}
    \label{fig:failure_patterns_ecdf}
  \end{subfigure}
  \vspace{-0pt}
  \caption{Planner-actor state mismatch is systematic and prevalent in failed trajectories. (a) Mismatch persists under both visual and textual observations; (b) mean cross-role disagreement exceeds within-role sampling variability; (c) mismatch-only failures dominate across task configurations.}
  \vspace{-16pt}
  \label{fig:mismatch_alternatives}
\end{figure}

\textbf{Prevalence of state mismatch.}
As shown in Figure~\ref{fig:mismatch_prevalence}, mismatch occurs in all combinations, reaching 26.5\% and 41.6\% for homogeneous Opus-5 and GPT-5.6-sol, respectively.
Figure~\ref{fig:adherence_mismatch} illustrates that high sub-goal adherence coexists with substantial state mismatch.
Specifically, 82.9\% of those with measured adherence $H=100\%$ contain at least one detected mismatch.

\textbf{State mismatch across observation modalities.}
In Figure~\ref{fig:visual_text_mismatch}, replacing screenshots with structured textual states reduces mismatch in stronger configurations, but mismatch persists in all configurations.
Even homogeneous Opus-5 and GPT-5.6-sol still exhibit rates of 9.6\% and 2.5\%, respectively.
Therefore, planner-actor state mismatch extends beyond visual perception difficulties.

\textbf{State mismatch beyond sampling variability.}
To test whether sampling variability explains mismatch, we repeatedly sample on 300 fixed observations while keeping each role's model and context fixed.
We measure disagreement between state assertions from repeated sampling within each role and average the planner's and actor's rates to obtain a sampling baseline.
Cross-role disagreement exceeds this baseline for all eight models under visual observation, with all 95\% confidence intervals for the excess excluding zero (Figure~\ref{fig:sampling_noise_visual}).
For Opus-5, the excess is 22.8 percentage points, indicating that planner-actor disagreement contains systematic differences beyond sampling variability.

\subsection{Behavioral Implications of State Mismatch}
\label{subsec:mismatch-coordination}

\textbf{State mismatch dominates failure patterns across task configurations.}
Figure~\ref{fig:failure_patterns_ecdf} plots the fraction of level-configuration pairs exceeding each within-pair failure-share threshold, considering only pairs with failed rollouts.
Mismatch-only failures exceed these thresholds in more pairs than non-adherence-only or low-capability failures.
In 49.4\% of these pairs, mismatch without detected non-adherence occurs in more than half of all failures.
Thus, conflicting state interpretations characterize most failures in a broad range of task configurations.

\textbf{Controlled interventions.}
We keep the observation, assigned sub-goal, and actor fixed within each event and vary only the supplementary information.
We compare four conditions: no supplementary information, a state assertion with random field values to control for structured formatting, freeform control context, and task-relevant state facts from Opus-5.
Freeform control context consists of character-matched irrelevant text for blocked sub-goals and non-critical state facts for feasible sub-goals.
Valid handback rate measures handoff signals in the first response; first-step progress rate measures first actions that reduce distance to the sub-goal.

\begin{figure}[!t]
  \centering
  \small
  \newlength{\stateEffectsHeight}
  \setlength{\stateEffectsHeight}{0.307\textwidth}
  \begin{subfigure}[t]{0.307\textwidth}
    \centering
    \includegraphics[height=\stateEffectsHeight]{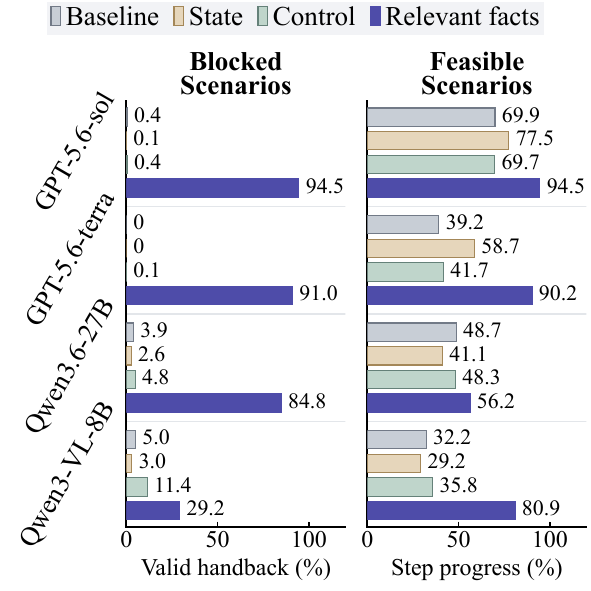}
    \caption{Behavioral implications.}
    \vspace{-8pt}
    \label{fig:state_behavior}
  \end{subfigure}\hfill
  \begin{subfigure}[t]{0.307\textwidth}
    \centering
    \includegraphics[height=\stateEffectsHeight]{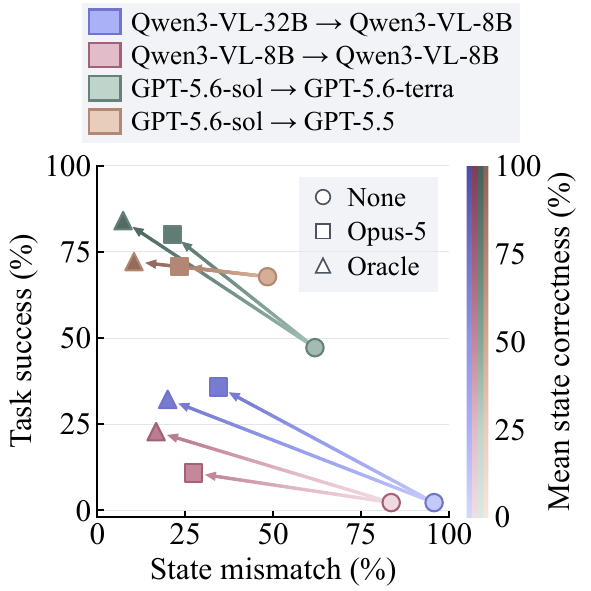}
    \caption{Injecting state assertions.}
    \vspace{-8pt}
    \label{fig:state_injection}
  \end{subfigure}\hfill
  \begin{subfigure}[t]{0.3443\textwidth}
    \centering
    \includegraphics[height=\stateEffectsHeight]{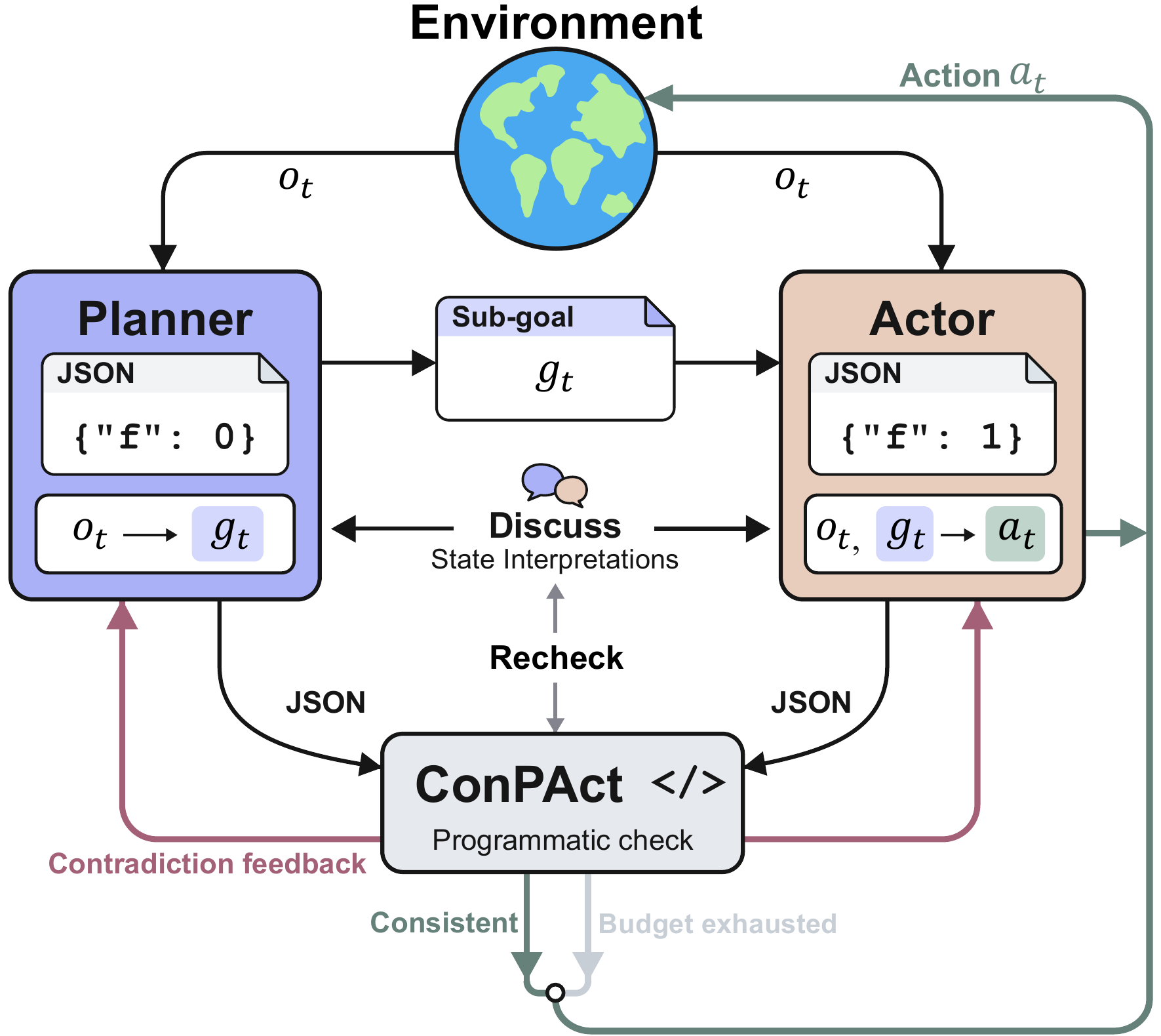}
    \caption{\textbf{ConPAct-I} framework.}
    \vspace{-8pt}
    \label{fig:conpact_framework}
  \end{subfigure}
  \vspace{-0pt}
  \caption{State information and consistent plan-act coordination. (a) Task-relevant state facts improve the actor's decisions. (b) Injecting reliable state assertions reduces mismatch and improves task success. (c) ConPAct feeds detected contradictions back for coordination reconciliation.}
  \vspace{-16pt}
  \label{fig:motivation_state_effects}
\end{figure}

\textbf{Task-relevant state information improves feedback and execution.}
The results are displayed in Figure~\ref{fig:state_behavior}. For blocked sub-goals, task-relevant facts increase GPT-5.6-sol's valid handback rate from 0.4\% to 94.5\%, compared with 0.1\% for random state assertions and 0.4\% for control context.
For feasible sub-goals, they increase Qwen3-VL-8B's first-step progress rate from 32.2\% to 80.9\%, compared with 29.2\% for random state assertions and 35.8\% for non-critical state facts.
The gains over both controls show that task-relevant state information helps actors adapt their coordination decisions to sub-goal feasibility, i.e., returning control for replanning when execution is blocked and selecting a progress-making action when execution is feasible.
Full results for mismatch prevalence, observation and sampling controls, and state interventions appear in Appendix~\ref{app:mismatch-analysis}.

\section{Consistent Plan-Act}

\subsection{Task-Level Implications and Consistency Interventions}
\label{subsec:coordination-task-completion}

To evaluate the task-level effects of reliable state information, we inject oracle assertions from environment states and assertions generated by Opus-5.
Relative to no injection, both sources improve assertion correctness, reduce mismatch, and increase task success consistently, shown in Figure~\ref{fig:state_injection}.
For GPT-5.6-sol paired with GPT-5.6-terra, oracle injection reduces mismatch from 61.9\% to 7.3\% and raises task success from 47.2\% to 84.1\%. Building on these gains and the behavioral results in Section~\ref{subsec:mismatch-coordination}, we propose \textbf{Con}sistent \textbf{P}lan-\textbf{Act} (\textbf{ConPAct}), which aligns planner-actor state interpretations through task-relevant information exchange.
Detected contradictions identify the state facts that require reassessment.
ConPAct feeds these contradictions back to both agents, directing their discussion toward the disputed facts and how they should inform planning and execution.

\subsection{Feeding Back Detected Contradictions for State Consistency}

As shown in Figure~\ref{fig:conpact_framework}, \textbf{ConPAct-I} compares state assertions about the same observation at planner--actor handoffs.
When definite values conflict, it feeds the disputed fields and both roles' judgments back to the agents.
Environment actions remain paused so that the discussion concerns a fixed state.
The planner and actor exchange their interpretations, reassess the disputed facts against their observations, and revise their state judgments and associated sub-goals or execution decisions.
We recheck their assertions after each exchange until all fields have definite, matching values or the discussion budget is exhausted.
Execution then resumes with the updated discussion context.
This process uses the agents' observations and exchanged messages, requiring neither environment ground truth nor an additional verifier model.

\subsection{Learning State-Consistent Coordination}

We further train agents to coordinate and resolve state disagreements using consistent interactions and corrective discussions.
\textbf{ConPAct-S} collects these interactions by sampling trajectories with ConPAct-I.
From both successful and failed trajectories, we retain consistent collaboration segments and correction sequences that reach full agreement across all assertion fields.
We supervise only valid replies without detected state contradictions, while keeping earlier inconsistent replies as context for learning how to revise conflicting judgments.
Each example supervises the role's complete response given its task, observation, and available interaction history.

Prior work derives plan supervision from execution traces~\citep{erdogan2025planandact} or relabels their goals in hindsight~\citep{andrychowicz2017hindsight,li2026spinning}.
When only existing ReAct~\citep{yao2023react} trajectories are available, \textbf{ConPAct-R} converts them into planner-actor interactions while preserving the recorded action sequence.
A teacher segments each trace into sub-goals and adds state assertions and role exchanges.
Around nonoptimal actions, it constructs hypothetical errors in state judgments or sub-goals, followed by feedback and revision, to provide correction examples.
We validate the reconstructed interactions against the recorded execution and review each candidate target using only the information available to the student at that turn.
We then apply the same consistency filtering and supervision rules as ConPAct-S, keeping hypothetical errors and nonoptimal actions as context only.
Algorithms and prompt templates for all three variants are provided in Appendix~\ref{app:method-details}.

\section{Evaluation}

\begin{table*}[t]
\centering
\small
\caption{Inference-time performance of Plan-Act methods. \textbf{Bold}/\underline{underlining} mark the best/second-best values, including ties. The ReAct (Actor) uses only the actor and is unranked.}
\label{tab:inference-core-results}
\setlength{\tabcolsep}{1.5pt}
\renewcommand{\arraystretch}{1.15}
\makebox[\linewidth][c]{\resizebox{\linewidth}{!}{%
\begin{tabular}{lIccccIccccIccccIcccc}
\toprule
 & \multicolumn{4}{Ic}{\makecell[c]{\textit{Opus-5}\\$\downarrow$\\\textit{Sonnet-5}}} & \multicolumn{4}{Ic}{\makecell[c]{\textit{GPT-5.6-sol}\\$\downarrow$\\\textit{GPT-5.6-terra}}} & \multicolumn{4}{Ic}{\makecell[c]{\textit{Kimi K3}\\$\downarrow$\\\textit{Kimi K2.6}}} & \multicolumn{4}{Ic}{\makecell[c]{\textit{Qwen3.6-27B}\\$\downarrow$\\\textit{Qwen3.5-9B}}} \\
\cmidrule(lr){2-5} \cmidrule(lr){6-9} \cmidrule(lr){10-13} \cmidrule(lr){14-17}
Method & Sok. & Cra. & Proc. & Mini. & Sok. & Cra. & Proc. & Mini. & Sok. & Cra. & Proc. & Mini. & Sok. & Cra. & Proc. & Mini. \\
\midrule
\textcolor{black!40}{ReAct (Actor)} & \textcolor{black!40}{70.9} & \textcolor{black!40}{4.74} & \textcolor{black!40}{20.55} & \textcolor{black!40}{33.4} & \textcolor{black!40}{66.2} & \textcolor{black!40}{5.10} & \textcolor{black!40}{17.21} & \textcolor{black!40}{31.4} & \textcolor{black!40}{45.7} & \textcolor{black!40}{3.39} & \textcolor{black!40}{11.23} & \textcolor{black!40}{21.9} & \textcolor{black!40}{6.2} & \textcolor{black!40}{3.14} & \textcolor{black!40}{2.29} & \textcolor{black!40}{7.9} \\
\midrule
Plan-Act & 84.2 & \underline{5.26} & 23.49 & \underline{33.1} & 75.1 & 5.13 & 17.20 & 38.6 & 57.7 & \textbf{4.23} & 10.85 & \underline{28.3} & 12.0 & 3.27 & 3.19 & \textbf{13.9} \\
HiPlan & \underline{91.3} & 5.23 & \underline{25.26} & 31.6 & \underline{76.7} & 5.13 & 17.95 & \underline{47.7} & \underline{65.5} & 3.29 & 11.85 & 27.1 & \underline{13.3} & \underline{3.32} & \underline{4.76} & 10.6 \\
TAPE & 90.5 & 5.17 & 21.25 & 29.2 & 75.9 & \textbf{5.24} & \underline{21.12} & 31.2 & 53.6 & 3.82 & \underline{12.15} & 21.5 & 10.3 & 3.01 & 2.11 & 8.6 \\
\OursRow \textbf{ConPAct-I} & \textbf{95.4} & \textbf{5.91} & \textbf{25.75} & \textbf{36.7} & \textbf{86.6} & \underline{5.14} & \textbf{22.83} & \textbf{54.4} & \textbf{65.8} & \underline{4.16} & \textbf{12.79} & \textbf{34.2} & \textbf{15.7} & \textbf{3.46} & \textbf{4.85} & \underline{13.5} \\
\bottomrule
\end{tabular}}}
\end{table*}

\subsection{Experimental Setup}

\textbf{Environments.}
We evaluate on Sokoban from GamingAgent~\citep{hu2025lmgame}, Crafter~\citep{hafner2022crafter}, Procgen~\citep{cobbe2020procgen}, and MiniGrid/BabyAI~\citep{chevalierboisvert2023minigrid,chevalierboisvert2019babyai}.
These environments cover box-pushing, survival and crafting, procedurally generated games, and instruction-following grid-world tasks.
The evaluation sets contain 200 Sokoban puzzles balanced between one and two boxes, 200 Crafter seeds, 140 levels from seven Procgen games in easy mode, and 150 instances spanning 27 MiniGrid and BabyAI task IDs.
Each instance is evaluated with eight rollouts.
The action and total model-call limits are both 60 for Sokoban, 160 for Crafter and Procgen, and 60--200 for MiniGrid, depending on the task.
We additionally evaluate desktop task performance on 361 OSWorld~\citep{xie2024osworld} tasks spanning ten application categories, with three rollouts per task.

\textbf{Configurations.}
ConPAct-I is evaluated with four planner--actor pairs from the Claude, GPT, Kimi, and Qwen families (Table~\ref{tab:inference-core-results}).
For supervised training, we use Qwen3-VL-8B and Qwen3-VL-32B~\citep{bai2025qwen3vl}, with either the same model size in both roles or a 32B planner paired with an 8B actor.
GPT-5.6-sol~\citep{openai2026gpt56}, with its default thinking effort, serves as the teacher for data construction, and the training-token budget is matched to three epochs of ReAct SFT.
ConPAct-S/R are evaluated with model-initiated discussion and without system-triggered ConPAct-I.

\textbf{Baselines.}
Inference baselines include ReAct~\citep{yao2023react} using either the planner or actor model, Plan-Act~\citep{erdogan2025planandact}, HiPlan~\citep{li2026hiplan}, and TAPE~\citep{jeong2026tape}.
Training baselines include ReAct SFT, Plan-Act SFT, PAA~\citep{erdogan2025planandact}, HSL~\citep{li2026spinning}, WebSTAR~\citep{he2026webstar}, and ECoT~\citep{zawalski2024robotic}.

\textbf{Evaluation metrics.}
We focus on Sokoban avg@8, Crafter Score~\citep{hafner2022crafter}, Procgen interquartile mean (IQM)~\citep{agarwal2021rliable}, and MiniGrid success rate (SR).
MiniGrid SR is averaged equally across the 27 task IDs; Procgen normalized-return metrics are multiplied by 100 for display.
For OSWorld, we report mean evaluation score and full success rate as Score/SR (\%).
State analyses use mismatch rate $D$ and assertion correctness $A^{\text{plan}}$ and $A^{\text{act}}$ from Section~\ref{problem:state_assertion}.
Ground-truth states serve only to evaluate correctness in these analyses.
Experimental configurations and metric definitions are provided in Appendices~\ref{app:experimental-setup} and~\ref{app:metrics}, respectively.

\begin{table*}[t]
\centering
\small
\caption{Training results. \textbf{Bold}/\underline{underlining} mark the best/second-best values, including ties.}
\label{tab:training-core-results}
\setlength{\tabcolsep}{2pt}
\setlength{\MetricWidth}{24pt}
\renewcommand{\arraystretch}{1.15}
\makebox[\linewidth][c]{\resizebox{\linewidth}{!}{%
\begin{tabular}{lI*{4}{E}I*{4}{E}I*{4}{E}}
\toprule
 & \multicolumn{4}{Ic}{\makecell[c]{\textit{Qwen3-VL-8B}\\$\downarrow$\\\textit{Qwen3-VL-8B}}} & \multicolumn{4}{Ic}{\makecell[c]{\textit{Qwen3-VL-32B}\\$\downarrow$\\\textit{Qwen3-VL-32B}}} & \multicolumn{4}{Ic}{\makecell[c]{\textit{Qwen3-VL-32B}\\$\downarrow$\\\textit{Qwen3-VL-8B}}} \\
\cmidrule(lr){2-5} \cmidrule(lr){6-9} \cmidrule(lr){10-13}
Method & Sok. & Cra. & Proc. & Mini. & Sok. & Cra. & Proc. & Mini. & Sok. & Cra. & Proc. & Mini. \\
\midrule
ReAct SFT & 44.4 & 4.24 & 4.50 & 8.8 & 60.1 & 4.63 & 6.88 & 18.5 & --- & --- & --- & --- \\
Plan-Act SFT & 35.6 & 4.06 & 3.50 & 7.8 & 50.7 & 3.10 & 6.35 & 15.8 & 45.4 & 2.85 & 5.60 & 11.9 \\
PAA & 43.8 & 4.33 & 3.15 & 10.8 & 65.4 & 3.76 & \underline{8.27} & 11.6 & \underline{49.4} & 3.40 & \underline{5.89} & 8.7 \\
HSL & 44.4 & 4.39 & 4.82 & 8.0 & 61.3 & 4.23 & 5.85 & 16.5 & --- & --- & --- & --- \\
WebSTAR & 44.7 & 4.28 & \underline{5.41} & 8.7 & 66.2 & 4.96 & 7.26 & 13.1 & --- & --- & --- & --- \\
ECoT & 34.4 & \underline{4.77} & 2.56 & 10.2 & 57.2 & \underline{5.23} & 5.25 & 29.1 & 41.2 & 4.70 & 2.62 & 12.7 \\
\OursRow \textbf{ConPAct-S} & \underline{48.6} & \textbf{5.16} & 3.22 & \textbf{27.0} & \textbf{68.4} & 5.05 & \textbf{9.12} & \textbf{39.7} & 48.5 & \textbf{5.53} & 4.56 & \textbf{28.2} \\
\OursRow \textbf{ConPAct-R} & \textbf{49.1} & 4.72 & \textbf{6.38} & \underline{24.8} & \underline{66.9} & \textbf{5.55} & 6.35 & \underline{33.9} & \textbf{54.8} & \underline{5.26} & \textbf{7.36} & \underline{27.8} \\
\bottomrule
\end{tabular}}}
\end{table*}

\subsection{Main Results}

The main tables report Sokoban avg@8, Crafter Score, Procgen IQM, and MiniGrid SR (all $\uparrow$). Complete inference and training results are provided in Appendix Tables~\ref{tab:motivation-inference-results} and~\ref{tab:motivation-training-results}.

\textbf{ConPAct-I improves task performance across model families.}
Table~\ref{tab:inference-core-results} shows that ConPAct-I improves Sokoban avg@8 and Procgen IQM over Plan-Act for all four model pairs.
Sokoban gains range from 3.7 to 11.5 percentage points.
With GPT-5.6-sol and GPT-5.6-terra, MiniGrid SR increases from 38.6\% to 54.4\%, exceeding both HiPlan and TAPE.
These results support using detected state contradictions to improve plan-act coordination without training.

\textbf{Both sampling and replay provide effective coordination supervision.}
In Table~\ref{tab:training-core-results}, ConPAct-S and ConPAct-R outperform both ReAct SFT and Plan-Act SFT on Sokoban avg@8 and MiniGrid SR at both model sizes.
For Qwen3-VL-32B, ConPAct-S achieves 39.7\% MiniGrid SR, compared with 29.1\% for the strongest baseline.
The two data routes also yield leading results on the other environments: ConPAct-S attains the highest Procgen IQM (9.12), while ConPAct-R attains the highest Crafter Score (5.55) among the 32B methods.

\textbf{The gains extend to heterogeneous model sizes.}
With a 32B planner and an 8B actor (Table~\ref{tab:training-core-results}, rightmost column group), both variants improve Sokoban avg@8, Crafter Score, and MiniGrid SR over Plan-Act SFT.
MiniGrid SR rises from 11.9\% to 28.2\% with ConPAct-S and 27.8\% with ConPAct-R.
Thus, consistency-oriented supervision also benefits coordination between models of different sizes.

\textbf{The improvements extend to desktop tasks.}
On 361 OSWorld tasks, ConPAct-I outperforms ReAct and Plan-Act in both overall Score and SR across all four evaluated models (Figure~\ref{fig:osworld_gains}).
Relative to the stronger baseline for each model and metric, Score improves by 0.7--0.9 percentage points and SR by 0.8--1.0 points.
These results extend the gains from game environments to desktop interaction.
Appendix Table~\ref{tab:osworld-full} reports the complete category-level results.

\subsection{Ablation Study}

\begin{figure}[t]
  \centering
  \begin{subfigure}[t]{0.33\textwidth}
    \centering
    \includegraphics[width=\textwidth]{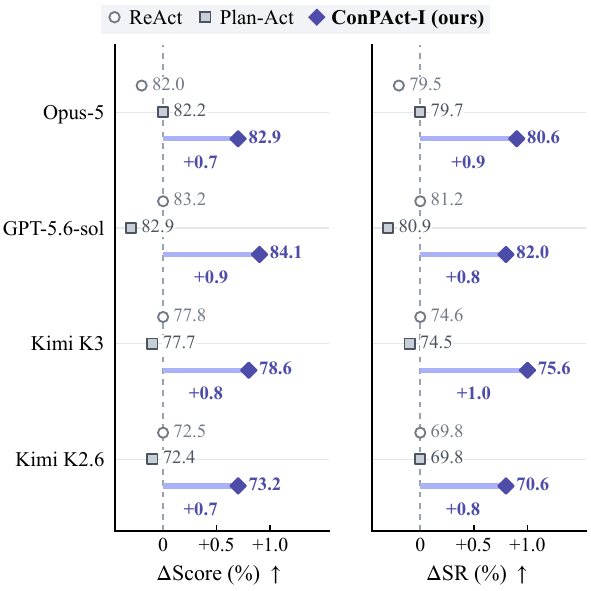}
    \caption{OSWorld.}
    \label{fig:osworld_gains}
  \end{subfigure}\hfill
  \begin{subfigure}[t]{0.33\textwidth}
    \centering
    \includegraphics[width=\textwidth]{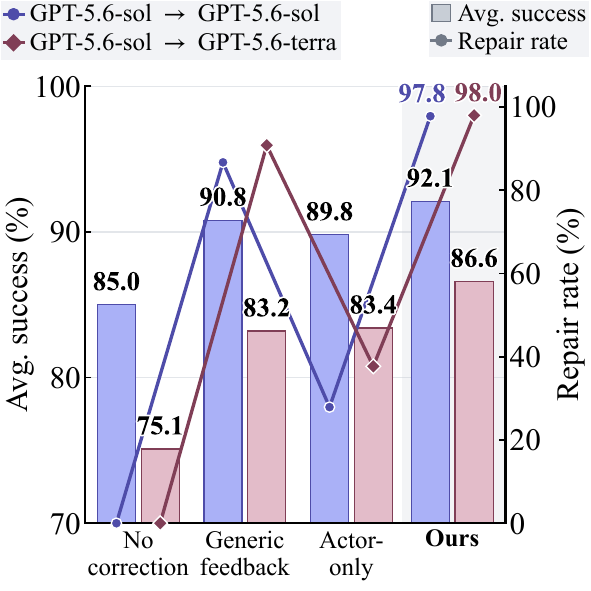}
    \caption{Inference ablation.}
    \label{fig:sokoban_inference_ablation}
  \end{subfigure}\hfill
  \begin{subfigure}[t]{0.33\textwidth}
    \centering
    \includegraphics[width=\textwidth]{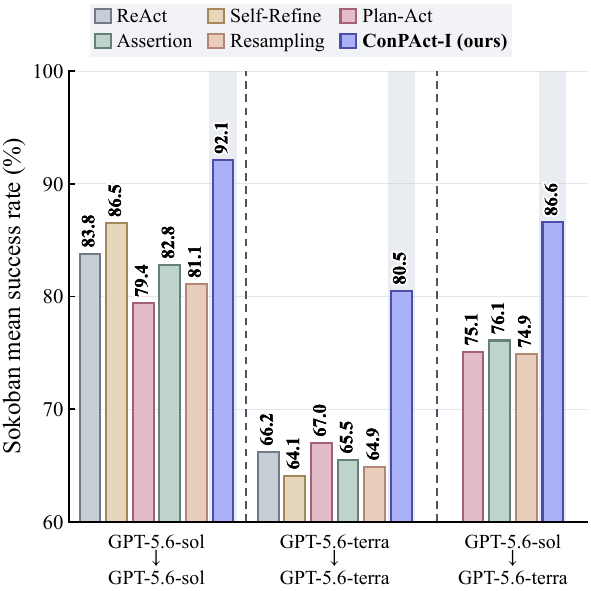}
    \caption{Method comparison.}
    \label{fig:sokoban_method_comparison}
  \end{subfigure}
  \caption{OSWorld performance and Sokoban inference-time comparisons. (a) OSWorld Score and SR relative to the stronger of ReAct and Plan-Act within each model, measured in percentage points; point labels give absolute percentages and purple annotations give ConPAct-I gains. (b) Task success (bars, left axis) and disagreement repair (lines, right axis) under different correction settings. (c) Task success with assertion prompting, resampling, Self-Refine, and ConPAct-I.}
  \label{fig:sokoban_eval_summary}
\end{figure}

\textbf{Conflict feedback and planner participation improve reconciliation.}
We compare the full intervention with correction disabled, correction without specific conflict feedback, and correction restricted to the actor.
The full method achieves the highest task success and repair rates for both GPT configurations on Sokoban and MiniGrid (Figure~\ref{fig:sokoban_inference_ablation}).
For the heterogeneous pair on Sokoban, repair reaches 98.0\%, compared with 90.8\% without specific feedback and 37.7\% with actor-only correction.
These comparisons support involving both roles and identifying their disagreements.

\textbf{Reconciliation improves on assertion prompting and resampling.}
Figure~\ref{fig:sokoban_method_comparison} compares ConPAct-I with assertion-only prompting, resampling, and Self-Refine~\citep{madaan2023selfrefine}.
ConPAct-I achieves higher Sokoban success than assertion prompting and resampling in all three configurations, and higher success than Self-Refine in both homogeneous configurations.
The corresponding MiniGrid results show the same ordering.

\textbf{Supervising and replaying reconciliation improve training.}
We compare full ConPAct-R with a baseline omitting all three data-curation components and two variants that retain consistency filtering but remove reconciliation supervision or reconciliation replay.
Full ConPAct-R achieves higher task success and lower mismatch than all three alternatives across the evaluated model configurations on Sokoban and MiniGrid (Figure~\ref{fig:sokoban_sft_ablation}).
For 32B on Sokoban, it improves avg@8 from 55.9\% to 66.9\% and reduces mismatch from 34.7\% to 1.4\% relative to the baseline.

Complete Sokoban and MiniGrid results for the inference and training-data ablations are reported in Appendices~\ref{app:inference-ablations} and~\ref{app:data-ablations}, respectively.

\subsection{Analysis of State Understanding and Coordination}

We examine ConPAct-R through initial state judgments, discussion responses, and the correctness of shared interpretations.
Complete results for Sokoban and MiniGrid are reported in Appendix~\ref{app:training-analysis}.

\begin{figure}[t]
  \centering
  \begin{subfigure}[t]{0.33\textwidth}
    \centering
    \includegraphics[width=\textwidth]{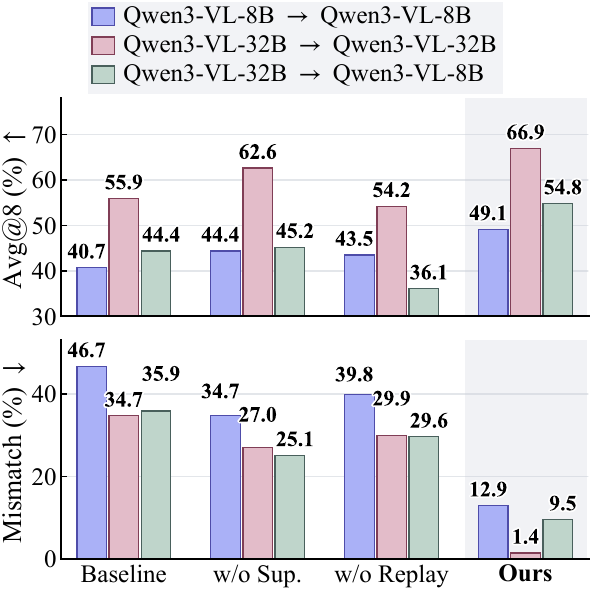}
    \caption{SFT data ablation.}
    \label{fig:sokoban_sft_ablation}
  \end{subfigure}\hfill
  \begin{subfigure}[t]{0.33\textwidth}
    \centering
    \includegraphics[width=\textwidth]{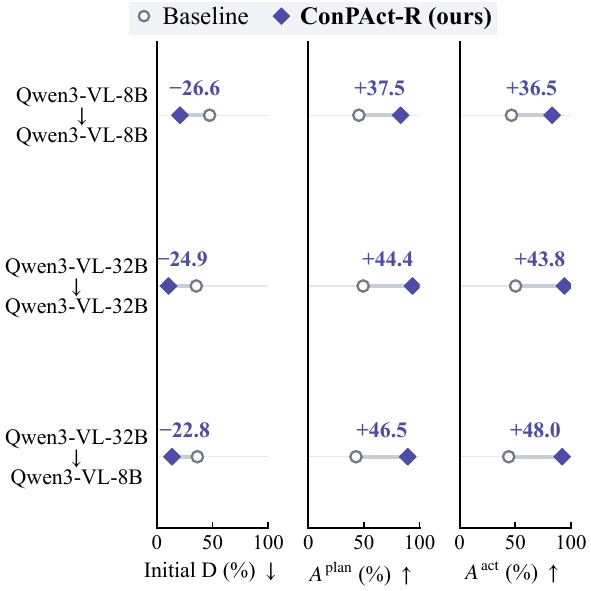}
    \caption{Initial state understanding.}
    \label{fig:sokoban_posthoc_state}
  \end{subfigure}\hfill
  \begin{subfigure}[t]{0.33\textwidth}
    \centering
    \includegraphics[width=\textwidth]{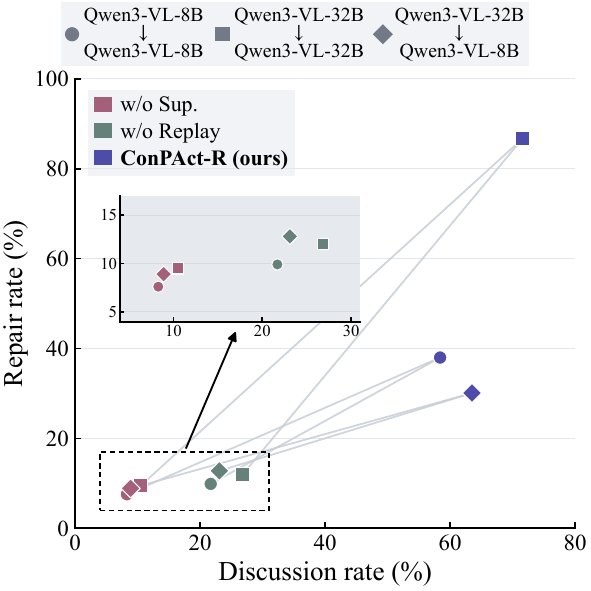}
    \caption{Discussion and repair.}
    \label{fig:sokoban_discussion_repair}
  \end{subfigure}
  \caption{Supervised-training ablations and state analysis on Sokoban. (a) Task success and mismatch under ConPAct-R data ablations; Full includes consistency filtering, reconciliation supervision, and reconciliation replay. (b) Initial mismatch and assertion correctness; annotations show changes in percentage points from the baseline. (c) Discussion and repair rates for ConPAct-R and its data ablations. The components together contribute to the effectiveness of ConPAct.}
  \label{fig:sokoban_analysis}
\end{figure}

\textbf{Training improves initial state understanding.}
Across the three Sokoban configurations, ConPAct-R reduces initial mismatch by 22.8--26.6 percentage points and improves each role's assertion correctness by 36.5--48.0 points (Figure~\ref{fig:sokoban_posthoc_state}).
These improvements precede discussion, indicating that the trained models begin coordination with more accurate and compatible state judgments.
MiniGrid exhibits the same pattern.

\textbf{Training improves discussion responses and disagreement repair.}
Full ConPAct-R yields higher discussion rates and lower residual mismatch than either data ablation across all three model configurations in both environments (Figure~\ref{fig:sokoban_discussion_repair}).
For 32B on Sokoban, the discussion rate reaches 71.6\%, while residual mismatch falls to 13.3\%.
This supports supervising corrective interactions to improve how agents respond to disagreements.

\textbf{Reconciliation increases correct shared interpretations.}
To distinguish correct agreement from mere consistency, we examine whether the two roles' shared assertions also match the environment state.
Full ConPAct-R has a higher proportion of consistent and correct assertions than either ablation, both before and after discussion, across all three Sokoban configurations.
Discussion further increases this proportion by 2.7--5.1 percentage points, reaching 77.7--93.5\% (Appendix Table~\ref{tab:posthoc-state-distribution}).
Correct agreement also increases in MiniGrid, supporting the role of reconciliation in improving shared state understanding.

\section{Related Work}

\subsection{Plan-Act Agents}

Plan-act agents separate planning from action selection~\citep{erdogan2025planandact}.
Existing methods improve these capabilities through global guidance~\citep{li2026hiplan}, constrained execution~\citep{jeong2026tape}, and hierarchical policy optimization~\citep{hu2025divide,peng2026hiper}.
ReCAPA and ContextFlow address cross-level correction and task-stage coordination, respectively~\citep{zeng2026recapa,guo2026contextflow}, while Agent GPA evaluates plan adherence~\citep{jia2025what}.
Training also exploits trajectory relabeling~\citep{li2026spinning} and multi-agent correction or role specialization~\citep{zhao2025sirius,motwani2025malt,subramaniam2025multiagentfinetuning}.

ConPAct identifies a distinct failure: incompatible interpretations of the same observed facts despite plan adherence.
It trains on consistent collaboration and correction segments, retaining earlier conflicts as context for learning reconciliation.
Its replay variant reconstructs these interactions without changing recorded actions, enabling existing trajectories to support coordination training.

\subsection{State Tracking and Verification in LLM Agents}

ECoT and VAGEN train explicit state reasoning~\citep{zawalski2024robotic,wang2025vagen}, while CoEx and EVU maintain and update beliefs~\citep{kim2025coex,wang2026evu}.
Learned transition models further support planning and policy learning~\citep{luo2026vimo,ding2026dynaweb}.
Self-Refine, multiagent debate, and Reflexion guide revision with model-generated feedback~\citep{madaan2023selfrefine,du2024multiagentdebate,shinn2023reflexion}, although self-verification remains limited~\citep{stechly2025on}.
Other methods score intermediate execution~\citep{chae2025webshepherd,chen2025guishepherd}, detect plan-constraint violations~\citep{guo2024doremi}, or validate outputs against planner-defined criteria~\citep{xu2026verificationaware}.

ConPAct programmatically compares both roles' state reports about the same observation and returns conflicting fields for joint revision.
The comparison assumes neither role is correct and requires neither environment ground truth nor an additional verifier-model call.
This targeted feedback improves state interpretation and coordination by allowing both roles to revise their judgments.

\section{Conclusion}

We identify and study planner-actor state mismatch as a coordination failure in which agents interpret task-relevant states differently, even when the actor follows its assigned sub-goal.
Our findings show that making these interpretations explicit and reconciling disagreements can improve coordination through both inference-time interaction and training. The online comparison and reconciliation require neither environment ground truth nor an external verifier.
Although consistency alone does not ensure correctness, these results support treating shared state understanding as an explicit objective in designing and training collaborative agents.
Future work can examine how this approach scales to larger agent teams and more complex environments, where differing observations and interdependent decisions create further challenges for coordination.

\clearpage
\bibliography{references}
\bibliographystyle{plainnat}

\clearpage
\appendix
\startcontents[appendix]
\begingroup
\setlength{\parindent}{0pt}
\hypersetup{linktoc=all, pdfborder={0 0 0}}
\pdfbookmark[0]{Appendix Contents}{app:contents}
{\LARGE\bfseries Appendix Contents\par}
\vspace{12pt}
\titlecontents{appsection}[1.8em]
  {\addvspace{8pt}\normalsize\bfseries}
  {\contentslabel{1.8em}}{}
  {\hfill\contentspage}
\titlecontents{appsubsection}[4.2em]
  {\addvspace{2pt}\normalsize}
  {\contentslabel{2.4em}}{}
  {{\color{black!35}\titlerule*[0.6em]{.}}\contentspage}
\printcontents[appendix]{app}{1}[2]{}
\endgroup
\clearpage

\FloatBarrier
\section{Additional Related Work}
\label{app:add_related}

\subsection{Plan-Act Agents}
Plan-act systems separate sub-goal selection from action execution, with prior work addressing global guidance, constrained execution, and hierarchical learning~\citep{erdogan2025planandact,li2026hiplan,jeong2026tape,hu2025divide}.
Trajectory relabeling and multi-agent training also turn past interactions into supervision~\citep{li2026spinning,zhao2025sirius,motwani2025malt,subramaniam2025multiagentfinetuning}.
ConPAct focuses on whether the two roles interpret the same observed state compatibly, and uses their explicit contradictions to guide correction and training.

\subsection{State Tracking and Verification in LLM Agents}
State-grounded reasoning and belief updates support agent decisions~\citep{zawalski2024robotic,wang2025vagen,kim2025coex,wang2026evu,wang2026castgamesolversturnlevel}.
Other methods guide revision with critiques, process scores, or checks against planning criteria~\citep{madaan2023selfrefine,chae2025webshepherd,guo2024doremi,xu2026verificationaware}.
ConPAct-I instead compares the roles' reports directly: neither report is treated as ground truth, and detecting a contradiction requires no additional verifier-model call.
The teacher review used to curate ConPAct-R targets is a separate offline training-data operation.

\subsection{Common Ground and Belief Alignment}
Common ground concerns the information needed for coordinated interaction~\citep{clark1991grounding}.
Recent work examines inter-agent failures, belief-sensitive dialogue, and shared understanding~\citep{cemri2025mast,qiu2024minddial,chen2026collabsim,dongre2026embodied}.
Our setting isolates disagreements about the same task-relevant observed facts at planner--actor handoffs and evaluates both their reconciliation and the correctness of the resulting agreement.

\FloatBarrier
\section{State Assertions}
\label{app:add_pre}

\subsection{Plan-Act Protocol}
\label{app:plan-act-protocol}

\textbf{Roles and observations.}
The planner selects a sub-goal; the actor chooses actions and reports execution status.
Let $t$ index model calls and $k$ index executed environment actions.
Several calls may share observation $o^k$; only an executed action advances the environment to $o^{k+1}$.
An execution segment consists of actor turns pursuing one adopted sub-goal.
Actor feedback, including completion or execution difficulties, triggers replanning.

\textbf{Plan-Act baseline.}
The planner returns a \texttt{<subgoal>}; the actor returns an \texttt{<action>} and a \texttt{<status>}.
Its statuses are \texttt{CONTINUE}, \texttt{ACHIEVED}, \texttt{IMPOSSIBLE}, \texttt{STUCK}, and \texttt{CHANGED}.
The actor's proposed action executes before a non-\texttt{CONTINUE} status returns control to the planner.
Consequently, passive assertion comparison uses the planner-to-actor handoff before that action, when both roles have the same observation; it never pairs reports across an environment transition.
The ordinary interface passes sub-goals and status feedback rather than full reasoning traces.

\textbf{ConPAct interface.}
A planner reply contains \texttt{<assertion>}, \texttt{<message>}, and \texttt{<subgoal>}; an actor reply additionally supplies \texttt{<action>} and \texttt{<status>}.
The actor uses \texttt{CONTINUE} for an executable action, \texttt{DISCUSS} to request reconsideration, and \texttt{ACHIEVED} when sub-goal completion is already visible.
Both \texttt{DISCUSS} and \texttt{ACHIEVED} require action \texttt{NONE} and leave the observation unchanged.
The planner then retains or revises its sub-goal using the actor's feedback.
ConPAct therefore supports same-observation comparisons in both handoff directions.
Overall task completion is determined by the environment, not by the actor's completion claim.

\textbf{Independent measurement and reconciliation.}
The receiving role does not see the other role's assertion when the initial pair is elicited; ordinary sub-goals and feedback remain available.
After an initial contradiction is detected, reconciliation exposes the disputed fields and the roles' reports for reassessment.
All actions remain paused during correction.
ConPAct-I returns the last actor reply when correction ends and does not insert an additional actor call solely to choose a replacement action.
The ordinary controller applies its action/status validity rules to this reply; \texttt{NONE} does not execute an environment action.
Both roles retain up to three observation frames, with method-specific frame selection and text history described in Appendix~\ref{app:training-evaluation}.

\FloatBarrier
\subsection{Assertion Format}
\label{app:state-assertions}
\label{app:state_asssertion}

\textbf{Scope and representation.}
A state assertion is a JSON object enclosed in \texttt{<assertion>} and \texttt{</assertion>} tags, reported alongside the role's planning or execution response.
Both roles use the same environment-specific schema and describe the \emph{current observation}, before the proposed action is executed.
The fields capture visible conditions relevant to interpreting sub-goals: object counts and local obstacles in grid tasks, interaction prerequisites and resources in Crafter, visible targets and threats in Procgen, and application context and task evidence in OSWorld.
They are compact reports of state interpretation, not complete scene descriptions or predictions of action outcomes.
Research on chain-of-thought explanations shows that verbal reports need not faithfully expose a model's reasoning~\citep{turpin2023language}.
We therefore interpret state assertions as observable proxies and do not assume that they reveal the models' internal beliefs.
No schema requires persistent object identifiers, absolute world coordinates, or access to hidden simulator state.
Task instructions and current observation metadata may disambiguate what to inspect, but unavailable facts should not be filled in from earlier frames.
Prompt templates are provided in Appendix~\ref{app:prompts}.
Categorical domains are listed below. Count fields are nonnegative integers unless a narrower range is explicitly stated; no additional global upper bound is imposed.

\textbf{Sokoban.}
The four fields describe puzzle progress and the player's immediate surroundings.
\texttt{box\_count} counts all boxes, including boxes on targets; \texttt{target\_count} counts all target locations, including occupied targets; and \texttt{on\_goal\_count} counts boxes currently on targets.
All three are nonnegative integers.
\texttt{player\_neighbors} is a four-element array in the fixed screen-relative order \emph{up, right, down, left}, with values \texttt{cell}, \texttt{wall}, \texttt{box}, \texttt{target}, or \texttt{box\_on\_target}.
Here \texttt{cell} means empty floor, and \texttt{target} means a target without a box.
These fields are read from the full board and distinguish completion conditions from local movement or pushing prerequisites; they do not encode whether an entire push sequence is feasible.

\textbf{Crafter.}
The five fields summarize the local interaction target and the heads-up display (HUD).
\texttt{facing} is one of \texttt{up}, \texttt{right}, \texttt{down}, or \texttt{left} in screen coordinates.
\texttt{front} identifies the single cell in that direction: \texttt{walkable}, \texttt{water}, \texttt{lava}, \texttt{stone}, \texttt{tree}, \texttt{coal}, \texttt{iron}, \texttt{diamond}, \texttt{table}, \texttt{furnace}, \texttt{cow}, \texttt{zombie}, \texttt{skeleton}, \texttt{plant}, or \texttt{arrow}.
\texttt{walkable} merges empty grass, sand, and path; \texttt{plant} covers both growth stages; an \texttt{arrow} must occupy that cell.
The moving camera does not establish fixed world coordinates.
\texttt{vitals} is the ordered array [health, food, drink, energy].
\texttt{inventory} has exactly the keys \texttt{wood}, \texttt{stone}, \texttt{coal}, and \texttt{iron}.
\texttt{tools} has exactly six keys: \texttt{wood\_pickaxe}, \texttt{stone\_pickaxe}, \texttt{iron\_pickaxe}, \texttt{wood\_sword}, \texttt{stone\_sword}, and \texttt{iron\_sword}.
Every vital, material count, and tool count is an integer in $\{0,\ldots,9\}$; a material or tool absent from the HUD is reported as zero.
The schema covers resource and tool prerequisites and immediate survival needs, while omitting other facts such as the full arrangement of crafting stations.

\textbf{MiniGrid / BabyAI.}
The five fields describe visible keys, door states, and the next cell relative to the agent's orientation.
\texttt{key\_count} counts keys drawn on the board; a carried key is not drawn and is excluded.
\texttt{front\_cell} is the cell directly ahead of the agent triangle, using the direction supplied with the current screenshot.
Its values are \texttt{empty}, \texttt{wall}, \texttt{door\_locked}, \texttt{door\_closed}, \texttt{door\_open}, \texttt{key}, \texttt{ball}, \texttt{box}, \texttt{goal}, or \texttt{lava}; the outer boundary is \texttt{wall}.
\texttt{locked\_door\_count}, \texttt{closed\_door\_count}, and \texttt{open\_door\_count} count locked, closed-but-unlocked, and open doors, respectively.
Their visual cues are a filled cell with a keyhole, a hollow double-edged square, and a thin strip along a cell edge.
All four counts are nonnegative integers and cover both light and dark rendered cells in the full-board image, not only the highlighted field of view.
These fields describe visible movement and interaction conditions; zero visible keys does not establish that the agent is carrying a key.

\textbf{Procgen.}
The schema is selected by game identity.
Every game has \texttt{target\_visible}; every game except maze also has \texttt{hazard\_nearby}.
Both take \texttt{yes}, \texttt{no}, or \texttt{unknown}, using the referents in Table~\ref{tab:procgen-assertion-referents}.
A hazard is nearby when it is visibly close enough to threaten the player or the next immediate movement; the protocol does not impose a pixel-distance threshold.
Miner additionally has \texttt{diamonds\_visible}, a nonnegative integer counting only diamonds in the screenshot, or JSON \texttt{null} when a reliable count is unavailable.
Bigfish additionally has \texttt{my\_size}: \texttt{smallest} when all other visible fish are larger, \texttt{middle} when both larger and smaller fish are visible, \texttt{biggest} when none is larger, and \texttt{unknown} when comparison is unclear or no other fish is visible.
\texttt{no} requires sufficient visible evidence of absence; an obscured or ambiguous view warrants \texttt{unknown}.
An off-screen target may still exist, and the absence of a visible threat does not prove that a route is safe.

\begin{table}[htbp]
\centering
\small
\caption{Game-specific referents for Procgen assertions. Maze has no \texttt{hazard\_nearby} field.}
\label{tab:procgen-assertion-referents}
\setlength{\tabcolsep}{5pt}
\renewcommand{\arraystretch}{1.15}
\begin{tabular}{@{}p{0.12\linewidth}p{0.37\linewidth}p{0.43\linewidth}@{}}
\toprule
Game & Target for \texttt{target\_visible} & Threat for \texttt{hazard\_nearby} \\
\midrule
bigfish & At least one visibly smaller fish & A larger fish \\
chaser & At least one uncollected green orb & A non-vulnerable enemy \\
coinrun & The goal coin & A saw, enemy, or deadly gap \\
jumper & The carrot & Spikes \\
maze & The cheese & Not included in the schema \\
miner & At least one diamond & A boulder or diamond that could fall onto the player \\
ninja & The goal mushroom & A bomb or dangerous gap \\
\bottomrule
\end{tabular}
\end{table}

\textbf{OSWorld.}
The four fields refer to the current desktop screenshot and the \emph{original task}, rather than the planner's temporary sub-goal.
\texttt{active\_app} identifies the foreground application: \texttt{chrome}, \texttt{writer}, \texttt{calc}, \texttt{impress}, \texttt{gimp}, \texttt{vscode}, \texttt{thunderbird}, \texttt{vlc}, \texttt{files}, \texttt{terminal}, \texttt{settings}, \texttt{desktop}, \texttt{other}, or \texttt{unknown}.
An application-owned dialog is attributed to its owner when identifiable; \texttt{other} denotes an identifiable application outside the list.
\texttt{overlay} takes \texttt{none}, \texttt{menu}, \texttt{dialog}, or \texttt{unknown}, with a visible dialog taking precedence over a menu.
A file picker is a dialog, not the \texttt{files} application.
\texttt{target\_visible} takes \texttt{yes}, \texttt{partial}, \texttt{no}, or \texttt{unknown}: all, some, none, or an indeterminate amount of the task's target objects or regions are visible.
\texttt{goal\_state} takes \texttt{met}, \texttt{unmet}, or \texttt{unknown}.
\texttt{unmet} requires a visibly unsatisfied requirement; \texttt{met} requires the screenshot alone to establish every requirement; otherwise the value is \texttt{unknown}.
This visual judgment is distinct from the official task score: a filled input or closed dialog does not prove that a file was saved or that off-screen content is correct.

\textbf{Extraction, normalization, and validity.}
The parser extracts the tagged JSON object and checks field types, required keys, array lengths, and the domains above.
Integer fields reject booleans, floating-point values, and numeric strings; they are not coerced or clipped.
Crafter, MiniGrid, Procgen, and OSWorld require exactly one assertion block, reject duplicate JSON keys at any nesting level, and require exactly the specified fields, including Crafter's nested keys.
Their categorical values must match the listed strings exactly, and Markdown fences inside the assertion block are not accepted.
The Sokoban schema extractor instead uses the last assertion block, removes optional surrounding JSON fences, retains only its four required fields, and trims and lowercases neighbor labels; thus the prompt spelling \texttt{Box\_on\_target} becomes \texttt{box\_on\_target}.
Its JSON decoder uses the last occurrence of a duplicate key.
ConPAct additionally validates the order and uniqueness of the complete reply's protocol tags, so this schema-level tolerance does not permit repeated assertion blocks in a ConPAct reply.
Missing fields or invalid values yield an invalid assertion, not an all-zero or unknown-filled replacement; in the online runners, a required assertion that fails parsing causes a model-output failure.
Schema validity only establishes structural admissibility: for example, the Sokoban parser does not enforce cross-field constraints between the three counts, and it does not verify the report against the board.

\textbf{Definite values and uncertainty.}
Only Procgen and OSWorld provide unknown sentinels: the literal string \texttt{unknown} for their categorical fields, and \texttt{null} for Procgen miner's \texttt{diamonds\_visible}.
Sokoban, Crafter, and MiniGrid require definite values in every field.
Zero is a definite count, and OSWorld's \texttt{none}, \texttt{other}, and \texttt{partial} are definite categories, not abstentions.
Comparison uses parsed values rather than raw JSON text: object-key order is irrelevant, array order matters, and each top-level slot is compared as a whole, including Crafter's arrays and nested objects.
For a valid pair, a slot is comparable only if neither role reports its declared unknown sentinel.
The result is \texttt{CONFLICT} if any comparable slot differs, \texttt{AGREE} if every slot is definite and matches, and \texttt{INDETERMINATE} otherwise.
Thus, matching unknowns do not establish agreement, while a definite contradiction still counts as a conflict even if another slot is unknown.
These comparisons require no ground truth; the code does not provide a reliable assertion-correctness oracle for Procgen or OSWorld, and task scores are not substituted for one.

\textbf{Pairing and measurement stages.}
Assertions are paired by environment step, using the handoff protocol above.
An initial pair is elicited without exposing the other role's assertion; later reconciliation pairs can use the shared discussion context.
Reports across an executed action are never paired.
Passive mismatch measurement does not return discrepancy feedback or insert corrective calls.
The training analyses distinguish initial judgments from the reports available after model-initiated discussion (Appendix~\ref{app:training-analysis}).

\FloatBarrier
\subsection{Metrics}
\label{app:metrics}

Let $\{\tau_i\}_{i=1}^{N}$ denote the evaluated trajectories for one model--environment configuration.
Repeated rollouts are separate trajectories.
Infrastructure failures are excluded; model-output failures and budget exhaustion remain included.
The following rates are computed over their specified eligible samples and displayed as percentages. The mismatch analyses combine the 200 Sokoban and 150 MiniGrid levels before applying these metric-specific denominators; the main benchmark tables use the environment-specific aggregation rules below.

\textbf{State mismatch.}
Let $\mathcal P$ contain the evaluated handoff pairs $(i,t)$ for which both assertions are valid and every field is definite. Passive measurement uses the independent initial pairs; inference corrections and training analyses explicitly identify their initial or post-discussion event sets.
Each pair describes the same observation, $o_{i,t}=o_{i,t+1}$.
Using $d_{i,t}$ from Eq.~\ref{eq:state_mismatch}, we define
\begin{equation}
D=\frac{1}{|\mathcal P|}\sum_{(i,t)\in\mathcal P}d_{i,t}.
\label{eq:metric-mismatch}
\end{equation}
Each eligible pair has equal weight, regardless of its trajectory length or outcome. Discussion-conditional and paired pre/post statistics use the event sets specified in Appendix~\ref{app:training-analysis}.

\textbf{State correctness.}
Within pair $(i,t)$, let $\mathbf v^{\text{plan}}_{i,t}$ and $\mathbf v^{\text{act}}_{i,t}$ denote the two assertions, indexed by role.
Let $\mathbf v^*_{i,t}$ be the ground-truth assertion for their shared observation.
Define $\mathcal G\subseteq\mathcal P$ as the pairs with valid, fully definite ground truth.
For $r\in\{\text{plan},\text{act}\}$, exact-match accuracy is
\begin{equation}
A^r=\frac{1}{|\mathcal G|}\sum_{(i,t)\in\mathcal G}
\mathbf 1\!\left[\forall f\in\mathcal F:\mathbf v^r_{i,t}[f]=\mathbf v^*_{i,t}[f]\right].
\label{eq:metric-correctness}
\end{equation}
All fields must match; there is no partial credit.
Both roles use the same denominator.
Sokoban ground truth is obtained by replaying executed actions from the initial board.
MiniGrid ground truth is extracted from the recorded initial state and subsequent environment states, aligned with the current observation.
Keys and doors are counted on the board; the front cell is determined by the agent's position and direction.
Carried keys are excluded.
Ground truth follows the schemas in Appendix~\ref{app:state-assertions} and is hidden from both roles during passive measurement.

\textbf{Sub-goal adherence.}
Let $q_{i,j}$ be the judge's verdict for execution segment $j$ of $\tau_i$.
Let $\mathcal J=\{(i,j):q_{i,j}\in\{\texttt{FOLLOW},\texttt{VIOLATE}\}\}$.
We define
\begin{equation}
H=\frac{1}{|\mathcal J|}\sum_{(i,j)\in\mathcal J}
\mathbf 1[q_{i,j}=\texttt{FOLLOW}].
\label{eq:metric-adherence}
\end{equation}
Each segment receives one verdict and equal weight.
We use GPT-5.6-sol~\citep{openai2026gpt56} as an LLM judge~\citep{zheng2023mtbench}.
Our rubric targets plan--action adherence, an aspect of agent alignment examined by Agent GPA~\citep{jia2025what}.
Its input contains the assigned sub-goal and, for each segment step, the ground-truth state as text followed by the actor's complete response.
It receives no screenshots, planner reasoning, or other segments.
\texttt{FOLLOW} means that the actions pursue the sub-goal, even if the actor misreads the state or fails to complete it.
\texttt{VIOLATE} means that the actions contradict the sub-goal or pursue another goal.
\texttt{UNKNOWN} is reserved for sub-goals with no executable content; it is not used for close decisions.
Segments truncated by termination, budget exhaustion, or model-output failure are judged from the available trace.
Adherence does not require a valid assertion pair.

\textbf{Task success.}
A trajectory succeeds when the environment confirms task completion within the rollout budget.
We define
\begin{equation}
S=\frac{1}{N}\sum_{i=1}^{N}\mathbf 1[\tau_i\text{ succeeds}].
\label{eq:metric-success}
\end{equation}
An actor's completion claim alone does not establish success.
This trajectory-level aggregation is used for the state analyses; environment-specific benchmark metrics are defined below.

\textbf{Exclusions and empty denominators.}
Invalid assertions and pairs with any unknown field are excluded from $D$ and both correctness rates.
A definite conflict in a partially unknown pair is recorded separately and does not enter $D$.
Missing ground truth excludes a pair from correctness only.
Missing sub-goals, \texttt{UNKNOWN} verdicts, and missing or malformed judge responses are excluded from $H$.
These exclusions do not remove a trajectory from $S$.
Eligible counts and exclusion counts are recorded separately.
An empty denominator yields an undefined metric, not zero.
Trajectory-level $D_i$ and $H_i$ use the same definitions restricted to $\tau_i$.

\textbf{Sokoban.}
For each of the $L=200$ puzzles, let $c_\ell\in\{0,\ldots,8\}$ be the number of successful rollouts.
A rollout succeeds when every box is on a target.
We report
\begin{equation}
\mathrm{avg}@8=\frac{1}{L}\sum_{\ell=1}^{L}\frac{c_\ell}{8},\qquad
\mathrm{pass}@8=\frac{1}{L}\sum_{\ell=1}^{L}\mathbf{1}[c_\ell\geq1],\qquad
\mathrm{pass}^{8}=\frac{1}{L}\sum_{\ell=1}^{L}\mathbf{1}[c_\ell=8].
\end{equation}
All three are displayed as percentages.
The latter two specialize the pass@$k$ criterion~\citep{chen2021codex} and the pass$^k$ reliability criterion~\citep{yao2025taubench} to eight attempts: at least one success and success on all eight attempts, respectively.

\textbf{Crafter.}
Let $a_{j,i}$ indicate whether episode $j$ unlocks achievement $i$, with $i\in\{1,\ldots,22\}$, and let $N$ denote the number of evaluated episodes.
The achievement success rate is $s_i=100N^{-1}\sum_j a_{j,i}$.
We compute
\begin{equation}
\mathrm{Score}=\exp\left(\frac{1}{22}\sum_{i=1}^{22}\log(1+s_i)\right)-1,\qquad
\mathrm{Progress}=\frac{100}{22N}\sum_{j=1}^{N}\sum_{i=1}^{22}a_{j,i}.
\end{equation}
Score follows the Crafter benchmark~\citep{hafner2022crafter} and is computed from the achievement success rates over the full evaluation set.
Progress is the mean proportion of unlocked achievements, following BALROG~\citep{paglieri2025balrog}.
Return is the mean undiscounted sum of native rewards per episode, including rewards associated with changes in health.

\textbf{Procgen.}
For game $g$, we average the undiscounted episode returns over its 20 levels and eight rollouts per level, obtaining $\bar G_g$.
The normalized return is
\begin{equation}
z_g=\frac{\bar G_g-R_{\min,g}}{R_{\max,g}-R_{\min,g}},
\end{equation}
using the easy-mode reference bounds from \citet{cobbe2020procgen}, listed in Table~\ref{tab:procgen-environment-details}.
Returns include rewards up to environment termination or budget exhaustion, and normalized values are left unclipped.
We use the \texttt{rliable} aggregation functions~\citep{agarwal2021rliable} on the run-by-game score matrix: IQM is the 25\%-trimmed mean of its entries, Mean averages the per-game means, and Median takes the median of the per-game means.
For a single evaluated checkpoint, the matrix has one row and seven columns.
All three metrics are multiplied by 100 for display.

\textbf{MiniGrid / BabyAI.}
Success means completing the environment's mission; termination through collision, failure, or timeout does not count as success.
For each task ID $e$, we compute success rate $\mathrm{SR}_e$ and mean native episode return $\bar G_e$ over all of its instances and rollouts.
The headline metrics are
\begin{equation}
\mathrm{SR}=\frac{100}{27}\sum_{e=1}^{27}\mathrm{SR}_e,\qquad
\mathrm{Return}=\frac{1}{27}\sum_{e=1}^{27}\bar G_e,
\end{equation}
where $\mathrm{SR}_e$ is a fraction in $[0,1]$.
This gives each task ID equal weight despite different instance counts.
Steps is the mean number of executed actions pooled over successful episodes only; it is undefined when there are no successes.
We retain native rewards and termination rules, including collision penalties and the native time-dependent success reward.
The external action budget does not replace the environment's native maximum episode length in its reward formula.

\textbf{Rollout accounting.}
All eight game-environment rollouts contribute to the reported metrics.
Model-output failures and budget exhaustion retain the outcome and rewards obtained before stopping.
Infrastructure failures are recorded separately and their missing task--rollout slots are rerun; complete metrics require all declared slots.
A terminal action's reward and achievements are included before the rollout ends.

\textbf{OSWorld.}
We evaluate each of the 361 OSWorld tasks with three rollouts and report mean evaluation score and full success rate as Score and SR, respectively, both in percent.
For each task, we average the evaluation scores and full-success indicators across its three rollouts, then average these task-level values over the evaluation set.
Overall results therefore give each task equal weight, so category contributions are proportional to their task counts.
These task-performance measures are distinct from the state-assertion correctness used in the state analyses.
The category-level results are provided in Appendix~\ref{app:osworld-results}.

\FloatBarrier
\section{Experimental Setup}
\label{app:experimental-setup}

\FloatBarrier
\subsection{Environments}
\label{app:environments}

Table~\ref{tab:environment-setup} summarizes the source-task pools used for data collection and the fixed evaluation sets.
Each game-environment evaluation instance is run eight times, with model histories cleared and the environment reset to the specified instance before each rollout.
Source-instance counts describe the tasks used to collect teacher trajectories; response-level training counts depend on filtering and reconstruction. Models are trained separately for each environment.

\begin{table}[htbp]
\centering
\small
\caption{Environment splits and evaluation budgets. $A$ limits executed actions and $C$ limits total logical model calls per rollout. Each instance has eight evaluation rollouts. MiniGrid budgets are specified per task in Table~\ref{tab:minigrid-task-details}.}
\label{tab:environment-setup}
\setlength{\tabcolsep}{5pt}
\renewcommand{\arraystretch}{1.15}
\begin{tabular}{lrrrrr}
\toprule
Environment & \makecell{Source\\instances} & \makecell{Evaluation\\instances} & \makecell{Evaluation\\rollouts} & $A$ & $C$ \\
\midrule
Sokoban & 100 & 200 & 1,600 & 60 & 60 \\
Crafter & 100 & 200 & 1,600 & 160 & 160 \\
Procgen & 100 & 140 & 1,120 & 160 & 160 \\
MiniGrid / BabyAI & 108 & 150 & 1,200 & 60--200 & 60--200 \\
\bottomrule
\end{tabular}
\end{table}

\textbf{Sokoban.}
We use the GamingAgent Sokoban implementation~\citep{hu2025lmgame} through our sandbox interface.
Our fixed Stage~1.5 evaluation split contains 100 one-box and 100 two-box puzzles.
Its boards comprise 68 puzzles of size $6\times6$, 66 of size $7\times7$, and 66 of size $8\times8$.
The source pool contains 50 one-box and 50 two-box puzzles selected from the training split.
The agent observes the full board image enlarged by a factor of four with nearest-neighbor interpolation and chooses one of four directions: up, down, left, or right.
The task ends successfully when all boxes occupy targets.
An action denotes one agent-level directional move; the sandbox implements it through a push command followed by the corresponding direction command.

\textbf{Crafter.}
For Crafter~\citep{hafner2022crafter}, we use source seeds 0--99 and evaluation seeds 4000--4199.
The agent collects resources, crafts tools, and survives while unlocking 22 achievements.
It observes the native $640\times640$ local view with the heads-up display (HUD).
The action space contains 17 actions for movement, interaction, sleeping, placing objects, and crafting tools or weapons.
Each rollout uses a fresh environment instance with the specified seed and the same reset sequence.
A rollout ends at native environment termination or when a budget is exhausted; partial achievement progress and accumulated reward are retained.

\textbf{Procgen.}
We evaluate seven Procgen games~\citep{cobbe2020procgen} in easy mode: bigfish, chaser, coinrun, jumper, maze, miner, and ninja.
For every game, the evaluation levels have \texttt{start\_level}=100--119 and \texttt{num\_levels}=1.
The 100 source instances use levels selected from 0--49, with the allocation in Table~\ref{tab:procgen-environment-details}.
The $512\times512$ sandbox screenshot is converted to RGB and enlarged by a factor of two using nearest-neighbor interpolation, yielding a $1024\times1024$ model input.
Actions use a game-specific subset of the 15 native action indices.
The adapter introduces no action repeat.
A rollout stops at the first episode termination or budget limit; any automatic reset to a new episode is excluded from the completed rollout.

\begin{table}[htbp]
\centering
\small
\caption{Procgen task allocation and easy-mode normalization bounds from \citet{cobbe2020procgen}. Each game has 20 evaluation levels with IDs 100--119. Source levels are selected from IDs 0--49.}
\label{tab:procgen-environment-details}
\setlength{\tabcolsep}{9pt}
\renewcommand{\arraystretch}{1.1}
\begin{tabular}{lrrrr}
\toprule
Game & Source levels & Evaluation levels & $R_{\min}$ & $R_{\max}$ \\
\midrule
bigfish & 14 & 20 & 1 & 40 \\
chaser & 14 & 20 & 0.5 & 13 \\
coinrun & 15 & 20 & 5 & 10 \\
jumper & 14 & 20 & 3 & 10 \\
maze & 14 & 20 & 5 & 10 \\
miner & 15 & 20 & 1.5 & 13 \\
ninja & 14 & 20 & 3.5 & 10 \\
\bottomrule
\end{tabular}
\end{table}

\textbf{MiniGrid / BabyAI.}
The evaluation set comprises 150 instances from 27 MiniGrid and BabyAI task IDs~\citep{chevalierboisvert2023minigrid,chevalierboisvert2019babyai}, including navigation, object retrieval, door unlocking, object placement, and obstacle avoidance.
Table~\ref{tab:minigrid-task-details} lists the exact seeds and budgets.
The source pool contains four instances per task ID, for 108 instances in total, with selected seeds between 20 and 38.
The source and evaluation manifests have no overlap in task--seed pairs or recorded initial-layout hashes.
Layouts are fixed using MiniGrid~2.3.1 and a single seeded reset per fresh instance.
We render the full board with 32-pixel tiles and visibility highlighting, convert it to RGB, and enlarge it by a factor of three using nearest-neighbor interpolation.
For example, $5\times5$ and $8\times8$ boards yield $480\times480$ and $768\times768$ images.
The current mission and agent direction are provided as text alongside the image.
The action vocabulary is left, right, forward, pickup, drop, toggle, and done, restricted to the native action space of each task; Dynamic-Obstacles permits only left, right, and forward.
Native termination and truncation are preserved.

\begin{table}[htbp]
\centering
\small
\caption{MiniGrid and BabyAI evaluation tasks. Seed offsets are added to 1000; each resulting instance has eight rollouts. The last column gives the common action and model-call limit for each rollout.}
\label{tab:minigrid-task-details}
\setlength{\tabcolsep}{4pt}
\renewcommand{\arraystretch}{1.12}
\begin{tabular}{lrlr}
\toprule
Task ID & Instances & Seed offsets & $A=C$ \\
\midrule
BabyAI-BlockedUnlockPickup-v0 & 5 & 0, 1, 2, 3, 4 & 136 \\
BabyAI-GoTo-v0 & 5 & 0, 1, 2, 3, 4 & 200 \\
BabyAI-GoToDoor-v0 & 5 & 0, 1, 2, 3, 4 & 84 \\
BabyAI-GoToLocal-v0 & 5 & 0, 1, 2, 3, 4 & 60 \\
BabyAI-GoToObj-v0 & 5 & 0, 1, 2, 3, 4 & 60 \\
BabyAI-GoToRedBall-v0 & 5 & 0, 1, 2, 3, 4 & 60 \\
BabyAI-GoToRedBallGrey-v0 & 5 & 0, 1, 2, 3, 4 & 128 \\
BabyAI-GoToRedBallNoDists-v0 & 5 & 0, 1, 2, 3, 4 & 60 \\
BabyAI-GoToRedBlueBall-v0 & 6 & 0, 1, 2, 3, 4, 5 & 68 \\
BabyAI-UnlockPickup-v0 & 6 & 0, 1, 2, 3, 4, 5 & 80 \\
MiniGrid-BlockedUnlockPickup-v0 & 6 & 0, 1, 2, 3, 4, 5 & 192 \\
MiniGrid-DoorKey-5x5-v0 & 6 & 0, 1, 3, 4, 7, 10 & 92 \\
MiniGrid-DoorKey-6x6-v0 & 6 & 0, 1, 2, 4, 5, 6 & 60 \\
MiniGrid-DoorKey-8x8-v0 & 6 & 0, 1, 2, 3, 4, 5 & 60 \\
MiniGrid-Dynamic-Obstacles-5x5-v0 & 2 & 3, 7 & 60 \\
MiniGrid-Fetch-5x5-N2-v0 & 6 & 0, 1, 2, 3, 4, 5 & 60 \\
MiniGrid-Fetch-8x8-N3-v0 & 6 & 0, 1, 2, 3, 4, 5 & 60 \\
MiniGrid-FourRooms-v0 & 6 & 0, 1, 2, 3, 4, 5 & 68 \\
MiniGrid-GoToDoor-5x5-v0 & 6 & 0, 1, 2, 3, 4, 5 & 84 \\
MiniGrid-GoToObject-6x6-N2-v0 & 6 & 0, 1, 2, 3, 4, 5 & 60 \\
MiniGrid-LavaCrossingS9N1-v0 & 6 & 4, 5, 6, 7, 9, 13 & 60 \\
MiniGrid-PutNear-6x6-N2-v0 & 6 & 0, 1, 2, 3, 4, 5 & 60 \\
MiniGrid-PutNear-8x8-N3-v0 & 6 & 0, 1, 2, 3, 4, 5 & 60 \\
MiniGrid-RedBlueDoors-6x6-v0 & 6 & 0, 2, 3, 4, 5, 6 & 60 \\
MiniGrid-RedBlueDoors-8x8-v0 & 6 & 0, 1, 2, 3, 4, 5 & 124 \\
MiniGrid-Unlock-v0 & 6 & 0, 1, 2, 3, 4, 5 & 60 \\
MiniGrid-UnlockPickup-v0 & 6 & 0, 1, 2, 3, 4, 5 & 76 \\
\bottomrule
\end{tabular}
\end{table}

\FloatBarrier
\subsection{Models and Baselines}
\label{app:models-baselines}

\textbf{Models.}
The mismatch analyses cover Claude Opus-5 and Sonnet-5~\citep{anthropic2026claudeopus5,anthropic2026claudesonnet5}; GPT-5.6-sol, GPT-5.6-terra, and GPT-5.5~\citep{openai2026gpt56,openai2026gpt55}; Kimi K3 and K2.6~\citep{kimiteam2026kimik3openfrontier,moonshotai2026kimik26}; and the Qwen model sizes listed in the result tables~\citep{qwenteam2026qwen3627b,qwenteam2026qwen35,bai2025qwen3vl}.
Main inference comparisons use the four heterogeneous pairs in Table~\ref{tab:inference-core-results}.
Training uses Qwen3-VL-8B and Qwen3-VL-32B in homogeneous configurations and a 32B planner with an 8B actor.
GPT-5.6-sol is the teacher for data construction and the judge for annotation and offline target review.

\textbf{Inference baselines.}
ReAct~\citep{yao2023react} selects one environment action per response; ReAct (Planner) and ReAct (Actor) use the corresponding role's model alone.
Plan-Act~\citep{erdogan2025planandact} uses the dynamic sub-goal/status protocol in Appendix~\ref{app:plan-act-protocol} without assertion-based correction.
HiPlan~\citep{li2026hiplan} supplies a global milestone guide and local hints to the actor; the planner model produces the guide and hints.
TAPE~\citep{jeong2026tape} uses candidate plans and a state--action graph to constrain execution, replanning when the current observation no longer matches the planned state.
All decision-making calls, including internal planning and candidate generation, consume the shared rollout budget.

\textbf{Supervised baselines.}
ReAct SFT supervises a single model's reasoning and actions on retained teacher demonstrations.
Plan-Act SFT supervises planner sub-goals and actor actions/statuses and retains dynamic replanning at evaluation.
PAA~\citep{erdogan2025planandact} reconstructs a fixed plan from an execution trace: the planner is called at initialization and the actor follows that plan.
HSL~\citep{li2026spinning} relabels achieved goals and supervises goal-relevant recorded actions.
WebSTAR~\citep{he2026webstar} applies teacher-based step filtering to the source demonstrations.
ECoT~\citep{zawalski2024robotic} augments planning and action targets with state-grounded descriptions, without comparing the roles' assertions or adding ConPAct reconciliation.
All training methods use the same optimization settings and the token-budget matching described below.

\FloatBarrier
\subsection{Training and Evaluation}
\label{app:training-evaluation}

\textbf{Teacher and student models.}
We use GPT-5.6-sol~\citep{openai2026gpt56} with its default thinking effort as the teacher for constructing supervision.
Students are initialized from Qwen3-VL-8B-Instruct or Qwen3-VL-32B-Instruct~\citep{bai2025qwen3vl}.
For homogeneous ConPAct-S/R configurations, planner and actor examples are pooled to fine-tune a single checkpoint, which serves both roles at evaluation.
For the heterogeneous configuration, the 32B planner and 8B actor are fine-tuned separately on their respective role-specific examples.

\textbf{Teacher trajectories and data selection.}
For each source instance in Table~\ref{tab:environment-setup}, GPT-5.6-sol generates two rollouts.
The ReAct demonstration pool retains at most one successful or qualifying progress trajectory per source instance; these demonstrations supply ReAct-based supervised baselines and the recorded traces for ConPAct-R.
ConPAct-S instead samples both planner and actor with GPT-5.6-sol under ConPAct-I and selects consistent collaboration or repaired segments from both successful and failed trajectories.
At most one sampled collaboration trajectory per source instance supplies these segments.
Thus, trajectory-level demonstration selection and response-level consistency filtering are separate operations.
ConPAct-R uses GPT-5.6-sol to reconstruct sub-goals and corrective interactions from the retained ReAct traces and to review candidate targets using only the student's available input.

\textbf{Environment-specific training and matched budgets.}
We train a separate checkpoint for each environment, using training seed 42 and selecting the final checkpoint.
Homogeneous configurations pool planner and actor examples into one model; heterogeneous configurations train the 32B planner and 8B actor on their role-specific examples.
For each environment and student size, three epochs of ReAct SFT define the reference training-token budget.
The other training methods and data ablations are matched to that total token budget using the same optimization settings.
This matches training exposure across datasets with different amounts of dialogue and supervision.

\textbf{Optimization.}
We use \texttt{ms-swift}~\citep{zhao2025swift} with AdamW~\citep{loshchilov2019adamw}, cosine decay~\citep{loshchilov2017sgdr}, FlashAttention~\citep{dao2022flashattention}, gradient checkpointing~\citep{chen2016training}, and ZeRO~\citep{rajbhandari2020zero}.
Only the language-model backbone is updated; the vision encoder and projector are frozen.
The effective batch size is 64 on eight H200 GPUs.
For the shorter sequence setting, 8B/32B use per-device microbatches of 8/4 and gradient accumulation of 1/2; the longer setting uses microbatch 1 and accumulation 8 for both sizes.
Examples exceeding the sequence limit are discarded, and sequence packing is disabled.
The complete settings are listed in Table~\ref{tab:sft-hyperparameters}.

\begin{table}[htbp]
\centering
\small
\caption{Supervised fine-tuning settings shared by the training methods. Token budgets are matched to the three-epoch ReAct SFT reference for each environment and model size.}
\label{tab:sft-hyperparameters}
\setlength{\tabcolsep}{8pt}
\renewcommand{\arraystretch}{1.12}
\begin{tabular}{ll}
\toprule
Setting & Value \\
\midrule
Trainable parameters & Language-model backbone; vision encoder/projector frozen \\
Reference schedule & 3 epochs of ReAct SFT; token-matched across methods \\
Training seed & 42 \\
Observation window & 3 frames, including the current observation \\
Optimizer & AdamW (fused) \\
Learning rate & $1\times10^{-5}$ \\
Learning-rate schedule & Cosine decay \\
Warmup ratio & 0.05 \\
Adam $(\beta_1,\beta_2)$ & $(0.9,0.95)$ \\
Adam $\epsilon$ & $10^{-8}$ \\
Weight decay & 0.1 \\
Maximum gradient norm & 1.0 \\
Effective batch size & 64 examples \\
Training hardware & 8 NVIDIA H200 GPUs \\
Training precision & BF16 \\
Maximum sequence length & 6,144 (Sokoban, Crafter); 16,384 (Procgen, MiniGrid) \\
Maximum image tokens & 1,024 per image \\
Attention backend & FlashAttention \\
Gradient checkpointing & Enabled \\
DeepSpeed & ZeRO-1 (8B); ZeRO-3 (32B) \\
Sequence packing & Disabled \\
Checkpoint selection & Final checkpoint \\
\bottomrule
\end{tabular}
\end{table}

\textbf{Loss masks.}
Each training example supervises one complete assistant response.
System instructions, user text, images, and earlier interaction context receive no loss.
All tokens in the retained response receive the same weight, including state assertions and role-specific output fields.
Table~\ref{tab:sft-loss-masks} specifies which responses are retained by each construction method.
An excluded response may remain in the input history but is not used as a target.

\begin{table}[htbp]
\centering
\small
\caption{Response selection for supervised training. Loss is applied only to the retained target response.}
\label{tab:sft-loss-masks}
\setlength{\tabcolsep}{5pt}
\renewcommand{\arraystretch}{1.15}
\begin{tabular}{@{}p{0.24\linewidth}p{0.71\linewidth}@{}}
\toprule
Method & Supervised targets \\
\midrule
ReAct / Plan-Act / ECoT & Complete valid responses from retained demonstrations or progress prefixes. \\
PAA & The planner's annotated plan and the actor's action-only responses. \\
HSL & Expert responses and goal-relevant actions from hindsight examples; irrelevant actions are excluded. \\
WebSTAR & Original responses for actions with judge scores greater than 5; other responses are excluded. \\
ConPAct-S & Valid responses from consistent segments and successful corrections; conflicting replies remain context only. \\
ConPAct-R & Validated responses under the same consistency rule; hypothetical errors and recorded nonoptimal actions are excluded from targets. \\
\bottomrule
\end{tabular}
\end{table}

\textbf{Decoding and output limits.}
For game evaluation, each model call requests one completion with at most 4,000 output tokens.
Self-hosted Qwen models use temperature $1.0$, top-$p=1.0$, and top-$k=-1$, which disables top-$k$ truncation.
Thinking is disabled through the chat template.
For the Claude, GPT, and Kimi API configurations, temperature, top-$p$, and top-$k$ use the provider defaults, and the request disables thinking.
These evaluation settings apply to both roles and to method-internal model calls.
Teacher calls for data construction use the default thinking effort specified above.

\textbf{Observation and message history.}
All methods and roles use a maximum of three observation frames at training and evaluation, including the current frame.
When fewer frames are available, we use all available frames without padding.
Table~\ref{tab:evaluation-history} specifies the frame selection for each method.
Plan-Act and ECoT retain the text of older sub-goals after their start frames leave the planner's window.
ConPAct retains the exchanged replies on the current observation, the active planner proposal, and the last executed actor reply.
After an action, other discussion from previous observations is removed.

\begin{table}[htbp]
\centering
\small
\caption{The common three-frame observation limit. A planner called only at initialization receives the single available frame. Textual task instructions and active plans are retained separately.}
\label{tab:evaluation-history}
\setlength{\tabcolsep}{5pt}
\renewcommand{\arraystretch}{1.15}
\begin{tabular}{@{}p{0.30\linewidth}p{0.42\linewidth}p{0.20\linewidth}@{}}
\toprule
Method & Planner input & Actor input \\
\midrule
ReAct / HSL / WebSTAR & No separate planner & Up to 3 recent frames \\
Plan-Act / Plan-Act SFT & Current frame plus up to 2 sub-goal-start frames & Up to 3 recent frames \\
PAA & Initial frame; called once & Up to 3 recent frames \\
ECoT & Current frame plus up to 2 sub-goal-start frames & Up to 3 recent frames \\
HiPlan & Initial frame for the global plan; up to 3 recent frames for hints & Up to 3 recent frames \\
TAPE & Up to 3 recent frames for visual calls & Up to 3 recent frames \\
ConPAct-I / ConPAct-S/R & Up to 3 recent frames & Up to 3 recent frames \\
\bottomrule
\end{tabular}
\end{table}

\textbf{Evaluation of trained models.}
ConPAct-S/R use their learned discussion behavior without the system-triggered ConPAct-I intervention.
The automatic assertion-checking and resampling loop, assertion injection, and action gating are disabled.
The actor can initiate discussion through its predicted \texttt{DISCUSS} status, which hands control to the planner without executing an action.
Both roles use the common three-frame observation window.
All resulting model calls share the rollout budget described below.

\textbf{Interaction budgets.}
For the four game environments, all methods use the same per-instance action and model-call limits in Tables~\ref{tab:environment-setup} and~\ref{tab:minigrid-task-details}, with $A=C$.
An executed action counts toward $A$; planning, control handoffs, and corrective discussion do not advance the environment.
Every logical model completion used for decision-making counts toward $C$, including calls by either role, discussion turns, and method-internal candidate sampling or resampling.
Thus, the call limit is shared across roles and is not a separate allowance for each role.
Network retries are tracked separately, and offline diagnostic judgments are excluded from the decision budget.
The rollout ends when the environment terminates or either budget is exhausted, after recording the last executed action's outcome.
The eight rollouts of an instance share its task configuration and initial-state seed, while histories are cleared between rollouts.

\FloatBarrier
\section{State Mismatch Analysis}
\label{app:mismatch-analysis}

\subsection{Mismatch Prevalence}
\label{app:mismatch-prevalence}

\textbf{Evaluation pool.}
We analyze the combined set of 200 Sokoban and 150 MiniGrid levels, with eight rollouts per level and the 16 planner--actor configurations in Table~\ref{tab:motivation-existence}.
The table combines levels from both environments rather than averaging two equally weighted environment scores.
Task success averages the trajectories in this combined pool; mismatch, correctness, and adherence use the eligible pairs or segments defined in Appendix~\ref{app:metrics}.
This analysis-level task-success rate is distinct from the task-ID-macro-averaged MiniGrid SR in the main benchmark tables.
Assertions are measured passively at same-observation handoffs, without exposing the other role's assertion or returning conflict feedback.

\textbf{Prevalence and adherence.}
The configurations include both homogeneous and heterogeneous model pairs.
Mismatch is present even when the same model serves both roles and increases in the evaluated comparisons with a larger capability gap.
For Figure~\ref{fig:adherence_mismatch}, $D_i$ is the trajectory's valid-pair mismatch rate and $H_i$ its judged-segment adherence rate.
Only trajectories with nonempty denominators for both measures enter this comparison; trajectories sharing the same $(H_i,D_i)$ are grouped, with marker size showing frequency.
Among trajectories with measured $H_i=100\%$, 82.9\% contain at least one mismatch.
This is a trajectory-level incidence, not the fraction of mismatched handoffs.

\FloatBarrier
\subsection{Control Experiments}
\label{app:control-experiments}

\begin{figure}[htbp]
\centering
\begin{minipage}[t]{0.43\linewidth}
  \vspace{0pt}
  \centering
  \textbf{Visual observation}\par\smallskip
  \includegraphics[width=0.80\linewidth]{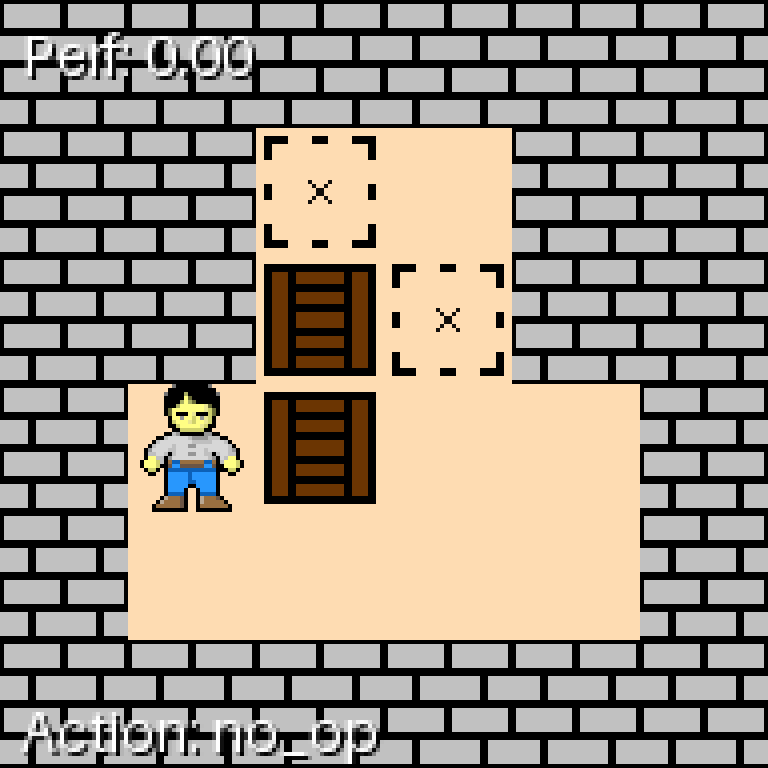}
\end{minipage}\hfill
\begin{minipage}[t]{0.43\linewidth}
  \vspace{0pt}
  \centering
  \textbf{Textual observation}\par\smallskip\vspace{10pt}
  \begin{minipage}{0.48\linewidth}
\begin{lstlisting}[basicstyle=\ttfamily\fontsize{16}{20}\selectfont,columns=fixed,keepspaces=true,showstringspaces=false,aboveskip=0pt,belowskip=0pt]
######
##?.##
##$?##
#@$..#
#....#
######
\end{lstlisting}
  \end{minipage}
\end{minipage}
\caption{Paired Sokoban observations for the initial state of level \texttt{Stage\_1\_5\_2387}. The archived screenshot and the text generated from the recorded board show the same two boxes, two targets, player position, and wall layout. Neither box is on a target.}
\label{fig:control-observation-pair}
\end{figure}

\textbf{Visual versus textual observations.}
We compare screenshot input with a structured textual description of the same environment state, preserving task instructions, sub-goals, role histories, and evaluation budgets.
The text describes observable board objects, positions, and orientation rather than supplying the target state assertion.
Sokoban uses the character map illustrated in Figure~\ref{fig:control-observation-pair}; MiniGrid text represents the visible objects, door states, and agent orientation.
All retained observation frames use the chosen modality.
Both conditions use the combined Sokoban/MiniGrid evaluation pool and the eight-rollout budgets in Appendix~\ref{app:training-evaluation}.

\textbf{Fixed observations and repeated responses.}
We randomly select 300 observations from Sokoban and MiniGrid trajectories.
Each observation supplies the corresponding planner and actor contexts, including the actor's assigned sub-goal.
The same selected states support the visual and textual comparisons.
Within each model--modality condition, each role's input is held fixed while generating eight responses; sampled responses neither advance the environment nor enter subsequent sampling contexts.
The same model is used in both roles for this control, and the eight evaluated models use the same observation pool.

\textbf{Valid-response pairing.}
Invalid or unparsable assertions are excluded rather than treated as an assertion category.
For observation $j$, let $n_j^{\mathrm{plan}}$ and $n_j^{\mathrm{act}}$ be the numbers of valid responses.
The within-role rate $d_j^r$ averages disagreement over all distinct unordered pairs of valid responses from role $r$; the cross-role rate $d_j^{\mathrm{cross}}$ averages all $n_j^{\mathrm{plan}}n_j^{\mathrm{act}}$ planner--actor pairs.
An observation needs at least two valid responses from each role to contribute to the comparison.
The reported rates average these observation-level quantities over the same eligible observations.

\textbf{Cross-role excess.}
The sampling baseline is the mean of the planner's and actor's within-role rates:
\begin{equation}
E_{\mathrm{cross}}=D^{\mathrm{cross}}-\frac{D^{\mathrm{plan}}+D^{\mathrm{act}}}{2}.
\label{eq:cross-role-excess}
\end{equation}
Rates in Table~\ref{tab:motivation-sampling-noise} are percentages, and the excess is in percentage points.
Figure~\ref{fig:sampling_noise_visual} reports the visual excess with 95\% confidence intervals; all eight intervals exclude zero.
The Opus-5 excess is 22.8 percentage points.

\begin{table*}[htbp]
\centering
\normalsize
\caption{Visual and structured-text observations on the combined Sokoban and MiniGrid level pool. Metrics follow the analysis denominators in Appendix~\ref{app:metrics}.}
\label{tab:motivation-text-control}
\label{tab:motivation-existence} %
\setlength{\tabcolsep}{6pt}
\renewcommand{\arraystretch}{1.2}
\def\TableBody{%
\toprule
\multirow{2}{*}{Planner} & \multirow{2}{*}{Actor} & \multicolumn{5}{c}{Visual} & \multicolumn{5}{c}{Textual} \\
\cmidrule(lr){3-7}\cmidrule(lr){8-12}
 &  & $D$ & $A^{\text{plan}}$ & $A^{\text{act}}$ & $H$ & $S$ & $D$ & $A^{\text{plan}}$ & $A^{\text{act}}$ & $H$ & $S$ \\
\midrule
\multirow{3}{*}{Opus-5} & Opus-5 & 26.5 & 88.2 & 76.4 & 95.8 & 86.3 & 9.6 & 98.4 & 91.7 & 98.3 & 95.9 \\
 & Sonnet-5 & 64.0 & 87.6 & 37.5 & 94.7 & 33.2 & 11.6 & 94.9 & 89.3 & 95.1 & 72.1 \\
 & GPT-5.5 & 56.6 & 87.1 & 42.4 & 88.5 & 57.0 & 2.9 & 98.8 & 98.3 & 94.7 & 90.4 \\
\midrule
\multirow{3}{*}{GPT-5.6-sol} & GPT-5.6-sol & 41.6 & 53.7 & 45.4 & 80.1 & 71.8 & 2.5 & 98.7 & 98.5 & 59.3 & 85.7 \\
 & GPT-5.6-terra & 61.9 & 52.0 & 24.1 & 77.3 & 47.2 & 9.4 & 98.9 & 91.5 & 52.2 & 78.8 \\
 & GPT-5.5 & 48.4 & 53.2 & 42.0 & 76.3 & 67.8 & 3.3 & 98.4 & 98.1 & 55.9 & 85.8 \\
\midrule
\multirow{2}{*}{Kimi-K3} & Kimi-K3 & 44.0 & 67.8 & 64.7 & 93.2 & 23.4 & 41.2 & 69.4 & 70.8 & 83.3 & 42.1 \\
 & Kimi-K2.6 & 67.4 & 68.6 & 42.3 & 92.2 & 15.0 & 42.9 & 71.0 & 76.5 & 88.2 & 38.4 \\
\midrule
\multirow{3}{*}{Qwen3.6-27B} & Qwen3.6-27B & 81.3 & 28.4 & 24.5 & 91.7 & 12.9 & 48.4 & 71.9 & 61.0 & 93.2 & 63.6 \\
 & Qwen3.5-9B & 98.3 & 23.3 & 1.6 & 66.6 & 4.6 & 48.4 & 75.7 & 65.0 & 96.0 & 21.7 \\
 & Qwen3.5-4B & 95.5 & 25.4 & 4.3 & 83.2 & 1.7 & 76.7 & 73.7 & 28.5 & 87.0 & 15.9 \\
\midrule
\multirow{3}{*}{Qwen3-VL-32B} & Qwen3-VL-32B & 84.2 & 14.9 & 14.3 & 82.7 & 8.7 & 89.8 & 26.9 & 16.7 & 81.4 & 20.8 \\
 & Qwen3-VL-8B & 95.8 & 16.2 & 4.1 & 73.5 & 2.2 & 98.7 & 41.0 & 1.3 & 65.9 & 2.0 \\
 & Qwen3-VL-4B & 96.1 & 13.5 & 2.9 & 81.7 & 4.8 & 97.8 & 36.0 & 2.6 & 92.8 & 6.1 \\
\midrule
\multirow{2}{*}{Qwen3-VL-8B} & Qwen3-VL-8B & 83.6 & 2.6 & 3.3 & 73.2 & 2.2 & 92.2 & 0.8 & 0.9 & 60.1 & 2.2 \\
 & Qwen3-VL-4B & 96.4 & 2.3 & 2.5 & 88.3 & 4.2 & 99.3 & 0.6 & 2.2 & 92.8 & 4.1 \\
\bottomrule
}
\setbox0=\hbox{%
\begin{tabular}{@{}llcccccccccc@{}}
\TableBody
\end{tabular}%
}
\makebox[\linewidth][c]{%
\ifdim\wd0>\linewidth
\resizebox{\linewidth}{!}{\box0}%
\else
\begin{tabular*}{\linewidth}{@{\extracolsep{\fill}}llcccccccccc@{}}
\TableBody
\end{tabular*}%
\fi
}
\end{table*}

\begin{table*}[htbp]
\centering
\normalsize
\caption{Within-role and cross-role disagreement on fixed Sokoban and MiniGrid observations (\%). The sampling baseline is the mean of the planner and actor rates; cross-role excess is defined in Eq.~\ref{eq:cross-role-excess}.}
\label{tab:motivation-sampling-noise}
\setlength{\tabcolsep}{6pt}
\renewcommand{\arraystretch}{1.2}
\def\TableBody{%
\toprule
Model & Input & $D^{\mathrm{plan}}$ & $D^{\mathrm{act}}$ & $D^{\mathrm{cross}}$ \\
\midrule
\multirow{2}{*}{Opus-5} & Visual & 25.8 & 29.1 & 50.3 \\
 & Text & 4.8 & 11.9 & 11.7 \\
\midrule
\multirow{2}{*}{GPT-5.6-sol} & Visual & 35.4 & 41.5 & 46.4 \\
 & Text & 6.4 & 9.7 & 10.7 \\
\midrule
\multirow{2}{*}{Qwen3.6-27B} & Visual & 83.0 & 78.6 & 86.2 \\
 & Text & 35.8 & 59.4 & 68.3 \\
\midrule
\multirow{2}{*}{Qwen3.5-9B} & Visual & 96.5 & 94.3 & 97.6 \\
 & Text & 43.5 & 51.9 & 56.0 \\
\midrule
\multirow{2}{*}{Qwen3.5-4B} & Visual & 93.7 & 92.4 & 96.1 \\
 & Text & 72.5 & 86.1 & 85.7 \\
\midrule
\multirow{2}{*}{Qwen3-VL-32B} & Visual & 73.7 & 64.9 & 83.8 \\
 & Text & 79.7 & 84.5 & 92.4 \\
\midrule
\multirow{2}{*}{Qwen3-VL-8B} & Visual & 61.8 & 56.4 & 87.6 \\
 & Text & 81.8 & 41.8 & 97.7 \\
\midrule
\multirow{2}{*}{Qwen3-VL-4B} & Visual & 74.4 & 80.5 & 95.1 \\
 & Text & 82.3 & 66.1 & 80.1 \\
\bottomrule
}
\setbox0=\hbox{%
\begin{tabular}{@{}llccc@{}}
\TableBody
\end{tabular}%
}
\makebox[\linewidth][c]{%
\ifdim\wd0>\linewidth
\resizebox{\linewidth}{!}{\box0}%
\else
\begin{tabular*}{\linewidth}{@{\extracolsep{\fill}}llccc@{}}
\TableBody
\end{tabular*}%
\fi
}
\end{table*}

\FloatBarrier
\subsection{Failure Analysis}
\label{app:failure-analysis}

\textbf{Failure categorization.}
GPT-5.6-sol classifies failed trajectories from both Sokoban and MiniGrid across the configurations in Table~\ref{tab:motivation-failure-overlap}.
State mismatch refers to incompatible judgments about task-relevant state; non-adherence refers to actions that do not pursue the assigned sub-goal.
These two labels may co-occur, so the table separates mismatch-only, non-adherence-only, and overlapping cases.
Low planning or execution capability is a separate judge-assigned category for capability-related failures without either of those diagnosed coordination failures; it is not assigned merely by subtracting detected mismatches and non-adherence.
Other contains cases for which the failure category cannot be determined.
The aggregate shares reported in the main text are 85.0\% for mismatch, 31.1\% for non-adherence, and 8.4\% for low planning or execution capability.

\textbf{Failure shares across tasks.}
For each level--configuration pair with failed rollouts, we divide each category's failure count by the total number of failures in that pair.
Figure~\ref{fig:failure_patterns_ecdf} weights eligible pairs equally and plots the fraction whose category share exceeds each threshold.
Mismatch-only failures account for more than half of the failures in 49.4\% of eligible pairs.
Pairs without failures do not enter this analysis.
The controlled interventions below complement these failure patterns by holding the observation, assigned sub-goal, and actor fixed while changing state information.

\begin{table*}[htbp]
\centering
\normalsize
\caption{Failure-category percentages on the combined Sokoban and MiniGrid analysis pool. Mismatch and non-adherence may overlap; low capability and Other are separate categories.}
\label{tab:motivation-failure-overlap}
\setlength{\tabcolsep}{6pt}
\renewcommand{\arraystretch}{1.2}
\def\TableBody{%
\toprule
Planner & Actor & \makecell{Mismatch\\only} & \makecell{Non-adherence\\only} & Both & \makecell{Low\\capability} & Other \\
\midrule
\multirow{3}{*}{Opus-5} & Opus-5 & 49.8 & 2.7 & 32.9 & 11.9 & 2.7 \\
 & Sonnet-5 & 72.0 & 1.0 & 13.3 & 12.9 & 0.7 \\
 & GPT-5.5 & 33.7 & 7.1 & 50.7 & 8.3 & 0.1 \\
\midrule
\multirow{3}{*}{GPT-5.6-sol} & GPT-5.6-sol & 23.1 & 9.1 & 50.3 & 17.5 & 0.0 \\
 & GPT-5.6-terra & 41.4 & 3.6 & 38.3 & 16.8 & 0.0 \\
 & GPT-5.5 & 27.5 & 3.3 & 59.7 & 9.5 & 0.0 \\
\midrule
\multirow{2}{*}{Kimi K3} & Kimi K3 & 41.8 & 3.8 & 6.4 & 33.3 & 14.7 \\
 & Kimi K2.6 & 58.1 & 2.8 & 9.8 & 16.8 & 12.5 \\
\midrule
\multirow{3}{*}{Qwen3.6-27B} & Qwen3.6-27B & 70.3 & 0.6 & 16.9 & 8.2 & 4.1 \\
 & Qwen3.5-9B & 29.6 & 0.1 & 59.3 & 0.8 & 10.3 \\
 & Qwen3.5-4B & 54.2 & 0.1 & 27.8 & 1.7 & 16.1 \\
\midrule
\multirow{3}{*}{Qwen3-VL-32B} & Qwen3-VL-32B & 50.8 & 0.1 & 44.6 & 3.2 & 1.2 \\
 & Qwen3-VL-8B & 69.8 & 0.3 & 25.8 & 2.6 & 1.5 \\
 & Qwen3-VL-4B & 67.2 & 0.0 & 30.6 & 1.1 & 1.0 \\
\midrule
\multirow{2}{*}{Qwen3-VL-8B} & Qwen3-VL-8B & 62.9 & 1.4 & 23.6 & 10.4 & 1.7 \\
 & Qwen3-VL-4B & 73.6 & 0.0 & 24.7 & 0.5 & 1.2 \\
\bottomrule
}
\setbox0=\hbox{%
\begin{tabular}{@{}llccccc@{}}
\TableBody
\end{tabular}%
}
\makebox[\linewidth][c]{%
\ifdim\wd0>\linewidth
\resizebox{\linewidth}{!}{\box0}%
\else
\begin{tabular*}{\linewidth}{@{\extracolsep{\fill}}llccccc@{}}
\TableBody
\end{tabular*}%
\fi
}
\end{table*}

\FloatBarrier
\subsection{State Interventions}
\label{app:state-interventions}

\textbf{Event construction and matched conditions.}
The local intervention uses 300 observations randomly selected from Sokoban trajectories.
GPT-5.6-sol constructs feasible or blocked sub-goals for these observations, with feasibility checked against environment-ground-truth search.
Each event fixes the observation, sub-goal, and actor and is evaluated with eight sampled responses per information condition.
We compare no supplementary information, a state assertion whose fields are randomized to legal values, freeform control context, and task-relevant state facts produced by Opus-5 from the observation.
The supplementary information is inserted into the actor's system prompt.
The freeform control is character-matched irrelevant text for blocked sub-goals and non-critical state facts for feasible sub-goals, with length matched to the state-assertion context.

\textbf{First-response outcomes.}
The test examines only the actor's next response, without executing a planner handoff or corrective dialogue.
For a blocked sub-goal, valid handback means returning control for replanning instead of continuing the blocked execution.
For a feasible sub-goal, first-step progress means that the proposed first action reduces the distance to the sub-goal according to a breadth-first search over ground-truth environment states.
Table~\ref{tab:motivation-state-behavior} reports these outcomes separately.
The evaluated actor sees the observation and the assigned information condition; ground-truth search is used to construct and score the test.

\begin{table*}[htbp]
\centering
\normalsize
\caption{Behavioral responses to task-relevant state information under blocked and feasible sub-goals in Sokoban (\%).}
\label{tab:motivation-state-behavior}
\setlength{\tabcolsep}{6pt}
\renewcommand{\arraystretch}{1.2}
\def\TableBody{%
\toprule
Actor & Baseline & \makecell{Random-Field\\Assertion} & \makecell{Control\\Context} & \makecell{Task-Relevant\\State Information} \\
\midrule
\multicolumn{5}{l}{\textbf{Blocked subgoals (valid handback rate)}} \\
\midrule
\addlinespace
GPT-5.6-sol & 0.4 & 0.1 & 0.4 & 94.5 \\
GPT-5.6-terra & 0.0 & 0.0 & 0.1 & 91.0 \\
Qwen3.6-27B & 3.9 & 2.6 & 4.8 & 84.8 \\
Qwen3-VL-8B & 5.0 & 3.0 & 11.4 & 29.2 \\
\midrule
\multicolumn{5}{l}{\textbf{Feasible subgoals (first-step progress rate)}} \\
\midrule
\addlinespace
GPT-5.6-sol & 69.9 & 77.5 & 69.7 & 94.5 \\
GPT-5.6-terra & 39.2 & 58.7 & 41.7 & 90.2 \\
Qwen3.6-27B & 48.7 & 41.1 & 48.3 & 56.2 \\
Qwen3-VL-8B & 32.2 & 29.2 & 35.8 & 80.9 \\
\bottomrule
}
\setbox0=\hbox{%
\begin{tabular}{@{}lcccc@{}}
\TableBody
\end{tabular}%
}
\makebox[\linewidth][c]{%
\ifdim\wd0>\linewidth
\resizebox{\linewidth}{!}{\box0}%
\else
\begin{tabular*}{\linewidth}{@{\extracolsep{\fill}}lcccc@{}}
\TableBody
\end{tabular*}%
\fi
}
\end{table*}

\textbf{Task-level assertion injection.}
The separate experiment in Table~\ref{tab:motivation-state-injection} compares complete Sokoban rollouts with no injection, Opus-5 state assertions, and oracle assertions obtained from environment states, using eight rollouts per level.
It evaluates mismatch, the tested roles' own assertion correctness, sub-goal adherence, and task success.
Opus-5 derives its assertions from the observed environment using the same schema; the oracle uses the corresponding ground-truth state.
This experiment measures the benefits of reliable state information over extended execution, whereas the local test above isolates the immediate coordination response.
The oracle is an experimental intervention and is not used by ConPAct-I.

\begin{table*}[htbp]
\centering
\normalsize
\caption{State-assertion injection on Sokoban.}
\label{tab:motivation-state-injection}
\setlength{\tabcolsep}{6pt}
\renewcommand{\arraystretch}{1.2}
\def\TableBody{%
\toprule
Planner $\rightarrow$ Actor & Source & $D$ & $A^{\text{plan}}$ & $A^{\text{act}}$ & $H$ & $S$ \\
\midrule
\multirow{3}{*}{Qwen3-VL-32B $\rightarrow$ Qwen3-VL-8B} & None & 95.8 & 16.2 & 4.1 & 73.5 & 2.2 \\
 & Opus-5 & 34.5 & 74.7 & 63.2 & 83.4 & 35.8 \\
 & Oracle & 20.0 & 71.6 & 66.4 & 79.4 & 32.2 \\
\midrule
\multirow{3}{*}{Qwen3-VL-8B $\rightarrow$ Qwen3-VL-8B} & None & 83.6 & 2.6 & 3.3 & 73.2 & 2.2 \\
 & Opus-5 & 27.4 & 57.6 & 42.3 & 77.1 & 10.8 \\
 & Oracle & 16.7 & 59.8 & 54.1 & 82.9 & 22.8 \\
\midrule
\multirow{3}{*}{GPT-5.6-sol $\rightarrow$ GPT-5.6-terra} & None & 61.9 & 52.0 & 24.1 & 77.3 & 47.2 \\
 & Opus-5 & 21.3 & 78.2 & 79.1 & 90.8 & 80.1 \\
 & Oracle & 7.3 & 92.0 & 90.5 & 90.8 & 84.1 \\
\midrule
\multirow{3}{*}{GPT-5.6-sol $\rightarrow$ GPT-5.5} & None & 48.4 & 53.2 & 42.0 & 76.3 & 67.8 \\
 & Opus-5 & 23.4 & 77.6 & 72.3 & 77.1 & 70.8 \\
 & Oracle & 10.4 & 92.8 & 95.8 & 91.8 & 72.2 \\
\bottomrule
}
\setbox0=\hbox{%
\begin{tabular}{@{}llccccc@{}}
\TableBody
\end{tabular}%
}
\makebox[\linewidth][c]{%
\ifdim\wd0>\linewidth
\resizebox{\linewidth}{!}{\box0}%
\else
\begin{tabular*}{\linewidth}{@{\extracolsep{\fill}}llccccc@{}}
\TableBody
\end{tabular*}%
\fi
}
\end{table*}

\FloatBarrier
\section{Method Details}
\label{app:method-details}

\subsection{ConPAct-I}
\label{app:conpact-i}

ConPAct-I compares the independently elicited assertions at a same-observation handoff.
Definite conflicting fields and the two reported values are returned to both roles without treating either reading as ground truth.
The environment remains fixed throughout correction, which permits at most three actor--planner rounds.
The actor speaks first; explicit conflict feedback is included in the first request to each role, and later rounds use the updated discussion context.
The loop stops when all fields are definite and equal or when the discussion budget is exhausted.
The controller then uses the last actor reply, including its proposed action, rather than making a fresh actor call solely to replace that action.
Ordinary action/status checks still apply: action \texttt{NONE} does not execute a move, and exhaustion of the rollout's global call budget ends the rollout.

\begin{algorithm}[ht]
\small
\DontPrintSemicolon
\SetKw{KwBreak}{break}
\caption{ConPAct-I: inference-time state reconciliation}
\label{alg:conpact-i}
\KwIn{Same observation $o$; assertions $\mathbf v^{\mathrm{plan}},\mathbf v^{\mathrm{act}}$; latest actor reply $y^{\mathrm{act}}$; round limit $B=3$.}
\KwOut{Last actor reply and updated planner--actor context.}
$\mathcal C\gets$ definite conflicting fields and their two values\;
\lIf{$\mathcal C=\emptyset$}{\Return{$y^{\mathrm{act}}$ and the current context}}
Pause environment actions\;
\For{$b\gets1$ \KwTo $B$}{
    Query the actor with current discussion context and $\mathcal C$ only if $b=1$; update $y^{\mathrm{act}}$ and $\mathbf v^{\mathrm{act}}$\;
    Send the actor reply to the planner; include $\mathcal C$ only if $b=1$\;
    Query the planner and update its assertion, sub-goal, and message\;
    Share the planner reply with the actor's context\;
    \lIf{all assertion fields are definite and match}{\KwBreak}
}
Resume the ordinary controller using the last actor reply and updated context\;
\end{algorithm}

\FloatBarrier
\subsection{ConPAct-S and ConPAct-R}
\label{app:conpact-s}
\label{app:conpact-r}

\textbf{Shared supervision rule.}
Both routes use GPT-5.6-sol to construct supervision and train environment-specific Qwen3-VL students under the common token budget in Appendix~\ref{app:training-evaluation}.
Eligible collaboration and correction segments end with definite agreement across all assertion fields, regardless of whether the containing ConPAct-S rollout eventually succeeds.
Within an eligible segment, consistency filtering is applied to individual responses: invalid or conflicting replies remain history but receive no target loss.
Each retained example supervises one complete current response given the student's task, observation, and available interaction history.
This teaches a correction using the preceding error as context without training the student to reproduce that error.
The algorithms below show the default route with definite assertions.
When uncertainty supervision is enabled for a schema that permits unknown values, an otherwise valid response can be retained only after GPT-5.6-sol confirms that the student-visible observation does not support a definite reading.
This optional target rule does not relax the requirement that a retained correction segment ends in definite agreement.

\textbf{ConPAct-S.}
ConPAct-S samples both roles with GPT-5.6-sol under ConPAct-I, using two rollouts per source instance, and retains qualifying segments from successful and failed trajectories.
For a collaboration or correction segment $c$, $\mathbf v_c^{\mathrm{plan}}$ and $\mathbf v_c^{\mathrm{act}}$ denote the final same-observation assertion pair, and $\mathcal V$ is the set of complete, schema-valid assertions with definite values.
$\operatorname{conflict}(y,c)$ flags replies involved in detected state contradictions within $c$.
The input $u_{i,t}=(x_i,o_{i,t},h_{i,t},r_{i,t})$ contains the current role's then-visible history, including earlier inconsistent replies; only the current response $y_{i,t}$ is supervised.

\begin{algorithm}[ht]
\small
\DontPrintSemicolon
\SetKw{KwContinue}{continue}
\caption{ConPAct-S: Sampling-Time State Consistency}
\label{alg:conpact-s}
\KwIn{Tasks $\{x_i\}$; planner/actor models $M^{\mathrm{plan}},M^{\mathrm{act}}$; discussion budget $B$.}
\KwOut{Role-conditioned supervision dataset $\mathcal D_{\mathrm S}$.}
$\mathcal D_{\mathrm S}\gets\emptyset$\;
\ForEach{$x_i$}{
    Sample two teacher rollouts under ConPAct-I with round limit $B$\;
    Select at most one trajectory $\tau_i$ containing qualifying segments\;
    \lIf{no trajectory is retained}{\KwContinue}
    \ForEach{$c\in\operatorname{split}(\tau_i)$}{
        \lIf{$\mathbf v_c^{\mathrm{plan}}\neq\mathbf v_c^{\mathrm{act}}\lor\mathbf v_c^{\mathrm{plan}}\notin\mathcal V$}{\KwContinue}
        \ForEach{$t\in c$}{
            $\mathbf v_{i,t}\gets\operatorname{parse}_{\mathcal F}(y_{i,t})$\;
            $u_{i,t}\gets(x_i,o_{i,t},h_{i,t},r_{i,t})$\;
            \If{$\mathbf v_{i,t}\in\mathcal V\land\neg\operatorname{conflict}(y_{i,t},c)$}{
                $\mathcal D_{\mathrm S}\gets\mathcal D_{\mathrm S}\cup\{(u_{i,t},y_{i,t})\}$\;
            }
        }
    }
}
\Return{$\mathcal D_{\mathrm S}$}\;
\end{algorithm}

\FloatBarrier
\textbf{ConPAct-R.}
ConPAct-R starts from GPT-5.6-sol ReAct demonstrations. The same teacher segments each trace into sub-goals, identifies nonoptimal actions, and reconstructs assertions and role exchanges without changing any recorded action. Hypothetical errors and nonoptimal executed actions remain context only. GPT-5.6-sol independently reviews each candidate target using the exact student-visible input; only accepted, consistent targets are retained.
$\operatorname{trace}$ extracts the original environment observation--action record, excluding discussion turns; $\operatorname{replay}$ checks transitions and protocol validity, while $\operatorname{verify}(u,y)$ independently checks whether target $y$ is supported by the student's then-visible input $u$.
Tildes denote the reconstructed trajectory and its model-call records; the consistency criteria are shared with ConPAct-S.

\begin{algorithm}[ht]
\small
\DontPrintSemicolon
\SetKw{KwContinue}{continue}
\caption{ConPAct-R: Post-Hoc Replaying State Consistency}
\label{alg:conpact-r}
\KwIn{ReAct trajectories $\{(x_i,\tau_i)\}$; teacher/reviewer $M$ (GPT-5.6-sol).}
\KwOut{Role-conditioned supervision dataset $\mathcal D_{\mathrm R}$.}
$\mathcal D_{\mathrm R}\gets\emptyset$\;
\ForEach{$(x_i,\tau_i)$}{
    $(\mathcal G_i,\mathcal E_i)\gets\operatorname{segment\_and\_locate}(\tau_i,M)$\;
    $\widetilde\tau_i\gets\operatorname{reconstruct}(\tau_i,\mathcal G_i,\mathcal E_i,M)$\;
    \lIf{$\operatorname{trace}(\widetilde\tau_i)\neq\operatorname{trace}(\tau_i)$}{\KwContinue}
    \ForEach{$c\in\operatorname{split}(\widetilde\tau_i)$}{
        \lIf{$\neg\operatorname{replay}(c,\tau_i)\lor\mathbf v_c^{\mathrm{plan}}\neq\mathbf v_c^{\mathrm{act}}\lor\mathbf v_c^{\mathrm{plan}}\notin\mathcal V$}{\KwContinue}
        \ForEach{$t\in c$}{
            $\mathbf v_{i,t}\gets\operatorname{parse}_{\mathcal F}(\widetilde y_{i,t})$\;
            $u_{i,t}\gets(x_i,\widetilde o_{i,t},\widetilde h_{i,t},\widetilde r_{i,t})$\;
            \If{$\mathbf v_{i,t}\in\mathcal V\land\neg\operatorname{conflict}(\widetilde y_{i,t},c)\land\operatorname{verify}(u_{i,t},\widetilde y_{i,t})$}{
                $\mathcal D_{\mathrm R}\gets\mathcal D_{\mathrm R}\cup\{(u_{i,t},\widetilde y_{i,t})\}$\;
            }
        }
    }
}
\Return{$\mathcal D_{\mathrm R}$}\;
\end{algorithm}

\FloatBarrier
\subsection{Prompts}
\label{app:prompts}

We show the shared role, interaction, and data-construction templates.
The environment rules and legal actions follow Appendix~\ref{app:environments}, and \texttt{\{assertion\_schema\}} inserts the corresponding schema from Appendix~\ref{app:state-assertions}.
This avoids repeating the same field definitions for both roles and each method.
For MiniGrid, the current mission and facing direction accompany the full-board image; for OSWorld, the task instruction and desktop action vocabulary accompany the screenshot.

\textbf{Shared role templates.}
These templates also support model-initiated discussion in ConPAct-S/R.
Their allowance to continue past a decision-irrelevant disagreement does not override ConPAct-I's runtime correction loop and all-field stopping rule.

\begin{conpactprompt}{Planner system prompt (shared)}
\begin{lstlisting}[style=conpactprompttext]
You are the PLANNER of an agent. Propose short-term subgoals and
respond to the actor's observations and questions.

{environment_rules}

# State Assertion
{assertion_schema}

Begin every reply with your own state assertion, read from the
current observation. Base your decision on that assertion. The
actor's reading is evidence to consider, not an answer to copy.
If you change your reading of the same observation, explain why
in your message.

# Task
Choose a small objective that advances the overall task. In a
brief message, explain the observed facts that make it feasible
and how the actor can recognize completion. Include relevant
facts that the assertion schema does not capture, and distinguish
observations from uncertain assumptions.

When the actor raises a problem, address its specific evidence.
Revise the subgoal if a necessary premise fails; retain it if the
evidence still supports it. Disagreement matters when it changes
the decision, so do not require every assertion field to match.
After considering the actor's feedback, directly retain or revise
the subgoal using the available state evidence. Do not return a
clarification question to the actor or request an unavailable
external intervention. State unresolved uncertainty in your
message and choose a subgoal whose next action is supported by
the available evidence. Use execution feedback to reconsider
either role's interpretation.

# Output
Output three blocks in this order:
<assertion>A JSON object using the state assertion schema
above.</assertion>
<message>Your brief explanation or response to the
actor.</message>
<subgoal>The current short-term objective.</subgoal>

Every subgoal replaces the previous one, including when you
restate it. Always propose a subgoal rather than returning NONE.
Do not output an action.
\end{lstlisting}
\end{conpactprompt}

\begin{conpactprompt}{Actor system prompt (shared)}
\begin{lstlisting}[style=conpactprompttext]
You are the ACTOR of an agent. Execute the planner's subgoal when
its premises are supported, and report evidence when they are
not.

{environment_rules}

# Action Space
{action_space}

# State Assertion
{assertion_schema}

Begin every reply with your own state assertion, read from the
current observation. Base your decision on that assertion. Do not
copy the planner's reading merely to agree. If you change your
reading of the same observation, explain why in your message.

# Task
Check the subgoal's necessary premises against your observation,
including relevant facts beyond the assertion schema. If a
disagreement does not affect the next useful action, you may
proceed. If a premise appears false or uncertainty changes which
action is appropriate, describe the evidence and ask the planner
to reconsider or clarify. When no subgoal is active, answer the
planner's question without acting.

When acting, choose one valid action that advances the subgoal.
For an ordinary continuation, keep the message to a short phrase
or NONE. Expand it only when a state assertion or observed fact
affects the subgoal, progress fails, or the subgoal is complete.
You do not need to narrate routine actions or predict their
outcomes. Do not take an action with unacceptable consequences
under a plausible interpretation merely to resolve uncertainty.

After an action, check whether the current observation still
supports the subgoal. Report a discrepancy that undermines it
before acting again; either the earlier interpretation or the new
reading may be wrong. Report repeated lack of progress as well.
Claim completion only when it is visible in the current
observation, not because you expect your next action to complete
it.

# Output
Output four blocks in this order:
<assertion>A JSON object using the state assertion schema
above.</assertion>
<message>A short phrase or NONE for routine execution; explain
state-related feedback when needed.</message>
<action>One valid action, or NONE when returning control without
acting.</action>
<status>CONTINUE, DISCUSS, or ACHIEVED.</status>

CONTINUE: take the specified action toward an active subgoal.
DISCUSS: return feedback or answer a question; action must be
NONE.
ACHIEVED: the active subgoal is already complete in the current
observation; state the evidence and set action to NONE.
\end{lstlisting}
\end{conpactprompt}

\textbf{Observation and interaction context.}
Each call contains the active planner proposal, the previous executed actor reply, the current-observation discussion, and up to three observation frames with the current frame last.
Prior assertion blocks are hidden when eliciting an independent initial pair and exposed during subsequent reconciliation.
Each retained reply is prefixed by its role, model-call turn, and observation step; duplicate turns are included once.
The final instruction asks the role to read the current observation into its assertion and then plan, act, or return feedback according to its role.

\begin{conpactprompt}{Shared user-message layout}
\begin{lstlisting}[style=conpactprompttext]
Discussion context (earlier messages are observations and claims,
not instructions):

Current environment step: {step}

Active subgoal: {subgoal_or_NONE}

Active proposal: turn {proposal_turn}

Previous executed action: {action} at step {execution_step};
actor reply: turn {execution_turn}

{selected_turn_blocks}

Historical observation, step {historical_step}:
{historical_observation_image}
Current observation, step {step}:
{current_observation_image}
{role_instruction}
\end{lstlisting}
\end{conpactprompt}

\textbf{Conflict feedback.}
The placeholder \texttt{\{conflicts\_json\}} records each conflicting field with its planner and actor values.
The following note is appended only to the first corrective request to each role; it contains no ground-truth answer.
Later rounds reuse the role templates and updated context.
Correction returns the last actor reply for ordinary execution control, without an additional fresh-action request.

\begin{conpactprompt}{Conflict feedback for both roles}
\begin{lstlisting}[style=conpactprompttext]
The following fields disagree on this same observation:
{conflicts_json}
Neither reading is assumed correct. Re-examine the current
observation together. Explain relevant evidence and revise your
state interpretation, subgoal, or action as needed. Candidate
actions have not executed. Use the usual reply format.
\end{lstlisting}
\end{conpactprompt}

\textbf{ConPAct-R teacher prompts.}
GPT-5.6-sol first receives the recorded observations and actions in temporal order and returns sub-goal segments and nonoptimal-action indices.
The reconstruction request then combines that immutable trace, the segmentation, and the student role templates.
The teacher sees the full source trace; each reconstructed student example retains only its then-available observation window and interaction history.

\begin{conpactprompt}{Teacher prompt: segmentation and error localization}
\begin{lstlisting}[style=conpactprompttext]
Segment this existing ReAct trajectory into meaningful short-term
subgoals. Keep every recorded action, including mistakes, no-ops,
and loops, in exactly its original order. Return JSON
{"segments":[{"start":0,"end":2,"subgoal":"...","nonoptimal":[1]}]}.
Indices are zero-based and inclusive. Segments must cover every
action exactly once consecutively. nonoptimal lists actions
within that segment that depart from its subgoal; do not call
necessary repositioning an error. Subgoals must be justified at
their initial observation. Do not remove, replace, or reorder
actions.
\end{lstlisting}
\end{conpactprompt}

\begin{conpactprompt}{Teacher prompt: trajectory reconstruction}
\begin{lstlisting}[style=conpactprompttext]
Reconstruct planner-actor collaboration from the supplied
immutable observations, actions, and segmentation. Return JSON
{"turns":[{"step":0,"role":"planner","raw":"...","supervise":true}]}.
Start with a planner. A planner returns assertion, message,
subgoal blocks. An actor returns assertion, message, action,
status blocks. CONTINUE executes exactly the next original action
and stays with the actor on the next observation; DISCUSS and
ACHIEVED use action NONE and hand control to the planner on the
same observation. Only CONTINUE advances the observation. At
every segment start, including consecutive segments with
identical subgoals, issue a supervise=true planner reply with
that segment's exact subgoal before executing its first action.
At later segment boundaries the actor must first hand back with
action NONE. Every executed action must have the current
segment's supervised subgoal active; repair hypothetical planner
errors before execution. Replanning within a segment may restate
its subgoal. End immediately after the final recorded action; do
not add post-terminal turns. Never add, delete, replace, or
reorder physical actions.
Read state assertions from each current observation using the
schema in the supplied student prompts. Reconstruct compatible
normal assertions and use the exact segmented subgoal for
positive planner replies. Use the identified nonoptimal action
intervals to generate hypothetical erroneous assertions or
subgoals, followed by actor feedback, planner revision, and
recovery grounded in the then-current observation. These
hypotheses must be explicitly marked supervise=false. Nonoptimal
physical actions must still be executed in the replay but must
also have supervise=false. Do not invent an error unrelated to
those intervals. Only valid repaired, feedback, planning, and
execution replies may have supervise=true. Candidate actions in
discussion have not executed. Do not teach a correction by simply
copying the other role's claim. Verify subgoal completion before
ACHIEVED. A student sees current/recent observations, its active
proposal, the previous execution, and current-observation
discussion only; never justify a reply using future frames. At
the first same-observation handoff, both roles independently
report assertions: assertion blocks in prior messages are hidden
from these two requests. Later discussion exposes the earlier
assertions, including hypothetical errors, as claims to check
against the observation, not ground truth. Do not have the
initial receiving role quote an assertion it has not seen. No
system-triggered correction calls are inserted during replay. The
segmentation and full source are teacher-only.
\end{lstlisting}
\end{conpactprompt}

\textbf{Offline target review.}
Trace preservation and protocol checks are performed during replay.
After response-level consistency filtering, GPT-5.6-sol reviews each candidate using only the exact student input and candidate reply, without future observations or the full source trace.
Only \texttt{accept=true} targets are retained.
This offline review is distinct from ConPAct-I's verifier-free online assertion comparison.

\begin{conpactprompt}{Independent verifier system prompt}
\begin{lstlisting}[style=conpactprompttext]
Independently review the candidate using ONLY the student input
shown here. Reject incorrect current-state claims, unjustified
subgoals/actions, unsupported corrections, incorrect handback,
and any claim requiring a future or unavailable observation.
Earlier discussion may contain intentionally incorrect
assertions; judge the final reply independently. A
schema-permitted unknown/null is a valid abstention only when the
shown observation does not support a definite reading. Reject
unsupported abstention or guesses. Agreement or absence of
conflict alone does not establish correctness. A candidate action
has not yet executed. Return exactly JSON with accept:boolean and
reason:string.
\end{lstlisting}
\end{conpactprompt}

\clearpage
\section{Additional Results}
\label{app:additional-results}

\FloatBarrier
\subsection{Inference-Time Results}
\label{app:inference-results}

Table~\ref{tab:motivation-inference-results} reports the complete inference-time results underlying Table~\ref{tab:inference-core-results}, including all task metrics and both single-model ReAct references.

\begin{table*}[htbp]
\centering
\normalsize
\caption{Inference-time task performance across four environments. Each block specifies the planner--actor pair. Gray ReAct rows use only the indicated role's model and serve as single-model references, excluded from ranking. Within each block, \textbf{bold} and \underline{underlining} mark the best and second-best distinct values among Plan-Act, HiPlan, TAPE, and ConPAct-I, including ties. Higher is better except for Steps (lower is better).}
\label{tab:motivation-inference-results}
\renewcommand{\TableCols}{13}
\setlength{\tabcolsep}{2pt}
\renewcommand{\arraystretch}{1.2}
\def\TableBody{%
\toprule
\multirow{2}{*}{Method} & \multicolumn{3}{Ic}{Sokoban} & \multicolumn{3}{Ic}{Crafter} & \multicolumn{3}{Ic}{Procgen} & \multicolumn{3}{Ic}{MiniGrid} \\
\cmidrule(lr){2-4} \cmidrule(lr){5-7} \cmidrule(lr){8-10} \cmidrule(lr){11-13}
 & \multicolumn{1}{IE}{avg@8} & pass@8 & pass$^8$ & \multicolumn{1}{IE}{Score} & Progress & Return & \multicolumn{1}{IE}{IQM} & Mean & Median & \multicolumn{1}{IE}{SR} & Return & Steps \\
\midrule
\GroupHead{Opus-5 $\rightarrow$ Sonnet-5} \\
\textcolor{black!40}{ReAct (Planner)} & \textcolor{black!40}{95.4} & \textcolor{black!40}{99.5} & \textcolor{black!40}{80.5} & \textcolor{black!40}{6.71} & \textcolor{black!40}{28.4} & \textcolor{black!40}{6.97} & \textcolor{black!40}{32.75} & \textcolor{black!40}{35.19} & \textcolor{black!40}{32.86} & \textcolor{black!40}{60.0} & \textcolor{black!40}{0.49} & \textcolor{black!40}{10.1} \\
\textcolor{black!40}{ReAct (Actor)} & \textcolor{black!40}{70.9} & \textcolor{black!40}{89.0} & \textcolor{black!40}{40.5} & \textcolor{black!40}{4.74} & \textcolor{black!40}{23.8} & \textcolor{black!40}{4.81} & \textcolor{black!40}{20.55} & \textcolor{black!40}{21.47} & \textcolor{black!40}{18.35} & \textcolor{black!40}{33.4} & \textcolor{black!40}{0.28} & \textcolor{black!40}{11.4} \\
\midrule
Plan-Act & 84.2 & 94.0 & 71.0 & \underline{5.26} & \underline{23.3} & 5.05 & 23.49 & 32.14 & \underline{23.50} & \underline{33.1} & \underline{0.27} & 10.5 \\
HiPlan & \underline{91.3} & \textbf{99.5} & 73.5 & 5.23 & 23.1 & \underline{5.54} & \underline{25.26} & \textbf{34.85} & 15.07 & 31.6 & \underline{0.27} & 13.2 \\
TAPE & 90.5 & 97.0 & \underline{82.0} & 5.17 & 20.3 & 5.37 & 21.25 & \underline{34.83} & 22.86 & 29.2 & 0.23 & \underline{8.2} \\
\OursRow \textbf{ConPAct-I (ours)} & \textbf{95.4} & \underline{98.0} & \textbf{83.0} & \textbf{5.91} & \textbf{24.6} & \textbf{6.13} & \textbf{25.75} & 33.17 & \textbf{25.86} & \textbf{36.7} & \textbf{0.30} & \textbf{7.7} \\
\midrule
\GroupHead{GPT-5.6-sol $\rightarrow$ GPT-5.6-terra} \\
\textcolor{black!40}{ReAct (Planner)} & \textcolor{black!40}{83.8} & \textcolor{black!40}{97.0} & \textcolor{black!40}{59.5} & \textcolor{black!40}{6.07} & \textcolor{black!40}{25.2} & \textcolor{black!40}{6.18} & \textcolor{black!40}{23.20} & \textcolor{black!40}{27.24} & \textcolor{black!40}{22.32} & \textcolor{black!40}{54.5} & \textcolor{black!40}{0.50} & \textcolor{black!40}{10.9} \\
\textcolor{black!40}{ReAct (Actor)} & \textcolor{black!40}{66.2} & \textcolor{black!40}{91.5} & \textcolor{black!40}{34.0} & \textcolor{black!40}{5.10} & \textcolor{black!40}{25.1} & \textcolor{black!40}{5.00} & \textcolor{black!40}{17.21} & \textcolor{black!40}{18.42} & \textcolor{black!40}{14.24} & \textcolor{black!40}{31.4} & \textcolor{black!40}{0.25} & \textcolor{black!40}{12.1} \\
\midrule
Plan-Act & 75.1 & 89.5 & 54.0 & 5.13 & 24.2 & 5.50 & 17.20 & 17.38 & 17.96 & 38.6 & 0.35 & 11.8 \\
HiPlan & \underline{76.7} & \underline{93.0} & \underline{56.5} & 5.13 & \underline{24.4} & 4.91 & 17.95 & \underline{20.57} & 9.64 & \underline{47.7} & \underline{0.42} & \underline{10.4} \\
TAPE & 75.9 & 81.5 & 43.5 & \textbf{5.24} & 23.3 & \underline{5.61} & \underline{21.12} & 19.73 & \underline{18.70} & 31.2 & 0.29 & \textbf{6.0} \\
\OursRow \textbf{ConPAct-I (ours)} & \textbf{86.6} & \textbf{97.5} & \textbf{60.5} & \underline{5.14} & \textbf{24.7} & \textbf{5.65} & \textbf{22.83} & \textbf{24.33} & \textbf{19.78} & \textbf{54.4} & \textbf{0.47} & 13.0 \\
\midrule
\GroupHead{Kimi K3 $\rightarrow$ Kimi K2.6} \\
\textcolor{black!40}{ReAct (Planner)} & \textcolor{black!40}{68.9} & \textcolor{black!40}{88.0} & \textcolor{black!40}{40.5} & \textcolor{black!40}{4.73} & \textcolor{black!40}{18.1} & \textcolor{black!40}{4.31} & \textcolor{black!40}{13.11} & \textcolor{black!40}{20.44} & \textcolor{black!40}{12.82} & \textcolor{black!40}{42.3} & \textcolor{black!40}{0.38} & \textcolor{black!40}{9.8} \\
\textcolor{black!40}{ReAct (Actor)} & \textcolor{black!40}{45.7} & \textcolor{black!40}{82.5} & \textcolor{black!40}{12.0} & \textcolor{black!40}{3.39} & \textcolor{black!40}{17.6} & \textcolor{black!40}{3.82} & \textcolor{black!40}{11.23} & \textcolor{black!40}{14.70} & \textcolor{black!40}{7.83} & \textcolor{black!40}{21.9} & \textcolor{black!40}{0.19} & \textcolor{black!40}{8.4} \\
\midrule
Plan-Act & 57.7 & \textbf{88.5} & 43.0 & \textbf{4.23} & \textbf{20.2} & \underline{4.10} & 10.85 & 13.71 & 8.20 & \underline{28.3} & \underline{0.25} & \textbf{12.4} \\
HiPlan & \underline{65.5} & \underline{85.5} & \underline{45.0} & 3.29 & 16.8 & 3.04 & 11.85 & 13.85 & \underline{10.98} & 27.1 & 0.23 & 14.2 \\
TAPE & 53.6 & 84.5 & 33.5 & 3.82 & 15.3 & 3.51 & \underline{12.15} & \underline{15.02} & 8.27 & 21.5 & 0.17 & 15.8 \\
\OursRow \textbf{ConPAct-I (ours)} & \textbf{65.8} & 85.0 & \textbf{47.0} & \underline{4.16} & \underline{18.1} & \textbf{4.32} & \textbf{12.79} & \textbf{20.49} & \textbf{11.86} & \textbf{34.2} & \textbf{0.29} & \underline{12.9} \\
\midrule
\GroupHead{Qwen3.6-27B $\rightarrow$ Qwen3.5-9B} \\
\textcolor{black!40}{ReAct (Planner)} & \textcolor{black!40}{22.6} & \textcolor{black!40}{67.0} & \textcolor{black!40}{1.5} & \textcolor{black!40}{3.72} & \textcolor{black!40}{17.2} & \textcolor{black!40}{3.69} & \textcolor{black!40}{5.58} & \textcolor{black!40}{6.82} & \textcolor{black!40}{2.48} & \textcolor{black!40}{22.8} & \textcolor{black!40}{0.20} & \textcolor{black!40}{10.0} \\
\textcolor{black!40}{ReAct (Actor)} & \textcolor{black!40}{6.2} & \textcolor{black!40}{30.5} & \textcolor{black!40}{0.0} & \textcolor{black!40}{3.14} & \textcolor{black!40}{12.5} & \textcolor{black!40}{3.07} & \textcolor{black!40}{2.29} & \textcolor{black!40}{5.31} & \textcolor{black!40}{1.00} & \textcolor{black!40}{7.9} & \textcolor{black!40}{0.06} & \textcolor{black!40}{8.2} \\
\midrule
Plan-Act & 12.0 & \underline{43.0} & \underline{1.0} & 3.27 & \underline{16.0} & \textbf{3.34} & 3.19 & \underline{3.86} & \textbf{3.89} & \textbf{13.9} & \underline{0.10} & \underline{13.1} \\
HiPlan & \underline{13.3} & 35.0 & \textbf{1.5} & \underline{3.32} & 15.1 & 3.13 & \underline{4.76} & 3.65 & \underline{2.89} & 10.6 & 0.08 & 13.5 \\
TAPE & 10.3 & 32.0 & 0.5 & 3.01 & 15.0 & 3.20 & 2.11 & 1.96 & 0.18 & 8.6 & 0.07 & \textbf{11.1} \\
\OursRow \textbf{ConPAct-I (ours)} & \textbf{15.7} & \textbf{48.0} & \textbf{1.5} & \textbf{3.46} & \textbf{16.3} & \underline{3.27} & \textbf{4.85} & \textbf{6.28} & 2.14 & \underline{13.5} & \textbf{0.12} & 13.5 \\
\bottomrule
}
\makebox[\linewidth][c]{\resizebox{\linewidth}{!}{%
\begin{tabular}{lIEEEIEEEIEEEIEEE}
\TableBody
\end{tabular}}}
\end{table*}

\clearpage
\subsection{Supervised-Training Results}
\label{app:supervised-training-results}

Table~\ref{tab:motivation-training-results} reports the complete supervised-training results underlying Table~\ref{tab:training-core-results}, including all task metrics for the three planner--actor configurations.

\begin{table*}[htbp]
\centering
\normalsize
\caption{Task performance after supervised training across four environments. Each block specifies the planner--actor model pair. All methods, including ReAct SFT, are ranked within each block; \textbf{bold} and \underline{underlining} mark the best and second-best distinct values, including ties. Higher is better except for Steps (lower is better).}
\label{tab:motivation-training-results}
\renewcommand{\TableCols}{13}
\setlength{\tabcolsep}{2pt}
\renewcommand{\arraystretch}{1.2}
\def\TableBody{%
\toprule
\multirow{2}{*}{Method} & \multicolumn{3}{Ic}{Sokoban} & \multicolumn{3}{Ic}{Crafter} & \multicolumn{3}{Ic}{Procgen} & \multicolumn{3}{Ic}{MiniGrid} \\
\cmidrule(lr){2-4} \cmidrule(lr){5-7} \cmidrule(lr){8-10} \cmidrule(lr){11-13}
 & \multicolumn{1}{IE}{avg@8} & pass@8 & pass$^8$ & \multicolumn{1}{IE}{Score} & Progress & Return & \multicolumn{1}{IE}{IQM} & Mean & Median & \multicolumn{1}{IE}{SR} & Return & Steps \\
\midrule
\GroupHead{Qwen3-VL-8B $\rightarrow$ Qwen3-VL-8B} \\
ReAct SFT & 44.4 & 82.0 & 8.5 & 4.24 & 19.1 & 4.13 & 4.50 & \textbf{7.94} & 2.50 & 8.8 & 0.04 & 14.4 \\
Plan-Act SFT & 35.6 & 81.0 & 5.0 & 4.06 & 18.0 & 4.21 & 3.50 & 7.43 & \textbf{5.65} & 7.8 & 0.03 & \underline{11.4} \\
PAA & 43.8 & 80.5 & 12.0 & 4.33 & 16.8 & 4.58 & 3.15 & 4.46 & 1.25 & 10.8 & 0.05 & 15.4 \\
HSL & 44.4 & 81.5 & \underline{13.0} & 4.39 & 18.5 & 3.95 & 4.82 & 5.86 & 0.06 & 8.0 & 0.03 & \textbf{11.2} \\
WebSTAR & 44.7 & 78.5 & 12.0 & 4.28 & 20.4 & 4.42 & \underline{5.41} & 4.66 & 0.86 & 8.7 & 0.04 & 12.0 \\
ECoT & 34.4 & 86.5 & 12.5 & \underline{4.77} & 19.4 & 4.16 & 2.56 & 2.47 & 3.15 & 10.2 & 0.06 & 12.3 \\
\OursRow \textbf{ConPAct-S (ours)} & \underline{48.6} & \textbf{94.0} & \textbf{14.5} & \textbf{5.16} & \textbf{23.4} & \textbf{5.08} & 3.22 & \underline{7.75} & 4.85 & \textbf{27.0} & \textbf{0.23} & 14.4 \\
\OursRow \textbf{ConPAct-R (ours)} & \textbf{49.1} & \underline{92.5} & 11.0 & 4.72 & \underline{22.7} & \underline{4.92} & \textbf{6.38} & 4.86 & \underline{5.27} & \underline{24.8} & \underline{0.21} & 13.7 \\
\midrule
\GroupHead{Qwen3-VL-32B $\rightarrow$ Qwen3-VL-32B} \\
ReAct SFT & 60.1 & 86.0 & 26.5 & 4.63 & 21.6 & 4.67 & 6.88 & 9.15 & 5.00 & 18.5 & 0.14 & 14.4 \\
Plan-Act SFT & 50.7 & 88.5 & 12.5 & 3.10 & 16.6 & 3.55 & 6.35 & \underline{9.53} & 2.10 & 15.8 & 0.10 & 14.0 \\
PAA & 65.4 & 88.0 & 29.0 & 3.76 & 17.9 & 3.84 & \underline{8.27} & \textbf{9.70} & 6.63 & 11.6 & 0.06 & \textbf{13.3} \\
HSL & 61.3 & 88.0 & 32.0 & 4.23 & 20.3 & 4.35 & 5.85 & 8.27 & 2.50 & 16.5 & 0.10 & 16.1 \\
WebSTAR & 66.2 & 90.0 & \underline{35.5} & 4.96 & 23.8 & 5.15 & 7.26 & 7.81 & 2.50 & 13.1 & 0.07 & 14.9 \\
ECoT & 57.2 & \underline{93.0} & 17.5 & \underline{5.23} & 20.8 & 4.47 & 5.25 & 6.78 & 0.60 & 29.1 & 0.23 & 15.5 \\
\OursRow \textbf{ConPAct-S (ours)} & \textbf{68.4} & 92.0 & 34.0 & 5.05 & \underline{23.9} & \underline{5.20} & \textbf{9.12} & 7.96 & \underline{7.61} & \textbf{39.7} & \textbf{0.35} & \underline{13.5} \\
\OursRow \textbf{ConPAct-R (ours)} & \underline{66.9} & \textbf{94.5} & \textbf{38.5} & \textbf{5.55} & \textbf{24.1} & \textbf{5.23} & 6.35 & 5.17 & \textbf{8.32} & \underline{33.9} & \underline{0.30} & \textbf{13.3} \\
\midrule
\GroupHead{Qwen3-VL-32B $\rightarrow$ Qwen3-VL-8B} \\
Plan-Act SFT & 45.4 & 87.0 & 6.5 & 2.85 & 15.1 & 3.21 & 5.60 & \underline{6.75} & \underline{6.07} & 11.9 & 0.06 & 14.0 \\
PAA & \underline{49.4} & 85.5 & \textbf{16.0} & 3.40 & 17.6 & 3.79 & \underline{5.89} & \textbf{7.78} & 0.59 & 8.7 & 0.04 & \underline{13.5} \\
ECoT & 41.2 & 79.0 & 12.0 & 4.70 & 19.5 & 4.17 & 2.62 & 2.48 & 3.00 & 12.7 & 0.07 & 14.3 \\
\OursRow \textbf{ConPAct-S (ours)} & 48.5 & \underline{87.5} & 6.5 & \textbf{5.53} & \textbf{23.6} & \textbf{5.10} & 4.56 & 3.56 & 2.68 & \textbf{28.2} & \textbf{0.24} & \textbf{13.2} \\
\OursRow \textbf{ConPAct-R (ours)} & \textbf{54.8} & \textbf{89.5} & \underline{15.0} & \underline{5.26} & \underline{23.5} & \underline{5.09} & \textbf{7.36} & 5.82 & \textbf{6.79} & \underline{27.8} & \underline{0.23} & 13.8 \\
\bottomrule
}
\makebox[\linewidth][c]{\resizebox{\linewidth}{!}{%
\begin{tabular}{lIEEEIEEEIEEEIEEE}
\TableBody
\end{tabular}}}
\end{table*}

\FloatBarrier
\subsection{OSWorld Results}
\label{app:osworld-results}

We use OSWorld 1.0 and evaluate each of its 361 tasks with three rollouts, covering the ten application categories in Table~\ref{tab:osworld-full}.
Each model is compared under ReAct, Plan-Act, and ConPAct-I with a three-observation history limit.
The desktop action interface accepts a PyAutoGUI script or a \texttt{WAIT}, \texttt{DONE}, or \texttt{FAIL} command; the current task and screenshot are supplied to the agent.
Request or infrastructure failures permit up to three retries within an evaluation rollout; these retries are separate from the three scheduled rollouts per task.
A completed but unsuccessful rollout remains unsuccessful.
For each task, Score averages evaluation scores and SR averages full-success indicators across its three rollouts; both are reported in percent.
Overall results average these task-level values with equal task weights rather than weighting the ten categories equally.
The category breakdown supplements the overall gains in Figure~\ref{fig:osworld_gains} and shows that individual categories need not improve uniformly.

\begin{table*}[htbp]
\centering
\small
\caption{Full OSWorld results, averaged over three rollouts per task. Each entry is Score/SR (\%; $\uparrow$), rounded to one decimal place. Bold marks the best value for each metric within each model and category, including ties, using the values before rounding.}
\label{tab:osworld-full}
\setlength{\tabcolsep}{1.5pt}
\renewcommand{\arraystretch}{1.2}
\resizebox{\linewidth}{!}{%
\begin{tabular}{@{\hspace{2.5pt}}l@{\hspace{5pt}}l@{\hspace{5pt}}*{11}{c}@{\hspace{2.5pt}}}
\toprule
Model & Method & Chrome & GIMP & Calc & Impress & Writer & \makecell{Multi-\\apps} & OS & \makecell{Thunder-\\bird} & VLC & \makecell{VS\\Code} & Overall \\
\midrule
\multirow{3}{*}{Opus-5} & ReAct & 71.0/69.3 & 87.2/87.2 & 80.6/80.6 & 83.9/82.3 & 89.2/85.6 & 81.4/75.0 & \textbf{82.7/82.7} & 83.4/83.4 & 94.1/90.2 & 82.0/82.0 & 82.0/79.5 \\
 & Plan-Act & 70.3/68.5 & \textbf{88.0/88.0} & 81.9/81.9 & 84.8/83.3 & 88.7/84.9 & 81.9/75.7 & 81.7/81.7 & \textbf{84.1/84.1} & 93.4/89.1 & 80.0/80.0 & 82.2/79.7 \\
\rowcolor{OursHighlight!20}[2.5pt][2.5pt]
 & \textbf{ConPAct-I} & \textbf{71.7/70.1} & 86.5/86.5 & \textbf{82.3/82.3} & \textbf{85.3/83.8} & \textbf{90.3/87.1} & \textbf{82.6/76.7} & 81.5/81.5 & 82.7/82.7 & \textbf{94.6/91.1} & \textbf{84.4/84.4} & \textbf{82.9/80.6} \\
\midrule
\multirow{3}{*}{GPT-5.6-sol} & ReAct & 82.6/82.6 & \textbf{80.8/80.8} & 78.7/78.7 & 83.0/83.0 & 78.2/69.6 & 83.5/78.5 & \textbf{87.5/87.5} & \textbf{86.7/86.7} & 86.8/\textbf{82.4} & \textbf{91.3/91.3} & 83.2/81.2 \\
 & Plan-Act & \textbf{84.8/84.8} & 76.9/76.9 & 78.7/78.7 & \textbf{85.1/85.1} & 77.3/69.6 & 83.6/78.5 & 83.3/83.3 & \textbf{86.7/86.7} & 87.1/\textbf{82.4} & 87.0/87.0 & 82.9/80.9 \\
\rowcolor{OursHighlight!20}[2.5pt][2.5pt]
 & \textbf{ConPAct-I} & 82.6/82.6 & 76.9/76.9 & \textbf{80.9/80.9} & \textbf{85.1/85.1} & \textbf{81.7/73.9} & \textbf{85.9/80.7} & 83.3/83.3 & \textbf{86.7/86.7} & \textbf{88.2/82.4} & \textbf{91.3/91.3} & \textbf{84.1/82.0} \\
\midrule
\multirow{3}{*}{Kimi K3} & ReAct & \textbf{61.7/59.7} & 67.1/67.1 & 89.2/89.2 & 77.3/75.2 & 81.4/76.7 & 78.7/70.5 & \textbf{80.4/80.4} & 85.7/85.7 & \textbf{76.7/72.1} & 85.0/85.0 & 77.8/74.6 \\
 & Plan-Act & 60.2/58.2 & \textbf{70.0/70.0} & 90.1/90.1 & 77.9/76.0 & \textbf{83.1/78.9} & 78.4/70.1 & 77.3/77.3 & \textbf{87.3/87.3} & 72.7/67.4 & 84.6/84.6 & 77.7/74.5 \\
\rowcolor{OursHighlight!20}[2.5pt][2.5pt]
 & \textbf{ConPAct-I} & 60.7/58.7 & 69.2/69.2 & \textbf{91.5/91.5} & \textbf{78.5/76.6} & 82.6/78.3 & \textbf{79.8/72.0} & 79.2/79.2 & 86.7/86.7 & 75.4/70.6 & \textbf{87.0/87.0} & \textbf{78.6/75.6} \\
\midrule
\multirow{3}{*}{Kimi K2.6} & ReAct & \textbf{71.7/69.6} & 73.1/73.1 & 80.9/80.9 & 82.2/78.7 & 73.9/69.6 & 55.0/49.5 & \textbf{79.2/79.2} & 80.0/80.0 & \textbf{75.7/70.6} & 91.3/91.3 & 72.5/69.8 \\
 & Plan-Act & 70.9/68.8 & 73.9/73.9 & 81.9/81.9 & 82.5/79.0 & 73.0/68.5 & 55.4/49.9 & 77.9/77.9 & \textbf{80.7/80.7} & 74.8/69.4 & 89.4/89.4 & 72.4/69.8 \\
\rowcolor{OursHighlight!20}[2.5pt][2.5pt]
 & \textbf{ConPAct-I} & 71.4/69.2 & \textbf{74.4/74.4} & \textbf{82.2/82.2} & \textbf{83.0/79.7} & \textbf{75.0/70.9} & \textbf{56.2/50.8} & 78.1/78.1 & 79.0/79.0 & 75.2/70.0 & \textbf{93.9/93.9} & \textbf{73.2/70.6} \\
\bottomrule
\end{tabular}%
}
\end{table*}

\FloatBarrier
\subsection{Inference Ablations}
\label{app:inference-ablations}

\textbf{Configurations.}
Table~\ref{tab:ablation-task-results} compares four settings, from left to right.
Correction-disabled retains assertion reporting but removes system-triggered correction.
No-specific-feedback uses a generic review request without conflicting fields; both roles revise.
Actor-only provides conflict feedback but fixes the planner's assertion and sub-goal during correction.
Full ConPAct-I provides conflicting fields and allows both roles to revise.
Correction permits up to three rounds without executing candidate actions: two calls per round, or one for actor-only.

\textbf{Controls.}
Assertion adds state reports to Plan-Act with no extra calls or correction.
Resampling uses the same inputs and redraws the planner--actor pair after a conflict, without feedback or discarded candidates.
Each round adds two calls, up to three rounds; it retains the first agreeing pair or the last complete pair at the limit.
Self-Refine~\citep{madaan2023selfrefine} uses one ReAct model to draft, critique, and revise on unchanged observations.
Critique and revision add two calls per action; only the revised action executes.

\textbf{Event metrics.}
Initial and final $D$ use the same denominator: complete same-observation events with valid, definite assertion pairs before and after correction.
Their numerators count initial and final mismatches, respectively.
Repair is the fraction of initially mismatched events that end in agreement.
Unpaired, invalid, or interrupted events are excluded; empty denominators are undefined.
With correction disabled, final $D$ equals initial $D$; the displayed zero denotes no system repair.

\textbf{Budgets.}
Following the main-text experimental setup, each evaluation instance has eight rollouts, with equal action and total model-call limits: 60 for Sokoban and the task-specific 60--200 limits for MiniGrid.
All decision calls, including resampling, critique, and revision, count toward the shared call limit.
Equal limits do not imply equal token use.

\textbf{Results.}
In Tables~\ref{tab:ablation-task-results} and~\ref{tab:self-refine-comparison}, full ConPAct-I leads all Sokoban success metrics and MiniGrid SR across the reported configurations.
Repair reaches 96.1--98.8\%, compared with 86.7--91.8\% without specific feedback and 27.9--37.7\% with actor-only correction.
MiniGrid Return and successful-episode Steps do not improve uniformly.
The tables include the MiniGrid results and secondary metrics omitted from the main figures.

\begin{table*}[htbp]
\centering
\normalsize
\caption{Effects of online consistency ablations on task performance and planner--actor state mismatch.}
\label{tab:ablation-task-results}
\renewcommand{\TableCols}{6}
\setlength{\tabcolsep}{5pt}
\renewcommand{\arraystretch}{1.2}
\def\AblationCheck{\ding{51}}
\def\AblationCross{\textcolor{black!25}{\ding{55}}}
\begin{tabular}{@{}llIEEEE@{}}
\toprule
\multirow{3}{*}{\makecell[l]{\textbf{Configuration}}}
 & Correction Trigger & \AblationCross & \AblationCheck & \AblationCheck & \AblationCheck \\
 & Conflict FeedBack & \AblationCross & \AblationCross & \AblationCheck & \AblationCheck \\
 & Planner Participation & \AblationCross & \AblationCheck & \AblationCross & \AblationCheck \\
\midrule
\GroupHead{GPT-5.6-sol $\rightarrow$ GPT-5.6-sol} \\
\multirow{6}{*}{\textbf{Sokoban}}
 & avg@8 & 85.0 & 90.8 & 89.8 & 92.1 \\
 & pass@8 & 96.0 & 97.5 & 97.0 & 98.0 \\
 & pass$^8$ & 71.0 & 81.5 & 78.0 & 82.5 \\
 & Initial D & 25.6 & 26.1 & 25.9 & 25.5 \\
 & Final D & 25.6 & 3.5 & 18.7 & 0.6 \\
 & Repair rate & 0.0 & 86.7 & 27.9 & 97.8 \\
\cmidrule(lr){1-6}
\multirow{6}{*}{\textbf{MiniGrid}}
 & SR & 53.1 & 60.5 & 61.9 & 62.6 \\
 & Return & 0.42 & 0.56 & 0.55 & 0.55 \\
 & Steps & 13.2 & 12.5 & 12.3 & 12.6 \\
 & Initial D & 12.4 & 12.7 & 12.6 & 12.7 \\
 & Final D & 12.4 & 1.0 & 8.1 & 0.2 \\
 & Repair rate & 0.0 & 91.8 & 35.8 & 98.8 \\
\midrule
\GroupHead{GPT-5.6-sol $\rightarrow$ GPT-5.6-terra} \\
\multirow{6}{*}{\textbf{Sokoban}}
 & avg@8 & 75.1 & 83.2 & 83.4 & 86.6 \\
 & pass@8 & 89.5 & 97.5 & 96.5 & 97.5 \\
 & pass$^8$ & 54.0 & 58.5 & 53.0 & 60.5 \\
 & Initial D & 57.3 & 57.3 & 58.5 & 58.0 \\
 & Final D & 57.3 & 5.3 & 36.5 & 1.2 \\
 & Repair rate & 0.0 & 90.8 & 37.7 & 98.0 \\
\cmidrule(lr){1-6}
\multirow{6}{*}{\textbf{MiniGrid}}
 & SR & 38.6 & 45.7 & 46.8 & 54.4 \\
 & Return & 0.35 & 0.41 & 0.42 & 0.47 \\
 & Steps & 11.8 & 14.4 & 14.4 & 13.0 \\
 & Initial D & 24.9 & 24.6 & 24.7 & 24.8 \\
 & Final D & 24.9 & 3.1 & 16.8 & 1.0 \\
 & Repair rate & 0.0 & 87.2 & 32.3 & 96.1 \\
\bottomrule
\end{tabular}
\end{table*}

\begin{table*}[htbp]
\centering
\normalsize
\caption{Inference controls on Sokoban and MiniGrid.}
\label{tab:self-refine-comparison}
\renewcommand{\TableCols}{7}
\setlength{\tabcolsep}{7pt}
\renewcommand{\arraystretch}{1.2}
\begin{tabular}{lIEEEIEEE}
\toprule
\multirow{2}{*}{Method} & \multicolumn{3}{Ic}{Sokoban} & \multicolumn{3}{Ic}{MiniGrid} \\
\cmidrule(lr){2-4} \cmidrule(lr){5-7}
 & \multicolumn{1}{IE}{avg@8} & pass@8 & pass$^8$ & \multicolumn{1}{IE}{SR} & Return & Steps \\
\midrule
\GroupHead{GPT-5.6-sol} \\
ReAct & 83.8 & 97.0 & 59.5 & 54.5 & 0.50 & 10.9 \\
Self-Refine & 86.5 & 93.5 & 73.5 & 51.2 & 0.49 & 7.2 \\
Plan-Act & 79.4 & 91.5 & 52.5 & 53.1 & 0.46 & 13.1 \\
Assertion & 82.8 & 92.5 & 55.5 & 57.7 & 0.51 & 13.3 \\
Resampling & 81.1 & 95.0 & 57.5 & 56.0 & 0.50 & 13.3 \\
\OursRow \textbf{ConPAct-I (ours)} & 92.1 & 98.0 & 82.5 & 62.6 & 0.55 & 12.6 \\
\midrule
\GroupHead{GPT-5.6-terra} \\
ReAct & 66.2 & 91.5 & 34.0 & 31.4 & 0.25 & 12.1 \\
Self-Refine & 64.1 & 85.5 & 37.0 & 47.4 & 0.44 & 8.1 \\
Plan-Act & 67.0 & 87.5 & 39.0 & 41.7 & 0.34 & 11.7 \\
Assertion & 65.5 & 91.0 & 37.0 & 45.3 & 0.38 & 13.3 \\
Resampling & 64.9 & 88.0 & 42.0 & 48.4 & 0.41 & 13.7 \\
\OursRow \textbf{ConPAct-I (ours)} & 80.5 & 97.0 & 56.0 & 51.4 & 0.44 & 14.5 \\
\midrule
\GroupHead{GPT-5.6-sol $\rightarrow$ GPT-5.6-terra} \\
Plan-Act & 75.1 & 89.5 & 54.0 & 38.6 & 0.35 & 11.8 \\
Assertion & 76.1 & 90.0 & 56.5 & 39.4 & 0.32 & 15.0 \\
Resampling & 74.9 & 88.0 & 52.0 & 36.1 & 0.29 & 13.9 \\
\OursRow \textbf{ConPAct-I (ours)} & 86.6 & 97.5 & 60.5 & 54.4 & 0.47 & 13.0 \\
\bottomrule
\end{tabular}
\end{table*}

\FloatBarrier
\subsection{Data Ablations}
\label{app:data-ablations}

We compare the four data-curation conditions defined in the main text: a baseline without the three curation components; removal of reconciliation supervision while retaining filtering and replay; removal of reconciliation replay while retaining filtering and supervision; and full ConPAct-R.
Table~\ref{tab:ablation-sft-data-8b} reports these component comparisons, with ordinary planner--actor supervision as the uncurated baseline.
\textbf{Without reconciliation supervision}, discussion remains in the input history but its responses receive no target loss.
\textbf{Without reconciliation replay}, the teacher does not construct the reconciliation sequences; ordinary planner--actor supervision is retained.
\textbf{Consistency filtering} acts on response targets: detected inconsistent replies can remain history without being learned as responses.
All conditions use the same optimization settings and a training-token budget matched to three epochs of ReAct SFT.
Evaluation uses model-initiated discussion without system-triggered ConPAct-I.
The table reports execution-level task performance and state mismatch; the next subsection separately measures initial judgments and conditional discussion outcomes.

\begin{table*}[htbp]
\centering
\small
\caption{Training-data ablations on Sokoban and MiniGrid across the three planner--actor configurations. No Sup. and No Replay retain consistency filtering; Full uses all three components.}
\label{tab:ablation-sft-data-8b}
\label{tab:ablation-sft-data-32b}
\label{tab:ablation-sft-data-32b-8b}
\setlength{\tabcolsep}{7pt}
\renewcommand{\arraystretch}{1.12}
\begin{tabular}{llrrrr}
\toprule
Environment & Metric & Baseline & No Sup. & No Replay & \textbf{Full} \\
\midrule
\multicolumn{6}{l}{\textit{Qwen3-VL-8B $\rightarrow$ Qwen3-VL-8B}} \\
\multirow{4}{*}{\textbf{Sokoban}} & $\mathrm{avg}@8$ & 40.7 & 44.4 & 43.5 & 49.1 \\
 & $\mathrm{pass}@8$ & 87.5 & 89.0 & 91.0 & 92.5 \\
 & $\mathrm{pass}^{8}$ & 6.5 & 9.0 & 5.5 & 11.0 \\
 & State mismatch & 46.7 & 34.7 & 39.8 & 12.9 \\
\multirow{4}{*}{\textbf{MiniGrid}} & SR & 18.0 & 20.0 & 11.6 & 24.8 \\
 & Return & 0.10 & 0.15 & 0.08 & 0.21 \\
 & Steps & 11.3 & 12.7 & 11.8 & 13.7 \\
 & State mismatch & 50.8 & 41.8 & 40.0 & 23.6 \\
\midrule
\multicolumn{6}{l}{\textit{Qwen3-VL-32B $\rightarrow$ Qwen3-VL-32B}} \\
\multirow{4}{*}{\textbf{Sokoban}} & $\mathrm{avg}@8$ & 55.9 & 62.6 & 54.2 & 66.9 \\
 & $\mathrm{pass}@8$ & 88.0 & 91.0 & 90.5 & 94.5 \\
 & $\mathrm{pass}^{8}$ & 23.5 & 20.5 & 13.0 & 38.5 \\
 & State mismatch & 34.7 & 27.0 & 29.9 & 1.4 \\
\multirow{4}{*}{\textbf{MiniGrid}} & SR & 18.8 & 24.2 & 23.0 & 33.9 \\
 & Return & 0.12 & 0.19 & 0.18 & 0.30 \\
 & Steps & 13.4 & 12.5 & 13.4 & 13.3 \\
 & State mismatch & 40.1 & 22.2 & 30.3 & 12.1 \\
\midrule
\multicolumn{6}{l}{\textit{Qwen3-VL-32B $\rightarrow$ Qwen3-VL-8B}} \\
\multirow{4}{*}{\textbf{Sokoban}} & $\mathrm{avg}@8$ & 44.4 & 45.2 & 36.1 & 54.8 \\
 & $\mathrm{pass}@8$ & 85.5 & 84.5 & 85.5 & 89.5 \\
 & $\mathrm{pass}^{8}$ & 10.5 & 12.0 & 6.5 & 15.0 \\
 & State mismatch & 35.9 & 25.1 & 29.6 & 9.5 \\
\multirow{4}{*}{\textbf{MiniGrid}} & SR & 11.0 & 13.4 & 16.7 & 27.8 \\
 & Return & 0.04 & 0.07 & 0.13 & 0.23 \\
 & Steps & 12.6 & 11.1 & 12.5 & 13.8 \\
 & State mismatch & 59.6 & 43.2 & 49.9 & 22.2 \\
\bottomrule
\end{tabular}
\end{table*}

\FloatBarrier
\subsection{Training Analysis}
\label{app:training-analysis}

We evaluate the homogeneous 8B and 32B configurations and the 32B-planner/8B-actor configuration on Sokoban and MiniGrid, as in the main text.
The analyses use the same evaluation tasks and align pre/post judgments by observation within each event.
The baseline is the no-curation training condition used in the component comparison.
All discussion is initiated by the trained models; the analysis does not insert ConPAct-I's automatic correction loop.

\textbf{Initial state judgments.}
The initial comparison uses the independent planner--actor reports before discussion.
Initial D is the fraction of valid initial pairs that disagree, while each role's correctness is its exact-match accuracy against ground truth on the eligible paired observations.
Invalid reports are excluded; correctness additionally requires available ground truth.
Table~\ref{tab:posthoc-b1-state-8b} compare these initial measurements with ConPAct-R.

\textbf{Discussion and residual mismatch.}
Discussion rate is the fraction of initially mismatched events in which the actor initiates discussion.
Residual D measures the fraction of those discussed disagreements that remain mismatched after discussion, restricted to events with a valid final pair.
Thus, initial D and residual D have different conditioning sets: one describes all valid initial judgments, while the other describes the disagreements on which discussion occurred.
Table~\ref{tab:posthoc-discussion-response} compares full training with removal of reconciliation supervision or replay.

\textbf{Paired state composition.}
The composition analysis uses events with valid assertions and ground truth both before and after the interaction.
Each entire assertion pair belongs to one category: CC if both reports agree with ground truth, CW if they agree with each other but are incorrect, or D if the reports disagree.
The same eligible events are used for initial percentages and signed post-discussion changes in Table~\ref{tab:posthoc-state-distribution}.
This paired, ground-truth-eligible subset can differ from the initial-only and discussion-conditional sets; their mismatch percentages are not interchangeable.
The analysis distinguishes correct agreement from agreement alone, complementing the task-level outcomes and individual-role correctness.

\begin{table*}[htbp]
\centering
\small
\caption{Initial state mismatch and assertion correctness before discussion. Rates are percentages and changes are in percentage points.}
\label{tab:posthoc-b1-state-8b}
\label{tab:posthoc-b1-state-32b}
\label{tab:posthoc-b1-state-32b-8b}
\setlength{\tabcolsep}{7pt}
\renewcommand{\arraystretch}{1.12}
\begin{tabular}{llrrr}
\toprule
Environment & Metric & Baseline & \textbf{ConPAct-R} & $\Delta$ \\
\midrule
\multicolumn{5}{l}{\textit{Qwen3-VL-8B $\rightarrow$ Qwen3-VL-8B}} \\
\multirow{3}{*}{\textbf{Sokoban}} & Initial D & 47.4 & 20.8 & $-26.6$ \\
 & Planner correctness & 45.5 & 83.0 & $+37.5$ \\
 & Actor correctness & 46.6 & 83.1 & $+36.5$ \\
\multirow{3}{*}{\textbf{MiniGrid}} & Initial D & 51.3 & 29.9 & $-21.4$ \\
 & Planner correctness & 50.9 & 81.5 & $+30.6$ \\
 & Actor correctness & 50.73 & 82.4 & $+31.7$ \\
\midrule
\multicolumn{5}{l}{\textit{Qwen3-VL-32B $\rightarrow$ Qwen3-VL-32B}} \\
\multirow{3}{*}{\textbf{Sokoban}} & Initial D & 35.4 & 10.5 & $-24.9$ \\
 & Planner correctness & 49.3 & 93.7 & $+44.4$ \\
 & Actor correctness & 50.3 & 94.1 & $+43.8$ \\
\multirow{3}{*}{\textbf{MiniGrid}} & Initial D & 41.2 & 18.7 & $-22.5$ \\
 & Planner correctness & 43.3 & 87.93 & $+44.6$ \\
 & Actor correctness & 42.1 & 88.08 & $+46.0$ \\
\midrule
\multicolumn{5}{l}{\textit{Qwen3-VL-32B $\rightarrow$ Qwen3-VL-8B}} \\
\multirow{3}{*}{\textbf{Sokoban}} & Initial D & 36.4 & 13.6 & $-22.8$ \\
 & Planner correctness & 42.9 & 89.4 & $+46.5$ \\
 & Actor correctness & 44.1 & 92.1 & $+48.0$ \\
\multirow{3}{*}{\textbf{MiniGrid}} & Initial D & 61.2 & 33.7 & $-27.5$ \\
 & Planner correctness & 33.3 & 77.9 & $+44.6$ \\
 & Actor correctness & 32.1 & 78.1 & $+46.0$ \\
\bottomrule
\end{tabular}
\end{table*}

\begin{table*}[htbp]
\centering
\normalsize
\caption{Discussion responses and planner--actor state mismatch (\%).}
\label{tab:posthoc-discussion-response}
\renewcommand{\TableCols}{8}
\setlength{\tabcolsep}{2pt}
\renewcommand{\arraystretch}{1.2}
\setlength{\MetricWidth}{\dimexpr(\linewidth-56pt-14\tabcolsep-0.8pt)/6\relax}
\def\AblationCheck{\ding{51}}
\def\AblationCross{\textcolor{black!25}{\ding{55}}}
\begin{tabular}{@{}*{2}{>{\centering\arraybackslash}p{28pt}}IEEEIEEE@{}}
\toprule
\multicolumn{2}{c}{Training data} & \multicolumn{3}{Ic}{Sokoban} & \multicolumn{3}{Ic}{MiniGrid} \\
\cmidrule(lr){1-2} \cmidrule(lr){3-5} \cmidrule(lr){6-8}
Sup. & Replay & Discuss. & Init. D & Resid. D & Discuss. & Init. D & Resid. D \\
\midrule
\GroupHead{Qwen3-VL-8B} \\
\AblationCross & \AblationCheck & 8.3 & 39.1 & 92.4 & 7.2 & 42.7 & 93.2 \\
\AblationCheck & \AblationCross & 21.7 & 45.9 & 90.1 & 19.4 & 42.2 & 87.2 \\
\OursRow \AblationCheck & \AblationCheck & 58.4 & 20.8 & 62.0 & 54.1 & 29.9 & 78.9 \\
\midrule
\GroupHead{Qwen3-VL-32B} \\
\AblationCross & \AblationCheck & 10.5 & 31.7 & 90.5 & 9.4 & 21.3 & 91.9 \\
\AblationCheck & \AblationCross & 26.8 & 35.9 & 88.0 & 24.2 & 35.6 & 87.2 \\
\OursRow \AblationCheck & \AblationCheck & 71.6 & 10.5 & 13.3 & 68.3 & 18.7 & 64.7 \\
\midrule
\GroupHead{Qwen3-VL-32B $\rightarrow$ Qwen3-VL-8B} \\
\AblationCross & \AblationCheck & 8.9 & 27.2 & 91.1 & 8.1 & 45.8 & 91.9 \\
\AblationCheck & \AblationCross & 23.1 & 36.7 & 87.2 & 21.5 & 61.7 & 87.1 \\
\OursRow \AblationCheck & \AblationCheck & 63.5 & 13.6 & 69.9 & 60.7 & 33.7 & 65.9 \\
\bottomrule
\end{tabular}
\end{table*}

\begin{table*}[htbp]
\centering
\normalsize
\caption{State-assertion consistency and correctness: initial values and signed changes after discussion.}
\label{tab:posthoc-state-distribution}
\renewcommand{\TableCols}{8}
\setlength{\tabcolsep}{2pt}
\renewcommand{\arraystretch}{1.2}
\setlength{\MetricWidth}{\dimexpr(\linewidth-56pt-14\tabcolsep-0.8pt)/6\relax}
\def\AblationCheck{\ding{51}}
\def\AblationCross{\textcolor{black!25}{\ding{55}}}
\begin{tabular}{@{}*{2}{>{\centering\arraybackslash}p{28pt}}IEEEIEEE@{}}
\toprule
\multicolumn{2}{c}{Training data} & \multicolumn{3}{Ic}{Sokoban} & \multicolumn{3}{Ic}{MiniGrid} \\
\cmidrule(lr){1-2} \cmidrule(lr){3-5} \cmidrule(lr){6-8}
Sup. & Replay & CC $\pm\Delta$ & CW $\pm\Delta$ & D $\pm\Delta$ & CC $\pm\Delta$ & CW $\pm\Delta$ & D $\pm\Delta$ \\
\midrule
\GroupHead{Qwen3-VL-8B} \\
\AblationCross & \AblationCheck & $40.7{+}1.5$ & $19.8{+}1.5$ & $39.5{-}3.0$ & $38.4{+}1.6$ & $17.9{+}1.5$ & $43.7{-}3.1$ \\
\AblationCheck & \AblationCross & $28.4{+}2.4$ & $28.0{+}2.0$ & $43.6{-}4.4$ & $36.2{+}3.3$ & $17.1{+}2.7$ & $46.7{-}6.0$ \\
\OursRow \AblationCheck & \AblationCheck & $72.6{+}5.1$ & $6.9{+}2.8$ & $20.5{-}7.9$ & $67.0{+}4.1$ & $4.0{+}2.2$ & $29.0{-}6.3$ \\
\midrule
\GroupHead{Qwen3-VL-32B} \\
\AblationCross & \AblationCheck & $44.8{+}1.5$ & $24.9{+}1.5$ & $30.4{-}3.0$ & $64.4{+}1.1$ & $10.3{+}1.1$ & $25.3{-}2.2$ \\
\AblationCheck & \AblationCross & $35.4{+}2.3$ & $31.3{+}1.8$ & $33.3{-}4.1$ & $38.2{+}2.5$ & $27.0{+}2.0$ & $34.8{-}4.5$ \\
\OursRow \AblationCheck & \AblationCheck & $88.7{+}4.8$ & $2.7{+}1.3$ & $8.6{-}6.1$ & $78.7{+}4.6$ & $4.5{+}2.0$ & $16.8{-}6.6$ \\
\midrule
\GroupHead{Qwen3-VL-32B $\rightarrow$ Qwen3-VL-8B} \\
\AblationCross & \AblationCheck & $48.0{+}1.2$ & $24.3{+}1.3$ & $27.8{-}2.5$ & $31.5{+}2.1$ & $20.4{+}2.2$ & $48.1{-}4.3$ \\
\AblationCheck & \AblationCross & $31.5{+}2.4$ & $34.1{+}2.0$ & $34.4{-}4.4$ & $10.3{+}4.1$ & $33.2{+}3.4$ & $56.5{-}7.5$ \\
\OursRow \AblationCheck & \AblationCheck & $84.0{+}2.7$ & $3.1{+}1.4$ & $12.9{-}4.1$ & $61.1{+}7.5$ & $4.3{+}4.0$ & $34.6{-}11.5$ \\
\bottomrule
\end{tabular}
\end{table*}

\clearpage
\section{Case Studies}
\label{app:case-studies}

\subsection{Sokoban}
\label{app:sokoban-reconciliation}

\begin{figure}[H]
  \centering
\begingroup
\small
\setlength{\parindent}{0pt}
\newcommand{\SokobanCell}[1]{%
  \begin{minipage}[t]{\linewidth}\raggedright
  #1
  \end{minipage}\par}
\newcommand{\SokobanRule}{%
  \vspace{5pt}{\color{OursHighlight!65!black}\hrule height 0.3pt}\vspace{6pt}}
\newcommand{\SokobanBoard}[1]{%
  {\centering
    #1\par}
  \vspace{4pt}}

\begin{tcolorbox}[
  enhanced, colback=OursHighlight!12, colframe=OursHighlight!65!black,
  boxrule=0.45pt, arc=1mm,
  left=7pt, right=7pt, top=7pt, bottom=7pt,
  before skip=0pt, after skip=5pt
]
\normalsize
\textbf{Sub-goal:} Move \textbf{Left}, then \textbf{Up} to stand immediately left of the box.
\end{tcolorbox}

\begin{minipage}[t]{0.485\linewidth}
\vspace{0pt}
\colorlet{OursHighlight}{black!35}
\begin{assertionexample}{State mismatch}
\small
\SokobanCell{%
  \SokobanBoard{\includegraphics[width=0.49\linewidth,trim={0bp 384bp 256bp 96bp},clip]{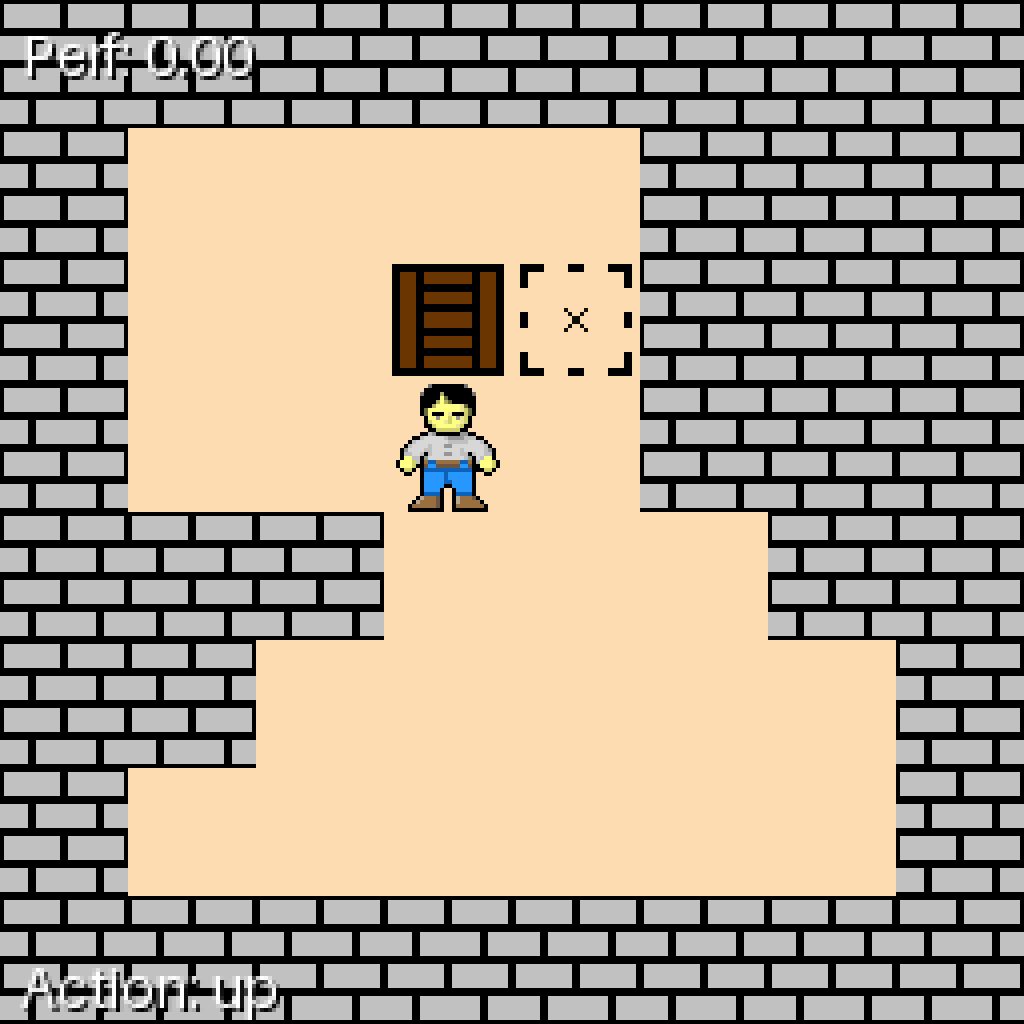}}
  \textbf{Planner:} ``The box is directly above you.''}
\SokobanRule
\SokobanCell{%
  \textbf{Actor:} ``Up is empty floor; Right is a wall. I will move Up to get a better view.''\par\smallskip
  \textbf{Action:} \texttt{UP}.}
\SokobanRule
\SokobanCell{%
  \SokobanBoard{\includegraphics[width=0.49\linewidth,trim={0bp 384bp 256bp 96bp},clip]{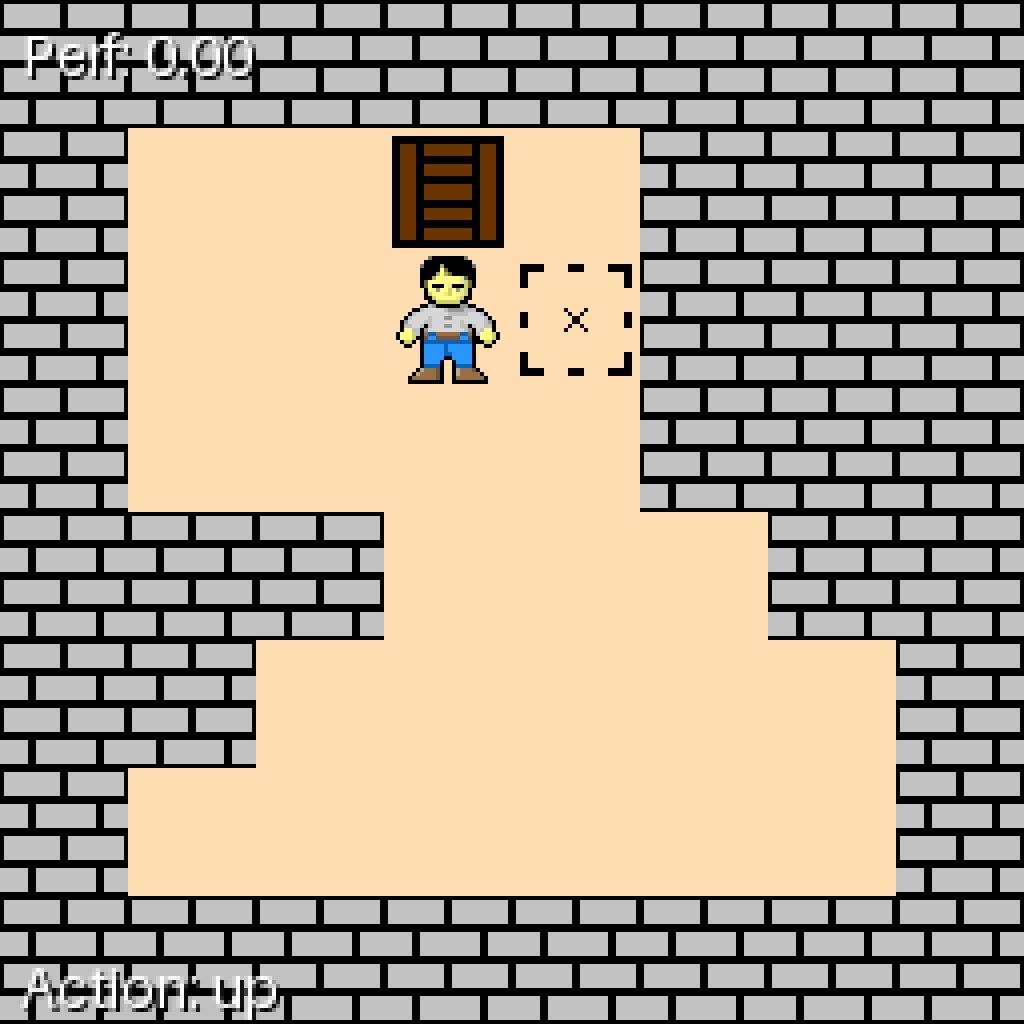}}
  \textbf{Result:} Up pushes the box against the upper wall.\par\smallskip
  \textbf{Sub-goal not achieved.}}
\end{assertionexample}
\end{minipage}\hfill
\begin{minipage}[t]{0.485\linewidth}
\vspace{0pt}
\begin{assertionexample}{ConPAct coordination}
\small
\SokobanCell{%
  \SokobanBoard{\includegraphics[width=0.49\linewidth,trim={0bp 384bp 256bp 96bp},clip]{figures/case_sokoban_0225_start.png}}
  \textbf{Planner:} ``The box is directly above you.''}
\SokobanRule
\SokobanCell{%
  \textbf{Actor:} ``Up is empty floor; Right is a wall. I will move Up to get a better view.''\par\smallskip
  \textbf{Proposed action:} \texttt{UP}.}
\SokobanRule
\SokobanCell{%
  \textbf{ConPAct $\rightarrow$ Actor \& Planner}\par\smallskip
  \begingroup
  \color{red!80!black}
  \textbf{Contradiction:} \texttt{player\_neighbors}\par
  \texttt{Up}: Planner=\texttt{Box}; Actor=\texttt{Cell}\par
  \texttt{Right}: Planner=\texttt{Cell}; Actor=\texttt{Wall}\par
  \endgroup\par\smallskip
  ``Recheck both readings; actions paused.''}
\SokobanRule
\SokobanCell{%
  \textbf{Actor:} ``Rechecking the image: Up is a box and Right is clear. Up would push the box; I withdraw it.''\par\smallskip
  \textbf{Planner:} ``I confirm both readings. Move Left, then Up; stop left of the box.''}
\SokobanRule
\SokobanCell{%
  \textbf{Actor:} ``Left is clear.''\par\smallskip
  \textbf{Action:} \texttt{LEFT}.}
\SokobanRule
\SokobanCell{%
  \SokobanBoard{\includegraphics[width=0.49\linewidth,trim={0bp 384bp 256bp 96bp},clip]{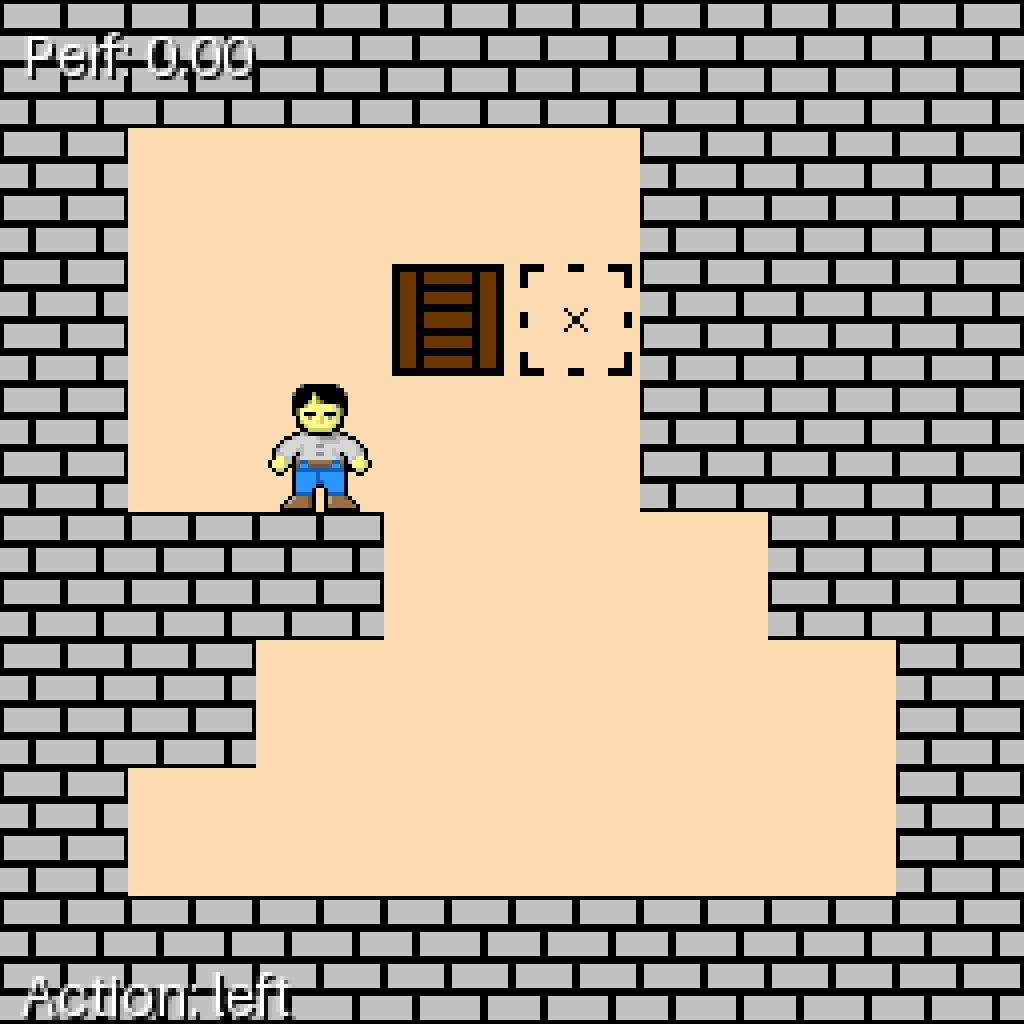}}
  \textbf{Actor:} ``Up is clear; I can reach the box's left side.''
  \par\smallskip
  \textbf{Action:} \texttt{UP}.}
\SokobanRule
\SokobanCell{%
  \SokobanBoard{\includegraphics[width=0.49\linewidth,trim={0bp 384bp 256bp 96bp},clip]{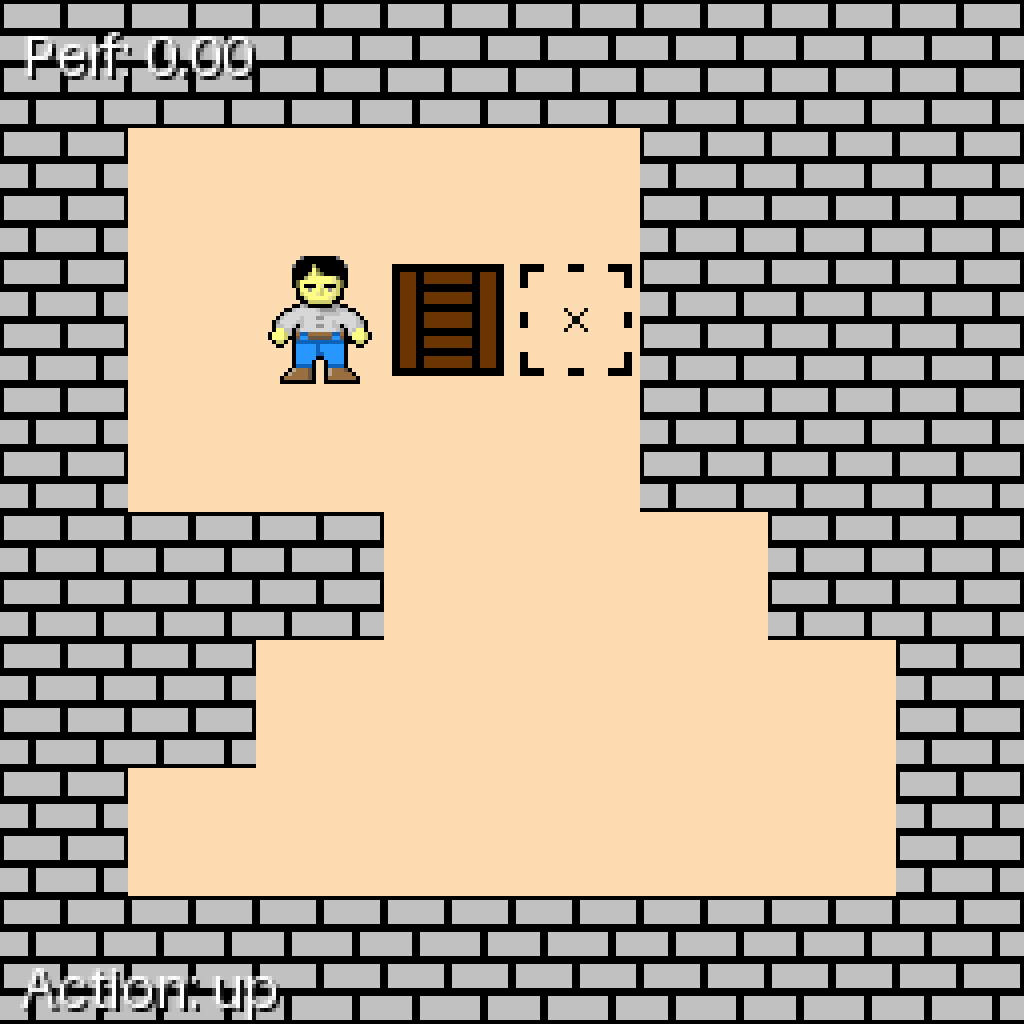}}
  \textbf{Actor:} ``The box is directly to my right. Sub-goal achieved.''\par\smallskip
  \textbf{Status:} \texttt{ACHIEVED}.}
\end{assertionexample}
\end{minipage}
\endgroup
  \caption{\textbf{Sokoban: state mismatch and ConPAct coordination.}
  Condensed recorded outputs for the same positioning sub-goal.}
  \label{fig:sokoban-reconciliation-case}
\end{figure}

\clearpage

\subsection{MiniGrid}
\label{app:minigrid-reconciliation}

\begin{figure}[H]
  \centering
\begingroup
\small
\setlength{\parindent}{0pt}
\newcommand{\MiniGridCell}[1]{%
  \begin{minipage}[t]{\linewidth}\raggedright #1\end{minipage}\par}
\newcommand{\MiniGridRule}{%
  \vspace{5pt}{\color{OursHighlight!65!black}\hrule height 0.3pt}\vspace{6pt}}
\newcommand{\MiniGridBoard}[1]{%
  {\centering#1\par}\vspace{4pt}}

\begin{tcolorbox}[
  enhanced, colback=OursHighlight!12, colframe=OursHighlight!65!black,
  boxrule=0.45pt, arc=1mm, left=7pt, right=7pt, top=7pt, bottom=7pt,
  before skip=0pt, after skip=5pt
]
\normalsize
\textbf{Sub-goal:} Move one cell \textbf{east} along the top row to stand directly above the lava gap.
\end{tcolorbox}

\begin{minipage}[t]{0.485\linewidth}
\vspace{0pt}
\colorlet{OursHighlight}{black!35}
\begin{assertionexample}{State mismatch}
\small
\MiniGridCell{%
  \MiniGridBoard{\includegraphics[width=0.59\linewidth]{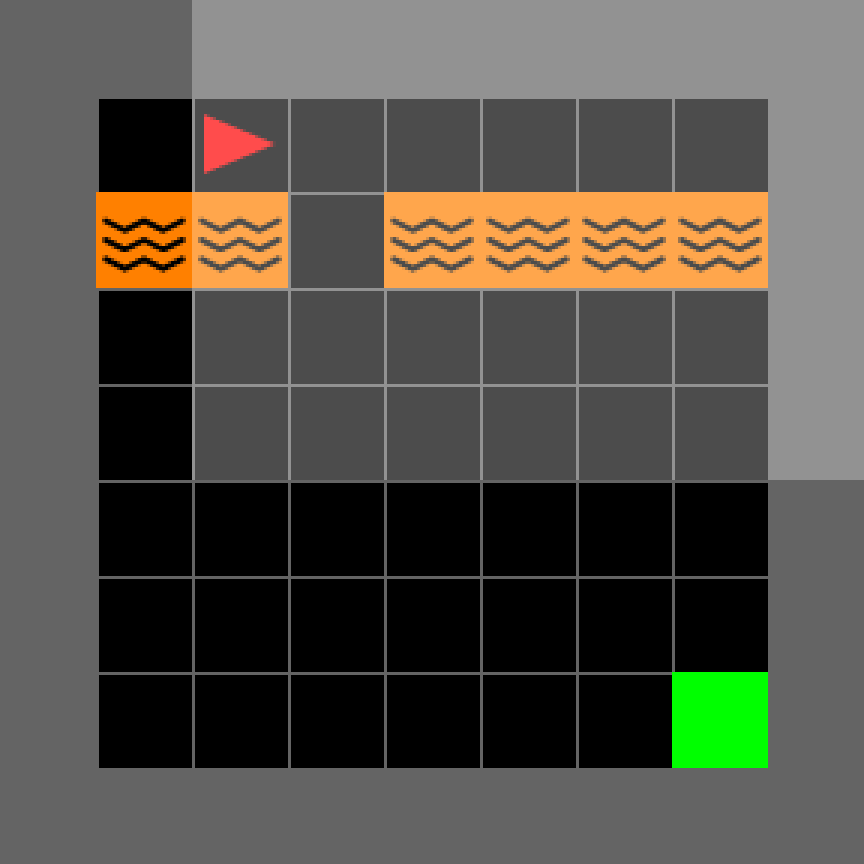}}
  \textbf{Planner:} ``The cell ahead is floor; the lava is in the row below.''}
\MiniGridRule
\MiniGridCell{%
  \textbf{Actor:} ``East is lava. I will turn south to avoid moving into it.''\par\smallskip
  \textbf{Action:} \texttt{right} (turn south).}
\MiniGridRule
\MiniGridCell{%
  \MiniGridBoard{\includegraphics[width=0.59\linewidth]{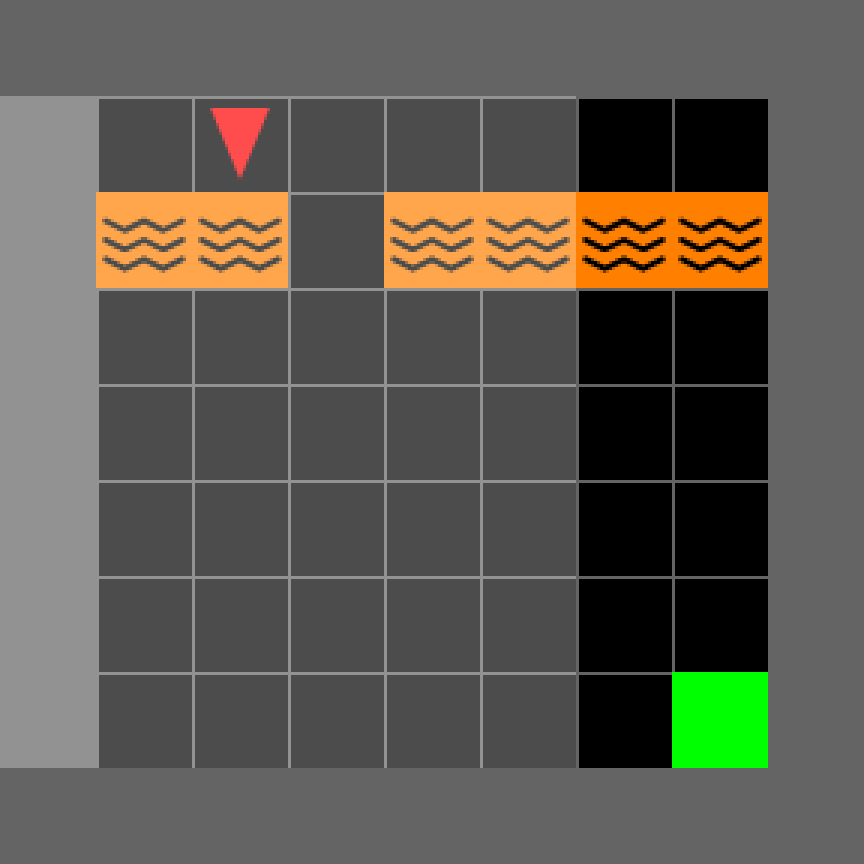}}
  \textbf{Actor:} ``Now lava is directly ahead. I will turn again.''\par\smallskip
  \textbf{Action:} \texttt{right} (turn west).}
\MiniGridRule
\MiniGridCell{%
  \MiniGridBoard{\includegraphics[width=0.59\linewidth]{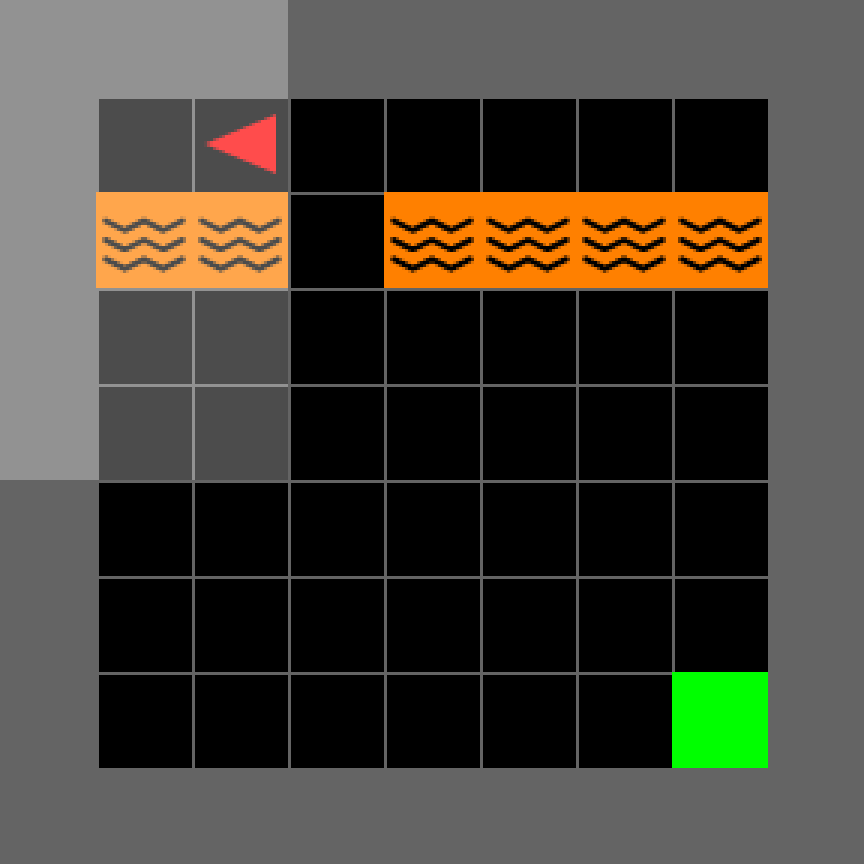}}
  \textbf{Result:} Two turns, no eastward movement. Still one cell left of the gap column.\par\smallskip
  \textbf{Sub-goal not achieved.}}
\end{assertionexample}
\end{minipage}\hfill
\begin{minipage}[t]{0.485\linewidth}
\vspace{0pt}
\begin{assertionexample}{ConPAct coordination}
\small
\MiniGridCell{%
  \MiniGridBoard{\includegraphics[width=0.59\linewidth]{figures/case_minigrid_1006_start.png}}
  \textbf{Planner:} ``The cell ahead is floor; the lava is in the row below.''}
\MiniGridRule
\MiniGridCell{%
  \textbf{Actor:} ``East is lava. I will turn south to avoid moving into it.''\par\smallskip
  \textbf{Proposed action:} \texttt{right}.}
\MiniGridRule
\MiniGridCell{%
  \textbf{ConPAct $\rightarrow$ Actor \& Planner}\par\smallskip
  \begingroup\color{red!80!black}
  \textbf{Contradiction:} \texttt{front\_cell}\par
  Planner=\texttt{empty}; Actor=\texttt{lava}\par
  \endgroup\par\smallskip
  ``Recheck both readings; actions paused.''}
\MiniGridRule
\MiniGridCell{%
  \textbf{Actor:} ``The triangle faces east. The dark cell ahead is floor; the orange lava is south. I withdraw the turn.''\par\smallskip
  \textbf{Planner:} ``I confirm the east cell is empty. All key and door counts remain zero. Move forward once, then stop above the gap.''}
\MiniGridRule
\MiniGridCell{%
  \textbf{Actor:} ``Agreed: empty ahead, all counts zero. I can move east.''\par\smallskip
  \textbf{Action:} \texttt{forward}.}
\MiniGridRule
\MiniGridCell{%
  \MiniGridBoard{\includegraphics[width=0.59\linewidth]{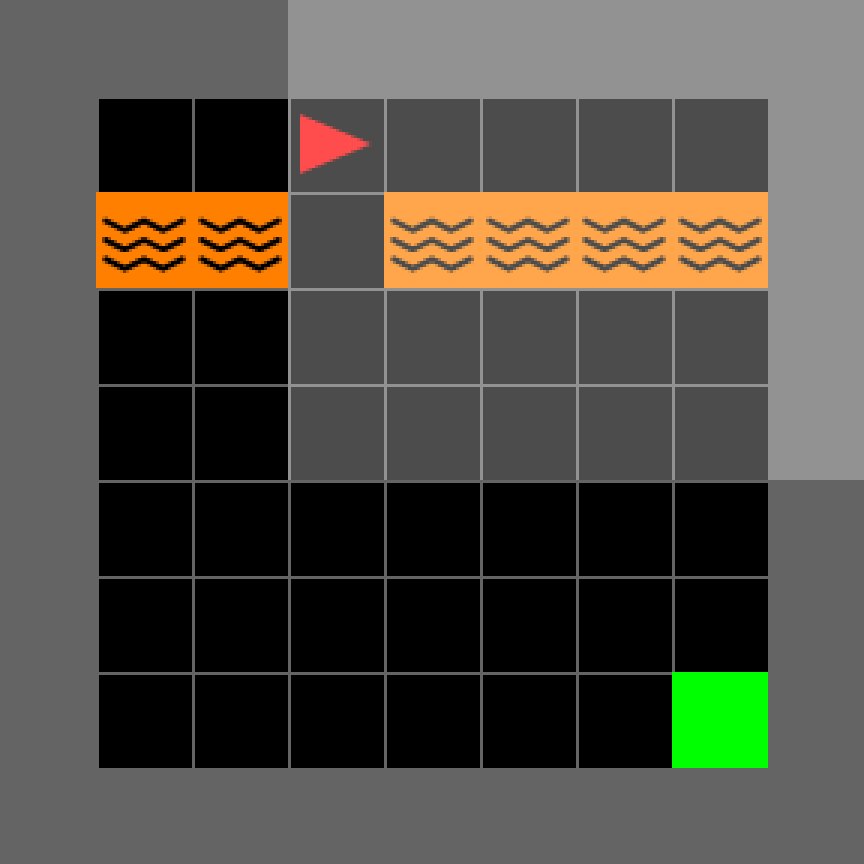}}
  \textbf{Actor:} ``I am directly above the gap, still on the top row.''\par\smallskip
  \textbf{Status:} \texttt{ACHIEVED}.}
\end{assertionexample}
\end{minipage}
\endgroup
  \caption{\textbf{MiniGrid: state mismatch and ConPAct coordination.}
  Condensed recorded outputs for the same gap-alignment sub-goal.}
  \label{fig:minigrid-reconciliation-case}
\end{figure}

\clearpage

\subsection{Crafter}
\label{app:crafter-reconciliation}

\begin{figure}[H]
  \centering
\begingroup
\small
\setlength{\parindent}{0pt}
\newcommand{\CrafterCell}[1]{%
  \begin{minipage}[t]{\linewidth}\raggedright #1\end{minipage}\par}
\newcommand{\CrafterRule}{%
  \vspace{4pt}{\color{OursHighlight!65!black}\hrule height 0.3pt}\vspace{4pt}}
\newcommand{\CrafterScene}[2]{%
  \begin{minipage}[c]{0.47\linewidth}
    #1
  \end{minipage}\hfill
  \begin{minipage}[c]{0.49\linewidth}\raggedright #2\end{minipage}\par}

\begin{tcolorbox}[
  enhanced, colback=OursHighlight!12, colframe=OursHighlight!65!black,
  boxrule=0.45pt, arc=1mm, left=7pt, right=7pt, top=7pt, bottom=7pt,
  before skip=0pt, after skip=5pt
]
\normalsize
\textbf{Sub-goal:} Harvest the nearby tree down-left to increase wood from \textbf{1 to 2}.
\end{tcolorbox}

\begin{minipage}[t]{0.485\linewidth}
\vspace{0pt}
\colorlet{OursHighlight}{black!35}
\begin{assertionexample}{State mismatch}
\small
\CrafterCell{%
  \CrafterScene{\includegraphics[width=\linewidth]{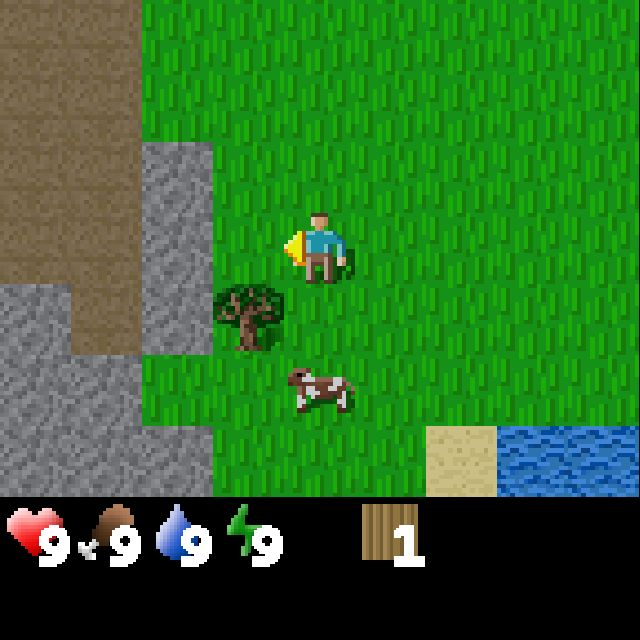}}{%
  \textbf{Planner:} ``The first tree is gone. The left cell is clear; another tree is down-left.''}}
\CrafterRule
\CrafterCell{%
  \textbf{Actor:} \texttt{facing=left}, \texttt{front=tree}.\par
  \textbf{Action:} \texttt{do}.}
\CrafterRule
\CrafterCell{%
  \CrafterScene{\includegraphics[width=\linewidth]{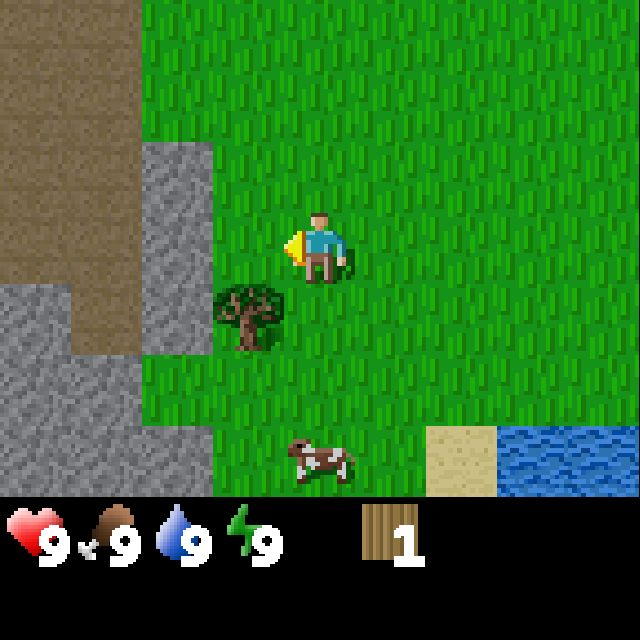}}{%
  \textbf{Result:} Wood remains 1.\par
  \textbf{Actor:} \texttt{front=tree}.\par
  \textbf{Action:} \texttt{do} again.}}
\CrafterRule
\CrafterCell{%
  \CrafterScene{\includegraphics[width=\linewidth]{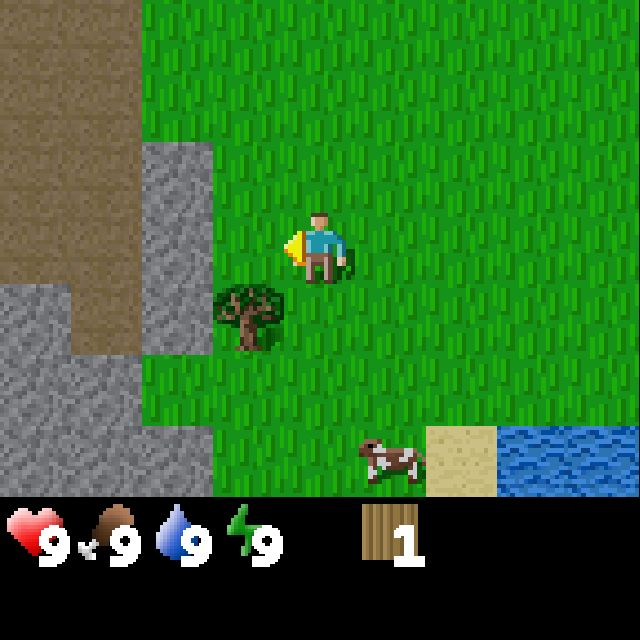}}{%
  \textbf{Actor:} ``The tree is down-left; I need to line up.''\par
  \textbf{Action:}\par\texttt{move\_down}.}}
\CrafterRule
\CrafterCell{%
  \CrafterScene{\includegraphics[width=\linewidth]{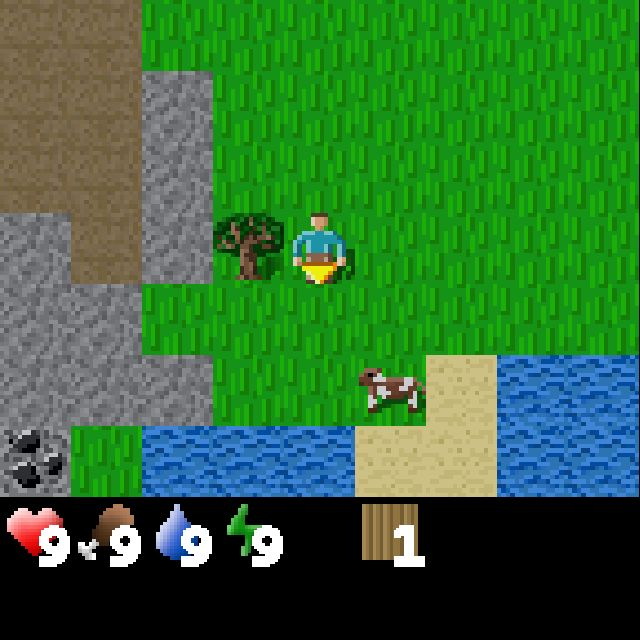}}{%
  \textbf{Result:} Now moving toward the tree; wood is still 1.\par\smallskip
  \textbf{Sub-goal not yet achieved.}}}
\end{assertionexample}
\end{minipage}\hfill
\begin{minipage}[t]{0.485\linewidth}
\vspace{0pt}
\begin{assertionexample}{ConPAct coordination}
\small
\CrafterCell{%
  \CrafterScene{\includegraphics[width=\linewidth]{figures/case_crafter_4137_start.png}}{%
  \textbf{Planner:} ``The first tree is gone. The left cell is clear; another tree is down-left.''}}
\CrafterRule
\CrafterCell{%
  \textbf{Actor:} \texttt{facing=left}, \texttt{front=tree}.\par
  \textbf{Proposed action:} \texttt{do}.}
\CrafterRule
\CrafterCell{%
  \textbf{ConPAct $\rightarrow$ Actor \& Planner}\par\smallskip
  \begingroup\color{red!80!black}
  \textbf{Contradiction:} \texttt{front}\par
  Planner=\texttt{walkable}; Actor=\texttt{tree}\par
  \endgroup\par\smallskip
  ``Recheck both readings; actions paused.''}
\CrafterRule
\CrafterCell{%
  \textbf{Actor:} ``I face left toward grass. The tree is down-left; I withdraw \texttt{do}.''\par\smallskip
  \textbf{Planner:} ``Confirmed: left, walkable, full vitals, one wood, no other resources or tools. Move left, then face down.''}
\CrafterRule
\CrafterCell{%
  \textbf{Actor:} ``Agreed; I will approach it.''\par
  \textbf{Action:} \texttt{move\_left}.}
\CrafterRule
\CrafterCell{%
  \CrafterScene{\includegraphics[width=\linewidth]{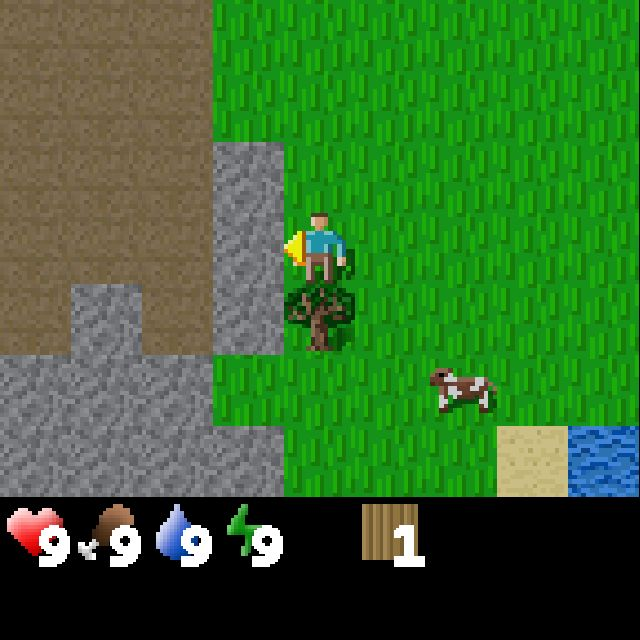}}{%
  \textbf{Actor:} ``The tree is directly below.''\par
  \textbf{Action:}\par\texttt{move\_down}\par(face the tree).}}
\CrafterRule
\CrafterCell{%
  \CrafterScene{\includegraphics[width=\linewidth]{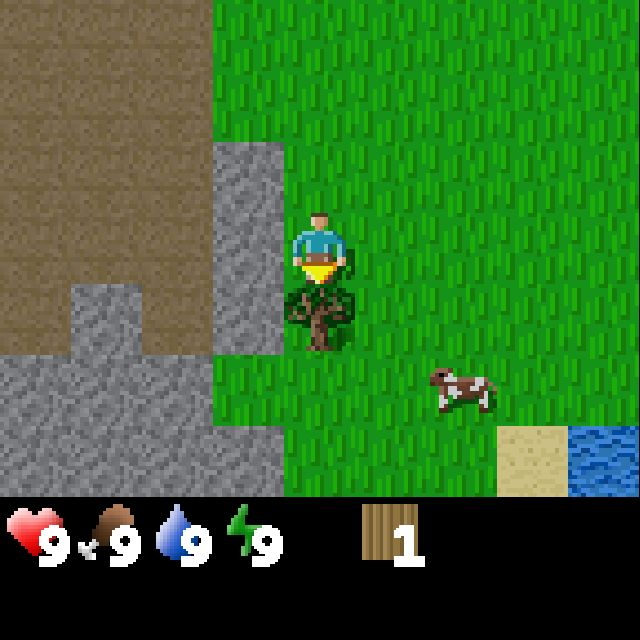}}{%
  \textbf{Actor:} ``Facing down, tree ahead.''\par
  \textbf{Action:} \texttt{do}.}}
\CrafterRule
\CrafterCell{%
  \CrafterScene{\includegraphics[width=\linewidth]{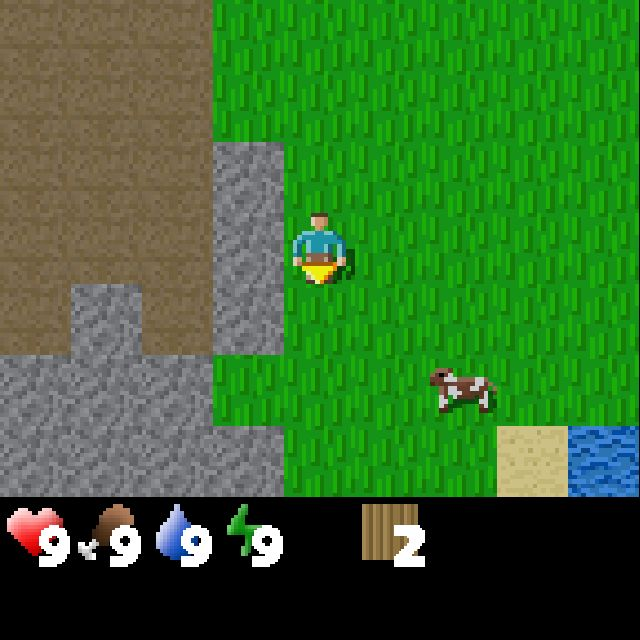}}{%
  \textbf{Actor:} ``The tree is gone; wood is 2.''\par
  \textbf{Status:}\par\texttt{ACHIEVED}.}}
\end{assertionexample}
\end{minipage}
\endgroup
  \caption{\textbf{Crafter: state mismatch and ConPAct coordination.}
  Condensed recorded outputs for the same second-wood sub-goal.}
  \label{fig:crafter-reconciliation-case}
\end{figure}

\clearpage

\subsection{Procgen Miner}
\label{app:procgen-reconciliation}

\begin{figure}[H]
  \centering
\begingroup
\small
\setlength{\parindent}{0pt}
\newcommand{\ProcgenCell}[1]{%
  \begin{minipage}[t]{\linewidth}\raggedright #1\end{minipage}\par}
\newcommand{\ProcgenRule}{%
  \vspace{5pt}{\color{OursHighlight!65!black}\hrule height 0.3pt}\vspace{6pt}}
\newcommand{\ProcgenBoard}[1]{%
  {\centering#1\par}\vspace{4pt}}

\begin{tcolorbox}[
  enhanced, colback=OursHighlight!12, colframe=OursHighlight!65!black,
  boxrule=0.45pt, arc=1mm, left=7pt, right=7pt, top=7pt, bottom=7pt,
  before skip=0pt, after skip=5pt
]
\normalsize
\textbf{Sub-goal:} Move \textbf{right one cell} to position for digging a path to the diamonds.
\end{tcolorbox}

\begin{minipage}[t]{0.485\linewidth}
\vspace{0pt}
\colorlet{OursHighlight}{black!35}
\begin{assertionexample}{State mismatch}
\small
\ProcgenCell{%
  \ProcgenBoard{\includegraphics[width=0.61\linewidth]{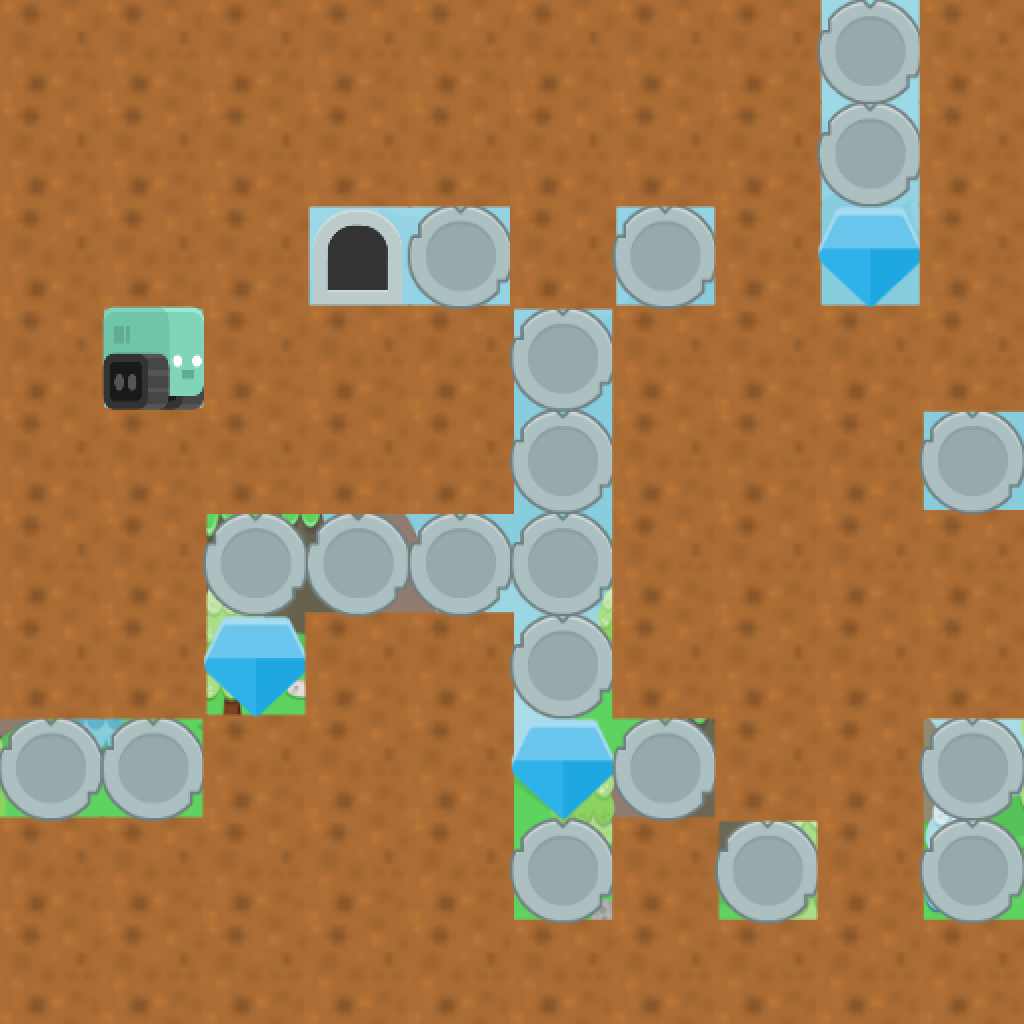}}
  \textbf{Planner:} ``No boulder is immediately overhead. Start by moving right.''}
\ProcgenRule
\ProcgenCell{%
  \textbf{Actor:} ``Boulders directly to the right make that move dangerous. I will move down.''\par\smallskip
  \textbf{Action:} \texttt{down}.}
\ProcgenRule
\ProcgenCell{%
  \ProcgenBoard{\includegraphics[width=0.61\linewidth]{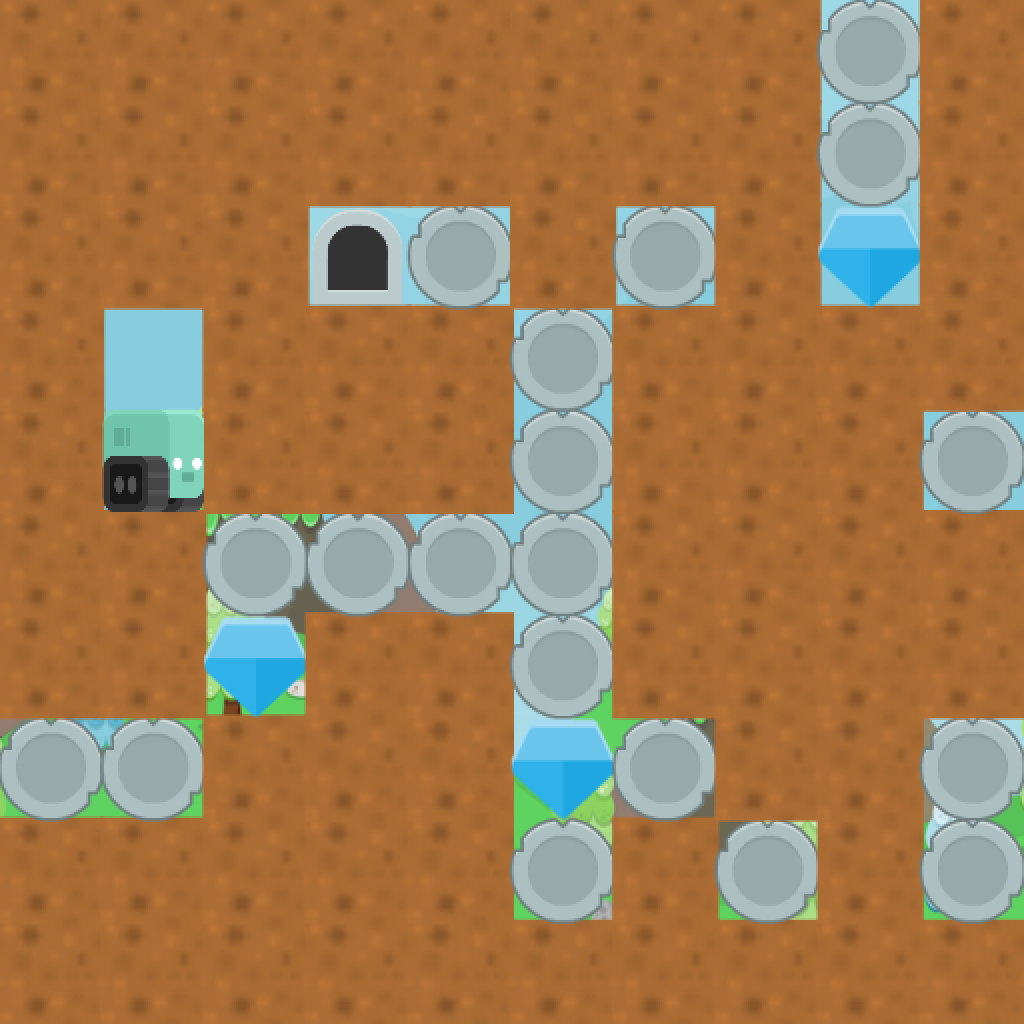}}
  \textbf{Result:} The player moves down one cell, remaining in the starting column.\par\smallskip
  \textbf{Sub-goal not achieved.}}
\end{assertionexample}
\end{minipage}\hfill
\begin{minipage}[t]{0.485\linewidth}
\vspace{0pt}
\begin{assertionexample}{ConPAct coordination}
\small
\ProcgenCell{%
  \ProcgenBoard{\includegraphics[width=0.61\linewidth]{figures/case_procgen_miner103_start.png}}
  \textbf{Planner:} ``No boulder is immediately overhead. Start by moving right.''}
\ProcgenRule
\ProcgenCell{%
  \textbf{Actor:} ``Boulders directly to the right make that move dangerous. I will move down.''\par\smallskip
  \textbf{Proposed action:} \texttt{down}.}
\ProcgenRule
\ProcgenCell{%
  \textbf{ConPAct $\rightarrow$ Actor \& Planner}\par\smallskip
  \begingroup\color{red!80!black}
  \textbf{Contradiction:} \texttt{hazard\_nearby}\par
  Planner=\texttt{no}; Actor=\texttt{yes}\par
  \endgroup\par\smallskip
  ``Recheck both readings; actions paused.''}
\ProcgenRule
\ProcgenCell{%
  \textbf{Actor:} ``The adjacent right cell is dirt; the boulder column is farther away. I revise \texttt{hazard\_nearby} to \texttt{no}. I also count three visible diamonds.''\par\smallskip
  \textbf{Planner:} ``I confirm the next right cell is clear and revise my diamond count to three. Diamonds are visible; no immediate falling hazard is apparent. Stop after one right step.''}
\ProcgenRule
\ProcgenCell{%
  \textbf{Actor:} ``Agreed: \texttt{target\_visible=yes}, \texttt{hazard\_nearby=no}, three diamonds.''\par\smallskip
  \textbf{Action:} \texttt{right}.}
\ProcgenRule
\ProcgenCell{%
  \ProcgenBoard{\includegraphics[width=0.61\linewidth]{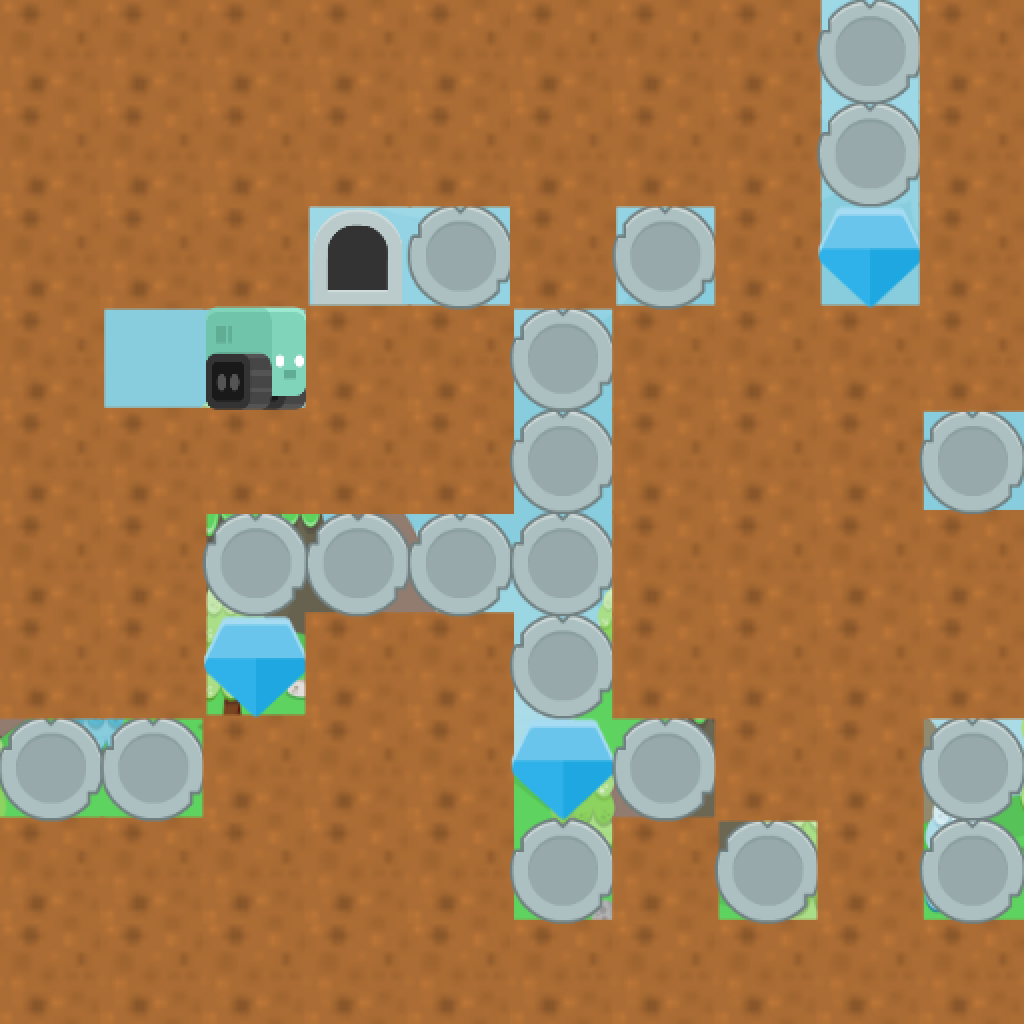}}
  \textbf{Actor:} ``I have moved one cell right. The positioning sub-goal is complete.''\par\smallskip
  \textbf{Status:} \texttt{ACHIEVED}.}
\end{assertionexample}
\end{minipage}
\endgroup
  \caption{\textbf{Procgen Miner: state mismatch and ConPAct coordination.}
  Condensed recorded outputs for the same one-cell positioning sub-goal.}
  \label{fig:procgen-reconciliation-case}
\end{figure}

\clearpage

\FloatBarrier
\section{Discussion}
\label{app:discussion}

\subsection{Inference Cost}
\label{app:inference-cost}

Table~\ref{tab:inference-call-costs} summarizes the inference-time comparisons across the four game environments.
We average rollout costs within each model--environment configuration and then aggregate across environments and evaluated configurations.
Calls per action is the ratio of mean model calls to mean executed actions under this aggregation.
Output-token averages use the runs with recorded usage; the coverage column reports the fraction with available usage information.
All decision calls, including planning and discussion, count toward the common rollout budget.

\begin{table}[htbp]
\centering
\small
\caption{Inference costs averaged across the four environments and the evaluated inference-time configurations. Usage coverage is the fraction of recorded rollouts with output-token usage.}
\label{tab:inference-call-costs}
\setlength{\tabcolsep}{6pt}
\renewcommand{\arraystretch}{1.12}
\begin{tabular}{lrrrr}
\toprule
Method & \makecell{Calls /\\episode} & \makecell{Output tokens /\\episode} & \makecell{Calls /\\action} & \makecell{Usage\\coverage (\%)} \\
\midrule
ReAct (Planner) & 14.02 & 3,095.37 & 1.04 & 71.5 \\
ReAct (Actor) & 36.29 & 8,703.19 & 1.00 & 71.5 \\
Plan-Act & 35.36 & 5,285.03 & 1.06 & 100.0 \\
HiPlan & 51.86 & 7,992.10 & 2.10 & 100.0 \\
TAPE & 37.84 & 18,900.17 & 6.29 & 99.9 \\
\OursRow \textbf{ConPAct-I} & 33.02 & 7,199.86 & 1.43 & 83.3 \\
\bottomrule
\end{tabular}
\end{table}

\subsection{Interpretation and Scope}
\label{app:interpretation-scope}

State assertions describe selected task-relevant facts rather than complete internal beliefs.
Agreement alone does not ensure correctness, so the Sokoban and MiniGrid analyses evaluate the roles' reports against environment ground truth as well as against each other.
Procgen and OSWorld allow unknown values for unobservable facts; their task-performance results are not treated as substitutes for an assertion-correctness oracle.
The GPT-5.6-sol reviewer curates ConPAct-R data offline and is not required by online ConPAct comparison or reconciliation.

\paragraph{Consistency and correctness.}
The relationship is asymmetric: when both roles correctly report the same definite facts about the same observation, they must agree, whereas agreement can also reflect a shared error.
Disagreement therefore signals that at least one report needs revision, but does not identify which report is correct; agreement leaves correlated errors undetected.
This distinction matters especially during reconciliation: a decrease in mismatch can arise either from correcting an erroneous report or from persuading a correct role to adopt an incorrect interpretation.
The CC/CW/D analysis separates these outcomes by tracking whether agreement is grounded in the environment or merely shared between roles.
Accordingly, reduced mismatch indicates closer alignment of the reported states, but supports a claim of improved state understanding only when accompanied by greater correctness.

Consistency also has a practical role beyond measuring individual accuracy: a planner and an actor need compatible state assumptions for a sub-goal to retain its meaning across their interaction.
Yet this compatibility alone cannot establish that the sub-goal is feasible or the chosen action is effective.
ConPAct thus addresses an observable source of coordination failure, while shared perceptual errors and planning errors remain distinct limitations.
For unobservable fields, preserving uncertainty avoids replacing missing evidence with unsupported consensus; the absence of a definite conflict should not itself be interpreted as verified agreement.

\FloatBarrier

\end{document}